\documentclass[twoside]{article}

\usepackage{siunitx,hyperref,lipsum}

\usepackage{nameref,zref-xr}
\zxrsetup{toltxlabel}
\zexternaldocument*{supplement}

\usepackage[round]{natbib}

\usepackage[utf8]{inputenc} % allow utf-8 input
\usepackage[T1]{fontenc}    % use 8-bit T1 fonts
\usepackage{hyperref}       % hyperlinks
\usepackage{url}            % simple URL typesetting
\usepackage{booktabs}       % professional-quality tables
\usepackage{amsfonts}       % blackboard math symbols
\usepackage{nicefrac}       % compact symbols for 1/2, etc.
\usepackage{microtype}      % microtypography
\usepackage{xcolor}         % colors

\usepackage[utf8]{inputenc} % allow utf-8 input
\usepackage[T1]{fontenc}    % use 8-bit T1 fonts
\usepackage{hyperref}       % hyperlinks
\usepackage{url}            % simple URL typesetting
\usepackage{booktabs}       % professional-quality tables
\usepackage{amsfonts}       % blackboard math symbols
\usepackage{nicefrac}       % compact symbols for 1/2, etc.
\usepackage{microtype}      % microtypography
\usepackage{xcolor}         % colors

\usepackage{microtype}
\usepackage{graphicx}
\usepackage{booktabs} % for professional tables

\usepackage{microtype}
\usepackage{graphicx}
\usepackage{booktabs} % for professional tables

\usepackage{amssymb, amsmath,amsthm}
\usepackage{amsfonts}
\usepackage{algorithm}
\usepackage{algorithmic}
\usepackage{paralist}
\usepackage{multirow}
\usepackage{wrapfig}
\usepackage{url,enumerate}
\usepackage{color,xcolor}
\usepackage{makeidx}  % allows for indexgeneration
\usepackage{amsmath,amssymb}
\usepackage{mathtools}
\usepackage[small, compact]{titlesec}
\usepackage{xspace}
\usepackage{epstopdf}
\usepackage{cite}
\usepackage{mathrsfs}
\usepackage{enumerate}
\usepackage{color}
\usepackage{graphicx,epsfig}
\usepackage{amsmath,amssymb,xspace}
\usepackage{url}
\usepackage{hyperref}
\usepackage{bm}
\usepackage{bbm}
\usepackage{upgreek}
\usepackage{cleveref}
\usepackage{multirow}
\usepackage{ulem}
\usepackage{cancel}
\usepackage{subcaption}
\usepackage{dsfont}
\usepackage{hyperref}
\newcommand{\zhec}[1]{[\textcolor{blue}{#1}]}

\newcommand{\ours}{{KBASS}\xspace}
\newcommand{\cmt}[1]{}
\newcommand{\eg}{{\textit{e.g.},}\xspace}
\newcommand{\ie}{{\textit{i.e.},}\xspace}
\newcommand{\etc}{{\textit{etc}.}\xspace}

\newcommand{\kron}{\otimes}

\renewcommand{\vec}{{\rm vec}}

\newcommand{\brho}{\boldsymbol{\rho}}
\renewcommand{\a}{{\bf a}}

\renewcommand{\d}{{\rm d}}  % for derivatives

\newcommand{\f}{{\bf f}}
\newcommand{\g}{{\bf g}}
\newcommand{\h}{{\bf h}}

\renewcommand{\k}{{\bf k}}
\newcommand{\m}{{\bf m}}

\newcommand{\s}{{\bf s}}

\renewcommand{\u}{{\bf u}}
\renewcommand{\v}{{\bf v}}
\newcommand{\w}{{\bf w}}
\newcommand{\x}{{\bf x}}
\newcommand{\y}{{\bf y}}
\newcommand{\z}{{\bf z}}
\newcommand{\A}{{\bf A}}

\newcommand{\Hcal}{{\mathcal{H}}}
\newcommand{\Dcal}{\mathcal{D}}
\renewcommand{\H}{{\bf H}}
\newcommand{\I}{{\bf I}}

\newcommand{\K}{{\bf K}}
\renewcommand{\L}{{\bf L}}
\newcommand{\Lcal}{{\mathcal{L}}}

\newcommand{\Mcal}{{\mathcal{M}}}
\newcommand{\gp}{\mathcal{GP}}  % for normal density
\newcommand{\Ocal}{{\mathcal{O}}}
\newcommand{\Fcal}{{\mathcal{F}}}
\newcommand{\N}{\mathcal{N}}  % for normal density

\newcommand{\Ucal}{{\mathcal{U}}}

\newcommand{\wy}{{\widehat{\y}}}

\newcommand{\X}{{\bf X}}
\newcommand{\Xcal}{{\mathcal{X}}}

\newcommand{\Zcal}{{\mathcal{Z}}}

\newcommand{\balpha}{\boldsymbol{\alpha}}

\newcommand{\bphi}{\boldsymbol{\phi}}
\newcommand{\bPhi}{\boldsymbol{\Phi}}
\newcommand{\bbeta}{\boldsymbol{\beta}}

\newcommand{\blambda}{\boldsymbol{\lambda}}

\newcommand{\btheta}{\boldsymbol{\theta}}

\newcommand{\bxi}{\boldsymbol{\xi}}
\newcommand{\bSigma}{\boldsymbol{\Sigma}}

\newcommand{\bOmega}{\boldsymbol{\Omega}}

\newcommand{\bgamma}{\boldsymbol{\gamma}}

\newcommand{\bmu}{\boldsymbol{\mu}}

\newcommand{\0}{{\bf 0}}

\newcommand{\ben}{\begin{enumerate}}
\newcommand{\een}{\end{enumerate}}

\newcommand{\argmin}{\operatornamewithlimits{argmin}}

\newcommand{\LN}{\mathcal{LN}}

\newcommand{\Pcal}{{\mathcal{P}}}

\newcommand{\EE}{\mathbb{E}}
\newcommand{\pa}[1]{{\partial_{{#1}}}}
\newcommand{\cov}{{\text{cov}}}

\newcommand{\tp}{{\widetilde{p}}}
\newcommand{\tmu}{{\widetilde{\mu}}}
\newcommand{\tv}{{\widetilde{v}}}

\newcommand{\bkh}{{\backslash}}

\newcommand{\kl}{{\text{KL}}}

\newcommand{\michael}[1]{{}}
\newcommand{\michaeladdressed}[1]{{}}
\newcommand{\shandian}[1]{{}}
\newcommand{\aditi}[1]{{}}

\usepackage[accepted]{aistats2024}
\begin{document}



\runningtitle{Equation Discovery with Bayesian Spike-and-Slab Priors and Efficient Kernels}
\runningauthor{Long, Xing, Krishnapriyan, Kirby, Zhe, and Mahoney}

% \newcommand{\affiliationOne}{theU}{University of Utah}
% \newcommand{\affiliationOne}{UCB}{University of California, Berkeley}
% \newcommand{\affiliationOne}{Lawrence}{Lawrence Berkeley National Laboratory}
% \newcommand{\affiliationOne}{ICSI}{International Computer Science Institute}

% \icmlcorrespondingauthor{Shandian Zhe}{zhe@cs.utah.edu}

\twocolumn[

\aistatstitle{Equation Discovery with Bayesian \\Spike-and-Slab Priors and Efficient Kernels}

\aistatsauthor{ Da Long \And Wei Xing \And  Aditi S. Krishnapriyan}  

\aistatsaddress{ The University of Utah \And The University of Sheffield \And  University of California, Berkeley}
\aistatsauthor{Robert M. Kirby  \And Shandian Zhe$^*$  \And Michael W. Mahoney }

\aistatsaddress{ The University of Utah \And The University of Utah  \And University of California, Berkeley\\Lawrence Berkeley National Laboratory\\International Computer Science Institute } ]

\begin{abstract}
Discovering governing equations from data is important to many scientific and engineering applications. Despite promising successes, existing methods are still challenged by data sparsity and noise issues, both of which are ubiquitous in practice. Moreover, state-of-the-art methods lack uncertainty quantification and/or are costly in training. To overcome these limitations, we propose a novel equation discovery method based on Kernel learning and BAyesian Spike-and-Slab priors (\ours). We use kernel regression to estimate the target function, which is flexible, expressive, and more robust to data sparsity and noises. We combine it with a Bayesian spike-and-slab prior --- an ideal Bayesian sparse distribution --- for effective operator selection and uncertainty quantification. We develop an expectation-propagation expectation-maximization (EP-EM) algorithm for efficient posterior inference and function estimation. To overcome the computational challenge of kernel regression, we place the function values on a mesh and induce a Kronecker product construction, and we use tensor algebra to enable efficient computation and optimization. We show the advantages of \ours on a list of benchmark ODE and PDE discovery tasks. The code is available at \url{https://github.com/long-da/KBASS}.

\cmt{
Discovering governing equations from data is important to many scientific and engineering applications. Despite promising successes, existing methods are still challenged by data sparsity as well as noise issues, both of which are ubiquitous in practice. Moreover, state-of-the-art methods lack uncertainty quantification and/or are costly in training. To overcome these limitations, we propose a novel equation discovery method based on Kernel learning and BAyesian Spike-and-Slab priors (\ours). We use kernel regression to estimate the target function, which is flexible, expressive, and more robust to data sparsity and noises. We combine with Bayesian spike-and-slab prior --- an ideal Bayesian sparse distribution --- for effective operator selection and uncertainty quantification. We develop an expectation propagation expectation-maximization (EP-EM) algorithm for efficient posterior inference and function estimation. To overcome the computational challenge of kernel regression, we place the function values on a mesh and induce a Kronecker product. We then use tensor algebra to enable efficient computation and optimization. We show significant advantages of \ours on a list of benchmark ODE and PDE discovery tasks.% 
}
\end{abstract}

%discover of governing equations is important ---> current methods reference --> problems of the current method: suffer from sparse & noisy data, lack UQ and/or computationally cost, require heavyt tunning --> our contributions: model (kernel method) , s&s prior, algorithm scalability, EP-EM, multitask, results

\section{\MakeUppercase{Introduction}}
Many scientific and engineering disciplines use differential equations to model systems of interest. These equations not only allow accurate predictions (by numerical solvers), but they also provide interpretation of the mechanism, \ie physical laws. However, for many realistic systems, it is difficult in practice to write down a full set of partial or ordinary differential equations (PDEs/ODEs). Hence, using data-driven machine learning methods to help discover underlying physical equations is a promising \cmt{(perhaps even critical, depending on the application)} approach to advance our understanding of these systems and to develop powerful, reliable predictive and analysis tools. 

There have been several approaches for equation discovery, including: 
Sparse Identification of Nonlinear Dynamics (SINDy)~\citep{brunton2016discovering} and extensions~\citep{rudy2017data,schaeffer2017learning,zhang2020data,lagergren2020learning}; 
physics-informed neural networks (PINNs)~\citep{raissi2019physics} with $L_1$ regularization~\citep{berg2019data,both2021deepmod}; PINN using alternating direction optimization to identify equation operators for symbolic regression (PINN-SR)~\citep{chen2021physics}; and the 
recent work~\citep{sun2022bayesian} using Bayesian spline learning (BSL) to estimate the target function and relevance vector machine~\citep{tipping2001sparse} to select the candidate~operators.

Despite promising successes, the existing  methods still have challenges with data sparsity and noise. In addition, for practical usage, they often lack uncertainty calibration and/or are costly in training.   
For example, SINDy uses numerical differentiation to evaluate the derivatives of candidate operators, and it can easily fail on sparsely sampled data. While PINN-based approaches avoid numerical differentiation by using neural networks (NNs) to approximate the target function, they need a careful choice of the network architectures and tedious tuning of many hyperparameters, and they are susceptible to well-known failure modes~\citep{krishnapriyan2021characterizing}. Applying differentiation operators on NNs further complicates the loss  and makes the training expensive~\citep{krishnapriyan2021characterizing,mojgani2022lagrangian}. Yet PINN-based methods typically underperform in ODE discovery tasks~\citep{sun2022bayesian}. In addition, 
both SINDy and PINN methods lack uncertainty quantification of the learned equations. Although the most recent BSL method comes with uncertainty estimation, %it also needs heavy tuning of many hyperparameters, such as truncation threshold, regularization strength, order, and knot locations and numbers. 
it is quite costly to learn the spline coefficients, especially in higher dimensions. 

To promote the performance of data-driven equation discovery and to be more amenable for practical usage, we propose \ours, a novel method based on Kernel learning and BAyesian Spike-and-Slab priors. 
The major contributions of our work are as follows:
%To promote the performance of data-driven equation discovery and to be more amenable for practical usage, we propose a novel equation discovery method based on Kernel learning and BAyesian Spike-and-Slab priors (\ours). 
%The major contributions are as follows.% of our work are as follows. 
\begin{compactitem}
   \item 
   \textbf{Model}. We use kernel regression to estimate the target function and its derivatives, an approach that is flexible, expressive, and more robust to data sparsity and noise. We combine this with a Bayesian spike-and-slab prior to select equation operators and to estimate the operator weights. As a gold standard in Bayesian sparse learning, the spike-and-slab prior has nice theoretical properties, such as selective shrinkage~\citep{ishwaran2005spike}. It also enables posterior inference of selection indicators, with which we can decide the selection results and avoid hard thresholding over the weight values. Tuning the weight threshold (as in existing methods) can be troublesome and inconvenient in practice.   
   
   \item 
   \textbf{Algorithm}. To overcome the computational challenge of kernel learning, we place the function values on a mesh and induce a Kronecker product in the kernel matrices. We use Kronecker product properties and tensor algebra methods to enable highly efficient computation.\cmt{, without the need for any low-rank approximations.}  We then develop an expectation-propagation expectation-maximization (EP-EM) algorithm for efficient alternating posterior inference and function estimation. In the E step, we perform EP to infer quickly the posterior of the selection indicators and operator weights; while in the M step, we maximize the expected model likelihood to solve the current equation (identified by EP) and update the function estimate and kernel parameters. 
   \item 
   \textbf{Results}.  
   We examine \ours in discovering a number of benchmark equations, including three ODE systems, a nonlinear diffusion-reaction equation (diffusion rate $10^{-4}$), Burgers' equations with small viscosity (0.1, 0.01, 0.005), and a Kuramoto-Sivashinsky equation. 
   We compared \ours with state-of-the-art data-driven equation discovery methods. 
   We observed that \ours can use a much smaller number of data points in the presence of much more noise and still recover the equations, while the competing methods failed in most cases; and that \ours obtained more accurate weight estimation and provided reasonable uncertainty calibration. 
   Meanwhile, \ours exhibits significant advantages in computational efficiency. 
\end{compactitem}
\cmt{
Many scientific and engineering disciplines use differential equations to model systems of interest. These equations allow accurate predictions with numerical solvers, and they also provide interpretation of the mechanism, \ie physical laws. However, for many realistic systems, it is difficult in practice to write down a full set of partial or ordinary differential equations (PDEs/ODEs). Hence, using data-driven machine learning methods to help discover underlying physical equations is a promising (perhaps even critical, depending on the application) approach to advance our understanding of these systems and to develop powerful, reliable predictive and analysis tools. 

There have been several approaches for equation discovery, including: 
Sparse Identification of Nonlinear Dynamics (SINDy)~\citep{brunton2016discovering} and extensions~\citep{rudy2017data,schaeffer2017learning,zhang2020data,lagergren2020learning}; 
physics-informed neural networks (PINNs)~\citep{raissi2019physics} with $L_1$ regularization~\citep{berg2019data,both2021deepmod}; PINN using alternating direction optimization to identify equation operators for symbolic regression (PINN-SR)~\citep{chen2021physics}; and the recent work~\citep{sun2022bayesian} that uses Bayesian spline learning (BSL) to estimate the target function and relevant vector machine~\citep{tipping2001sparse} to select the candidate operators.

Despite the promising successes, the existing  methods still have challenges with data sparsity and noise. In addition, for practical usage, they often lack uncertainty calibration and/or are costly in training. % demand heavy hyper-parameter tuning.  
For example, SINDy uses numerical differentiation to evaluate the derivatives of candidate operators, and can easily fail on sparsely sampled data. While PINN-based approaches avoid numerical differentiation by using neural networks (NNs) to approximate the target function,  they need a careful choice of the network architectures and tedious tuning of many hyperparameters. Applying differentiation operators on NNs further complicates the loss and makes the training expensive~\citep{krishnapriyan2021characterizing,mojgani2022lagrangian, du2023neural}. Further, PINN-based methods typically underperform in ODE discovery tasks~\citep{sun2022bayesian}. In addition, 
both the SINDy and PINN methods lack uncertainty quantification of the learned equations. Although the most recent BSL method comes with uncertainty estimation, 
it is quite costly to learn the spline coefficients, especially in higher dimensions. 

To promote the performance of data-driven equation discovery and to be more amenable for practical usage, we propose \ours, a novel approach that couples kernel learning and Bayesian spike-and-slab priors. 
The major contributions of our work are as follows. 
\begin{compactitem}
   \item 
   \textbf{Model}. We use kernel regression to estimate the target function and its derivatives, an approach that is flexible, expressive and more robust to data sparsity and noises. We then combine with Bayesian spike-and-slab prior to select equation operators and to estimate the operator weights. As a golden standard in Bayesian sparse learning, the spike-and-slab prior not only enjoys nice theoretical properties, such as selective shrinkage~\citep{ishwaran2005spike}, it also enables posterior inference of selection indicators, with which we can decide the selection results and avoid hard thresholding over the weight values. Tuning the weight threshold (as in existing methods) can be troublesome and inconvenient in practice.   
   
   \item 
   \textbf{Algorithm}. To overcome the computational challenge of kernel learning, we place the function values on a mesh and induce a Kronecker product in the kernel matrices. We use Kronecker product properties and tensor algebra to enable highly efficient computation.\cmt{, without the need for any low-rank approximations.}  We then develop an expectation-propagation expectation-maximization (EP-EM) algorithm for efficient alternating posterior inference and function estimation. In the E step, we perform EP to fast infer the posterior of the selection indicators and operator weights, while in the M step, we maximize the expected model likelihood to solve the current equation (identified by EP) and update the function estimate and kernel parameters. 
   \item 
   \textbf{Results}.  
   We examined \ours in discovering a list of benchmark equations, including three ODE systems, a nonlinear diffusion-reaction equation (diffusion rate $10^{-4}$), Burger's equations with small viscosity (0.1, 0.01, 0.005), and a Kuramoto-Sivashinsky equation. We compared with state-of-the-art data-driven equation discovery methods. 
   \ours can use a much smaller number of data points under much more noises to recover the equations, while the competing methods failed in most cases.  \ours obtained more accurate weight estimation and provided reasonable uncertainty calibration. Meanwhile, \ours exhibits significant advantages in computational efficiency. 
\end{compactitem}
}
%background --> Model --> Algorithm --> Related Work 
%model --> start setting, and talk about a grid --> mention this is useful and will discuss about it later --> and then go to the next 
%GP prior 
%\vspace{-0.13in}
%1. Gaussian process 2. horseshoe
\section{\MakeUppercase{Preliminaries}}
\label{sect:bk}
%\vspace{-0.05in}
%Gp part can still be tuncated, maybe rmeoving the noisy likelihood, just mention the kernel interpolation 

%Here, we describe some preliminaries needed to understand our main results.
%More details on other related work may be found in Section~\ref{sxn:related-work}.

%\noindent 
\noindent \textbf{Kernel Regression.} Denote by $\Hcal_\kappa$ the Reproducing Kernel Hilbert Space (RKHS) induced by a Mercer kernel function $\kappa(\cdot, \cdot)$, and by $\|\cdot\|_{\Hcal_\kappa}$ its norm. %A commonly used kernel function is the square exponential (SE) kernel, $\kappa(\x, \x') = \exp(-\eta \|\x - \x'\|^2)$
%\begin{align}
%    \kappa(\x, \x') = \exp(-\eta \|\x - \x'\|^2) \label{eq:se}
%\end{align}
%where $\eta>0$ is the kernel parameter. 
Given a training dataset $\Dcal = \{(\x_1, y_1), \ldots, (\x_N, y_N)\}$, we use the regularized regression framework to estimate the target function,  $f^*(\x) = \min_{f \in \Hcal_\kappa} \sum_{n=1}^N L(y_n, f(\x_n)) + \sigma^2 \|f\|_{\Hcal_\kappa},$ where the solution can be shown to take the form $f(\x)  = \sum\nolimits_{n=1}^N \alpha_n \kappa(\x, \x_n)$ and each $\alpha_n \in \mathbb{R}$. The learning is therefore equivalent to solving an optimization problem for the coefficients $\balpha = [\alpha_1, \ldots, \alpha_N]^\top$, and $\|f\|_{\Hcal_\kappa} = \balpha^\top \K \balpha$, where $\K$ is an $N \times N$ kernel matrix, and each $[\K]_{ij} = \kappa(\x_i, \x_j)$. Denote by $\f = [f(\x_1), \ldots, f(\x_N)]^\top$  the function values at the training inputs. Since $\f = \K \balpha$, we can instead model the target function as %in terms of $\f$, 
\begin{align}
	f(\x) = \kappa(\x, \X) \K^{-1} \f, \label{eq:ker-reg-model}
\end{align}
where $\X = [\x_1, \ldots, \x_N]^\top$ and $\kappa(\x, \X) = [\kappa(\x, \x_1), \ldots, \kappa(\x, \x_N)]$. The learning amounts to minimizing the following regularized loss, 
%\begin{align}
	$\argmin_\f \sum\nolimits_{n=1}^N L(y_n, f(\x_n)) + \sigma^2 \f^{\top} \K^{-1} \f$. %\label{eq:ker-reg}
%\end{align}
When we use the squared loss, the optimum is $\f^* = \K(\K+\sigma^2\I)^{-1}\y$, and the function estimate is $f^*(\x) = \kappa(\x, \X) (\K+\sigma^2\I)^{-1}\y$. This can also be explained in a Gaussian process (GP) regression framework, where $f^*(\x)$ is the posterior mean of GP prediction~\citep{williams2006gaussian}.   
\cmt{
This gives rise to a probabilistic interpretation of the kernel regression. We can view the target function $f(\cdot)$ as sampled from a zero-mean Gaussian process (GP) with the covariance function as $\kappa$. Accordingly, $\f$ is sampled from a multivariate Gaussian distribution with covariance matrix $\K$, and the observation $\y$ from a Gaussian noise model  conditioned on $\f$, 
\begin{align}
	p(\f) = \N(\f |\0, \K), \;\;\ p(\y| \f) = \N(\y|\f, \sigma^2\I).
\end{align}
This is known as GP regression. When we estimate $\f$ by maximizing the log joint probability, the optimization is equivalent to \eqref{eq:ker-reg}. The function estimate from the kernel regression is the same as the posterior (or conditional) mean of the GP regression, \ie $\EE [p(f(\x)|\Dcal)] = \kappa(\x, \X) (\K+\sigma^2\I)^{-1}\y$. In general, for any loss $L$ in \eqref{eq:ker-reg}, we can construct a corresponding GP model with the likelihood $p(y_n | f(\x_n)) \propto \exp(-\frac{1}{\sigma^2}L(y_n, f(\x_n)))$.
}
\cmt{
\noindent \textbf{Kernel Regression.} Denote by $\Hcal_\kappa$ the Reproducing Kernel Hilbert Space (RKHS) induced by a Mercer kernel function $\kappa(\cdot, \cdot)$, and by $\|\cdot\|_\kappa$ its norm. \cmt{The functions in $\Hcal_\kappa$ takes the form $f(\cdot) = \sum_{j=1} \alpha_j \kappa(\cdot, \x_j)$ where each $\x_j$ belongs to the input domain.} Given a training dataset $\Dcal = \{(\x_1, y_1), \ldots, (\x_N, y_N)\}$, we use regularized least square regression to estimate the target function from $\Hcal_\kappa$,  $f^*(\x) = \underset{f \in \Hcal_\kappa}{\min} \sum_{n=1}^N (y_n - f(\x_n))^2 + \tau \|f\|_{\Hcal_\kappa}$, where 
\begin{align}
	f(\x) = \sum\nolimits_{n=1}^N \alpha_n \kappa(\x, \x_n). 
\end{align}
The learning is equivalent to solving an optimization problem w.r.t the coefficients $\balpha = [\alpha_1, \ldots, \alpha_N]^\top$, 
\begin{align}
	\balpha^* = \underset{\balpha}{\min} \| \y - \K \balpha\|^2 + \tau \balpha^\top \K \balpha,
\end{align}
where $\K$ is an $N \times N$ kernel matrix, and each $[\K]_{ij} = \kappa(\x_i, \x_j)$. Note that $\|f\|_{\Hcal_\kappa} = \balpha^\top \K \balpha$. Obviously  $\balpha^* = \left(\K + \tau \I\right)^{-1}\y$ where $\y=[y_1, \ldots, y_N]^\top$,  and the function estimate  has the following interpolation form, 
\begin{align}
f^*(\x) = \kappa(\x, \X) \left(\K + \tau \I\right)^{-1}\y	\label{eq:kr-est}
\end{align}
where $\X = [\x_1, \ldots, \x_N]^\top$, and $\kappa(\x, \X) = [\kappa(\x, \x_1), \ldots, \kappa(\x, \x_N)]$. Equivalently, we can  optimize the function values at $\X$, namely, $\f = [f(\x_1), \ldots, f(\x_N)]$.  From $\f = \K \balpha$, we have
\begin{align}
	\f^* \cmt{=  \underset{\f}{\min} \| \y - \f\|^2  + \tau \|f\|_{\Hcal_\kappa} }=\underset{\f}{\min} \| \y - \f\|^2 + \tau \f^\top \K^{-1} \f. \label{eq:obj-1}
\end{align}
This gives rise to a probabilistic view of kernel regression. We can view the target function $f(\cdot)$ as sampled from a zero-mean Gaussian process (GP) with the covariance function as $\kappa$, accordingly $\f$ is sampled from a multivariate Gaussian distribution, and the training outputs $\y$ is sampled from a Gaussian noise model conditioned on $\f$, 
\begin{align}
	p(\f) = \N(\f |\0, \K), \;\;\ p(\y | \f) = \N(\y|\f, \tau\I).
\end{align}
When we estimate $\f$ by maximizing the log joint probability, the optimization is equivalent to \eqref{eq:obj-1}. The kernel regression estimate \eqref{eq:kr-est} is the posterior mean of the GP regression model. 
}
\cmt{
\textbf{Gaussian processes (GPs)} are powerful nonparametric function estimators that do not assume any parametric form of the target function.
They can flexibly capture the complexity of the function according to the data, \eg from simple linear, to quadratic, to highly nonlinear, and hence they can prevent both underfitting and overfitting. 
\michael{That sounds more like philosophy/hope than something we justify.  Can we stick with more precise claims.} \shandian{I will rephrase everything in the context of kernel methods.}
Specifically, GP models assume the target function $f: \mathbb{R}^d \rightarrow \mathbb{R}$ is sampled from a  Gaussian process, $f \sim \gp(m(\cdot), \kappa(\x, \x'))$, where $m(\cdot)$ is the mean function,  and $\kappa(\cdot, \cdot)$ is the covariance function,  \ie $\cov(f(\x), f(\x')) = \kappa(\x, \x')$, which characterizes how the function values are correlated according to their inputs.  In practice, one often sets $m(\cdot)$ to the constant zero and chooses a kernel function as the covariance function. For example, a popular choice is the Square Exponential (SE) kernel, $\kappa(\x, \x') = \exp\left(-\frac{1}{\eta}\|\x - \x'\|^2\right)$ where $\eta$ is the length-scale parameter. Under the GP prior, the values of the sampled function at an arbitrary finite set of inputs, $\f = [f(\x_1), \ldots, f(\x_N)]$, follow a multi-variate Gaussian distribution, $p(\f) = \N(\f|\0, \K)$ where $[\K]_{ij} = \kappa(\x_i, \x_j)$. Suppose we have collected the training dataset $\Dcal = \{(\x_1, y_1), \ldots, (\x_N, y_N)\}$. Given $\f$, we use a noise model to fit the data while capturing the possible noises in the data. For continuous observations,  a commonly used model is the Gaussian noise model, $p(\y|\f) = \N(\y | \f, \sigma^2\I)$ where $\sigma^2$ is the noise variance and $\y = [y_1; \ldots ; y_N]$. \cmt{The marginal likelihood of data (model evidence) is therefore 
\begin{align}
p(\y|\X) = \N(\y|\0, \K + \sigma^2\I).
\end{align}
One can maximize the marginal likelihood to estimate the kernel parameters and noise variance.}  Given a test input $\x^*$, since $[\y; f(\x^*)]$ jointly follow a multi-variate Gaussian distribution, we can immediately obtain the predictive distribution as a conditional Gaussian, 	$p\big(f(\x^*)|\X, \y\big) = \N\big(f(\x^*)|\mu(\x^*), v(\x^*)\big)$,
%\begin{align}
%	p\big(f(\x^*)|\X, \y\big) = \N\big(f(\x^*)|\mu^*, v^*\big),
%	\label{eq:predictive}
%\end{align}
where $\mu(\x^*) = \k_*^\top(\K + \sigma^2\I)^{-1}\y$, $v(\x^*)=\kappa(\x^*, \x^*) - \k_*^\top(\K + \sigma^2\I)^{-1}\k_*$ and  $\k_* = [\kappa(\x^*, \x_1), \cdots,\kappa(\x^*, \x_N)]^\top$. Note that the conditional mean $\mu(\x^*)$ is also known as the kernel interpolation from the data points $\y$.
 %We can see that due to the closed-form posterior, GP models are easy and convenient to quantify and reason under uncertainty. 
}

\textbf{Bayesian spike-and-slab prior} is a powerful Bayesian sparse distribution to identify dominant patterns or signals from data~\citep{mitchell1988bayesian}. 
Take basis selection as an example for illustration. Suppose we have $M$ basis functions, $\{\phi_j(\x)\}_{j=1}^M$, and the target function is $f(\x) = \sum_{j=1}^M w_j \phi_j(\x)$. To select dominant or relevant bases, the spike-and-slab prior first samples a binary selection indicator $s_j \in \{0, 1\}$ for each basis $j$, with which to sample the basis weight $w_j$, 
\begin{align}
	p(s_j) &= \text{Bern}(s_j|\rho_0) = \rho_0^{s_j}(1 - \rho_0)^{1 - s_j}, \notag \\
    p(w_j | s_j) &= s_j\N(w_j|0, \sigma_0^2) + (1 - s_j)\delta(w_j), \label{eq:ssprior}
\end{align}
where $\delta(\cdot)$ is a Dirac-delta function. The selection indicator $s_j$ decides the prior over $w_j$. If $s_j$ is $1$, meaning that basis $j$ is selected, the weight $w_j$ is sampled from a flat Gaussian prior with variance $\sigma_0^2$, \ie the slab component. If $s_j$ is $0$,  the basis $j$ is pruned, and the weight $w_j$ is sampled from the spike prior that concentrates on zero, \ie the spike component. Via the selection indicators $\{s_j\}$, the spike-and-slab prior fulfills a selective shrinkage. For the unselected bases, the weights are directly shrunk to zero by the Dirac-delta prior $\delta(\cdot)$. For the selected ones, the weights are sampled from a flat Gaussian, which performs a mild regularization and allows the weights to be adequately estimated from data. Not only is the selective shrinkage observed empirically to be better than the uniform shrinkage in $L_1$ regularization~\citep{mohamed2012bayesian,fang2020online}, but it is critical for effective  selection in terms of risk misclassification~\citep{ishwaran2005spike}.

%To accurately and efficiently estimate $u$ from the boundary conditions and equation, we create a dense grid on the domain $\Omega$. Note that the grid is unnecessary to be evenly spaced. We can randomly sample or specially design the locations at each input dimension, and then construct the grid through a Cartesian product

%algorithm: we propose EM algorihtm. First, if we use a product kernel, like SE kernel, the covariance can be represtned by a Kronecerk product ....---> convenince in computing deriative covariance  Second, we use the EP framework. basic description, link to appendix. introduce the approximation, and then udpate, algorithm table 
\section{\MakeUppercase{Model}}
%\vspace{-0.1in}
We now present \ours, our Bayesian model for  equation discovery. Without loss of generality, we consider a 2D dynamic system to illustrate the idea. It is straightforward to extend the idea to higher-dimensional problems or reduce the idea to ODE systems. 
We denote the target (or solution) function by $u(t, x_1, x_2)$.  Given a collection of measurement data, $\Dcal = \{(\z_1, y_1), \ldots, (\z_N, y_N)\} $ where each $\z_n = (t_n, x_{n1}, x_{n2})$,  we intend to estimate $u$ and its PDE representation. To this end,  we introduce a dictionary of candidate operators $\Ocal = \{P_1, \ldots, P_A\}$, which includes  basic differentiation operators, such as $\pa{x_1} u$, $\pa{x_2} u$, $\pa{x_1x_1} u$ and $\pa{x_1x_2} u$, their linear/nonlinear combinations, and composition with other functions: $u \pa{x_1} u$,  $u$, $u^2$, $\cos(u)$, $\sin(u)$, \etc We assume the dictionary is large enough to cover all the possible operators in the ground-truth PDE. We model the PDE  with the following form, %$u_t - \sum\nolimits_{j=1}^K w_j P_j[u] = 0$, 
\begin{align}
	u_t - \sum\nolimits_{j=1}^K w_j P_j[u] = 0, \label{eq:form}
\end{align}
where $\{P_j\}_{j=1}^K \subseteq \Ocal$, and each $w_j \in \mathbb{R}$ is the weight of the operator $P_j$.

To efficiently estimate $u$, we construct a mesh $\Mcal$ to cover the domain and estimate the function values at the mesh. Note that the mesh is \textit{not} necessary to be evenly spaced. We can randomly sample or design the locations at each input dimension, and then construct the mesh via a Cartesian product. In doing so, we will not only be able to compute efficiently (see Section \ref{sec:algo} for details), but also we will be able to flexibly handle different geometries in the domain by varying the dense regions; see Appendix Fig. \ref{fig:mesh-design} for an illustration. %Note that this is consistently with the practice in numerical solvers. 
Denote by $\bgamma_j$ the locations at each input dimension $j$, and by $d_j$ the size of $\bgamma_j$.  The mesh points are $\Mcal = \bgamma_1 \times \bgamma_2 \times \bgamma_3 = \{(t', x_1', x_2') | t' \in \bgamma_1, x_1' \in \bgamma_2, x_2' \in \bgamma_3\}$.
\cmt{
\begin{align}
	&\Mcal = \bgamma_1 \times \bgamma_2 \times \bgamma_3 \notag \\
        &= \{(t', x_1', x_2') | t' \in \bgamma_1, x_1' \in \bgamma_2, x_2' \in \bgamma_3\}.
\end{align}
}

Denote the function values at $\Mcal$ by $\Ucal = \{u(t', x_1', x_2') | (t', x_1', x_2') \in \Mcal\}$. Hence, $\Ucal$ is a $d_1 \times d_2 \times d_3$ tensor.  We use kernel regression to estimate $u(\cdot)$, and, according to \eqref{eq:ker-reg-model}, the target function can be modeled as a kernel interpolation from $\Ucal$, 
\begin{align}
	u(\z) = \kappa(\z, \Mcal) \K_{\Mcal,\Mcal}^{-1} \vec(\Ucal),  \label{eq:ker-int}
\end{align}
where $\z = (t, x_1, x_2)$ is an arbitrary point in the input domain, $\vec(\cdot)$ is vectorization, $\kappa(\cdot, \cdot)$ is a kernel function, $\K_{\Mcal,\Mcal}$ is the kernel matrix on the mesh points $\Mcal$, and $\kappa(\z, \Mcal) = [\kappa(\z, \brho_1), \ldots, \kappa(\z, \brho_d)]$, where $\brho_j$ are all the points in $\Mcal$, $1 \le j \le d$ and $d = d_1 d_2 d_3$.   According to \eqref{eq:ker-int}, we can compute any derivative of $u$. Since $\Mcal$ and $\K_{\Mcal,\Mcal}^{-1}\vec(\Ucal)$ are both constant to the input $\z$, we only need to compute the derivatives of the kernel w.r.t the variables in $\z$, \eg  
\begin{align}
	\pa{x_1x_2} u(\z) = \pa{x_1x_2}\kappa(\z, \Mcal) \cdot  \K_{\Mcal,\Mcal}^{-1} \vec(\Ucal). \label{eq:ker-int-der}
\end{align}	
Here $\pa{x_1x_2}\kappa(\z, \Mcal) \overset{\Delta}{=} [\pa{x_1x_2} \kappa(\z, \brho_1), \ldots, \pa{x_1x_2}\kappa(\z, \brho_d)]$, and all $\{\brho_j\}_{j=1}^d$ constitute $\Mcal$. In this way, given the current function estimate \eqref{eq:ker-int}\cmt{(based on $\Ucal$ and the kernel parameters)}, we can evaluate each operator in $\Ocal$ on mesh $\Mcal$. We denote them by $\widehat{\Pcal} = \{\Pcal_1, \ldots, \Pcal_A\}$ where each is a $d_1 \times d_2 \times d_3$ tensor. 

Based on  $\widehat{\Pcal}$,  we propose a Bayesian sparse model to identify the operators in the PDE and to estimate the operator weights. Specifically, for each operator $j$,  we use a spike-and-slab prior to sample a selection indicator $s_j$ and operator weight $w_j$ as in \eqref{eq:ssprior}.\cmt{
\begin{align}
    p(s_j) &= \text{Bern}(s_j|\rho_0) = \rho_0^{s_j}(1 - \rho_0)^{1 - s_j}, \notag \\
  p(w_j | s_j) &= s_j\N(w_j|0, \sigma_0^2) + (1 - s_j)\delta(w_j). \label{eq:ss-2}
\end{align} 
}

Then, conditioned on $\w = [w_1, \ldots, w_A]^\top$\cmt{, $\s = [s_1, \ldots, s_A]^\top$,} and $\widehat{\Pcal}$, we sample a virtual dataset $\wy = \0$, as
\begin{align}
	p(\wy|\bPhi, \w) = \N\left(\wy|\h - \bPhi \w, \tau \I\right) = \N\left(\h|\bPhi \w, \tau \I\right), 
 \label{eq:vll}
\end{align}
where $\bPhi = [\vec(\Pcal_1), \ldots, \vec(\Pcal_A)]$,   $\h$ is $\frac{\partial u}{\partial t}$ evaluated at $\Mcal$ and has been flattened into a vector. The equation likelihood \eqref{eq:vll} measures how consistent the selected equation (by the spike-and-slab prior) is with the estimated target function at the mesh $\Mcal$. 
Finally, we use a Gaussian noise model  to fit the measurement data, 
\begin{align}
	p(\y_{\text{tr}}|\Ucal) = \N(\y_{\text{tr}} | \u_{\text{tr}}, v\I), \label{eq:data-ll}
\end{align}
where $\u_{\text{tr}}$ is obtained via the kernel interpolation \eqref{eq:ker-int} at the training inputs and $\y_{\text{tr}} = [y_1, \ldots, y_N]^\top$ are the training outputs. 

Note that a particular advantage of using the spike-and-slab prior \eqref{eq:ssprior} is that we can estimate the posterior of each selection indicator $p(s_j|\Dcal, \wy)$, and use it to decide if an operator should be selected, \eg checking if $p(s_j=1|\Dcal, \wy)>0.5$. We never need to set a threshold over the weight values, \eg $|w_j|>10^{-3}$, as popular methods (like SINDy and PINN-SR) do. Tuning a weight threshold can be troublesome and inconvenient in practice. A bigger threshold can miss important operators with small weights, \eg in Burger's equation with a small viscosity; a smaller one, however, can easily select false operators. Moreover, the threshold is often not general. When switching to a different problem, one has to carefully tune it from scratch.  

\cmt{
 We then sample a virtual dataset $\wy = \0$, given the candidate operators' evaluation results  on $\Zcal$, 
\begin{align}
	p(\wy|\bPhi, \w) = \N\left(\wy|\h - \bPhi \w, \tau_2^{-1} \I\right) , 
 \label{eq:vll}
\end{align}
where $\bPhi = [\bphi_1, \ldots, \bphi_M]$, $\w = [w_1, \ldots, w_M]^\top$, and $\h$ is $\frac{\partial u}{\partial t}$ evaluated at $\Zcal$. Note that each $\bphi_j$ ($1\le j \le M$) and $\h$ have been flattened into a vector. The equation likelihood \eqref{eq:vll} measures how consistent the selected equation is with the estimate of $\Ucal$ and the operators. During the learning,  the data likelihood~\eqref{eq:gp-ll} and equation likelihood~\eqref{eq:vll} will synergize to learn $\Ucal$ and the underlying equation.  

that couples GPs and the horse-shoe (HS) prior to select  the operators in the PDE and to estimate their spatially varying coefficients. 
}

\cmt{
To discover underlying governing equations from scarce, noisy data, we propose a Bayesian equation discovery model based Gaussian processes.  
Without loss of generality, we consider a spatial-temporal system. 
\michael{As opposed to what?} \shandian{this is just to follow the description of other equation discovery papers. They illustrate them method with a spatial temporal system. I will add one sentence, saying the connection to ODE and spatial system discovery.}
We denote  the response (or solution) function by $u(t,x_1, \ldots, x_m)$. 
Note that when $m=0$, it reduces to an ODE system. Given a small collection of measurement data, $\Dcal = \{(\x_1, y_1), \ldots, (\x_P, y_P)\} $, we aim to learn $u$ and the governing equation about $u$. 
To this end, we first prepare a dictionary of candidate operators $\Fcal = \{\psi_1, \ldots, \psi_M\}$, which consists of basic differentiation operators, such as $D_0 [u]$, $D_{1} [u]$, and $D_{22} [u]$, their linear/nonlinear combinations, and composition with other functions: $u D_0 [u]$, $u D_1 [u]$,  $u$, $u^2$, $u^3$, $\cos(u)$, $\sin(u)$, \etc 
\michael{Something about how we prepare that dictionary.} \shandian{Yes, i will add one sentence.}
Here $D_\a [u]$ means the partial derivatives of $u$ w.r.t the input variable(s) indexed by $\a$, \eg  $D_0[u]  = \frac{\partial u}{\partial t}$, $D_{1}[u]  = \frac{\partial u}{\partial x_1}$, and $D_{122}[u] =\frac{\partial^3 u}{\partial x_1 \partial x_2^2}$. 
We assume the governing equation can be expressed as a linear combination of  a small set of operators from $\Fcal$,
%To discover the underlying governing equation from scarce, noisy data, we propose a Gaussian process (GP) based model. Without loss of generality, we consider a 1d spatial-temporal system to explain the idea.  We denote  the response (or solution) function by $u(t,x)$. Given a small collection of measurement data, $\Dcal = \{((t_1, x_1), y_1), \ldots, ((t_N, x_N), y_N)\} $, we aim to learn $u$ and the governing equation about $u$. To this end, similar to~\citep{brunton2016discovering,chen2021physics}, we first prepare a dictionary of candidate operators $\Fcal = \{\psi_1, \ldots, \psi_M\}$, which consists of basic differentiation operators, such as $u_x$, $u_t$, and $u_{xx}$, their linear/nonlinear combinations, and composition with other functions: $u u_t$, $u u_x$,  $u$, $u^2$, $u^3$, $\cos(u)$, $\sin(u)$, \etc We assume the governing equation can be represented by a linear combination of  a small set of operators from the dictionary,
\begin{align}
	\frac{\partial u}{\partial t} - \sum\nolimits_{j=1}^M w_j \psi_j[u] = 0 ,
 \label{eq: form}
\end{align}
where most of the coefficients $\{w_j\}$ are zero. 
%this sentence can be removed
Note that the linear structure does not restrict our flexibility of finding nonlinear equations, because the dictionary can include nonlinear operators. 
The framework can account for unknown source terms by adding the corresponding candidate functions into $\Fcal$ to explain the source.

To estimate the target function $u$ and perform operator selection in \eqref{eq: form}, we create a dense grid in the domain of interest, and we consider estimating the function values at the grid. 
\michael{This grid thing is coming out of the blue.  Can we define the setup somewhere, so it's clear there is a grid.  Also, maybe write out a PDE, saying that it is unknown but we seek to discover it, in general or when we assume some form for it.  Perhaps here, but this probably goes in the intro.} \shandian{thanks for the comments. I will make the connection to the practice of numerical solvers and to our motivation for efficient computation. I will also emphasize this setting can also adapt to complex geometries, like the figures you show. }
Denote the inputs  at each dimension by $\z_1, \ldots, \z_{m+1}$, where each $\z_j$ includes $N_j$ inputs. The grid points are 
\begin{align}
	\Zcal = \z_1 \times \cdots \times \z_{m+1} = \{(z_1, \ldots, z_{m+1}) | z_j \in \z_j, 1 \le j \le m+1\} .
\end{align}
Denote by $\Ucal$ the values of $u(\cdot)$ at the grid points $\Zcal$. Hence $\Ucal$ is a $m+1$ dimensional tensor, of size $N_1 \times \ldots \times N_{m+1}$.  We place a GP prior over $u$, and $\Ucal$ therefore follows a multi-variate Gaussian prior distribution, 
\begin{align}
	p(\Ucal) = \N(\vec(\Ucal)| \0, \K_{\Zcal\Zcal}) ,\label{eq:prior-u}
\end{align}
where $\K_{\Zcal\Zcal}$ is the kernel matrix computed from $\Zcal$.

Given $\Ucal$, we first obtain the function values at the training inputs $\Xcal = \{\x_1, \ldots \x_P\}$ via kernel interpolation, 
\begin{align}
	f_n \overset{\Delta}{=} u(\x_n) = \kappa(\x_n, \Zcal) \K_{\Zcal\Zcal}^{-1} \vec(\Ucal), \;\;\; 1 \le n \le P  ,
 \label{eq:kint-1}
\end{align}
with which we sample the observed outputs from a Gaussian noise model, %$p(\y|\u) = \N(\y|\u, \sigma_0^2I)$
\begin{align}
	p(\y|\f) = \N(\y|\f, \tau_1^{-1}\I ) ,
 \label{eq:gp-ll}
\end{align}
where $\f = [f_1, \ldots f_P]^\top$, and $\tau_1$ is the noise inverse variance. 

Next, 
\michael{Can we write this to be clearer, as to what is the high-level organization. Is this ``Next'' following the ``first'' in the ``Given $\Ucal$, we first obtain ...'' sentence? } \shandian{ok, i will add one leading sentence to summarize what we are doing in each step.}
to select the operators from $\Fcal$ and to identify the equation, we obtain all the needed derivatives values of $u$ on $\Zcal$, so as to obtain the evaluation of every candidate operator on $\Zcal$. 
For example, suppose one operator is $\psi [u] =  D_{jk}[u] u$. We need to obtain the values of $D_{jk}[u]$ on $\Zcal$ and then obtain the values of  $\psi[u]$ on $\Zcal$. 
This is can be done via kernel differentiation and interpolation. 
\michael{I'm trying to figure out what is ``new'' here.  Is this the first place there is something ``new''?} \shandian{Yes, using kernel interpolation/differentiation for eq discovery is new here. I will highlight the benefit as compared with using NN network. Because it will always gives a linear combination of the function estimate. The nolinear part only comes into the kenels. It will make the estimation of the function values eaiser.}
Specifically, consider arbitrary inputs $\balpha$ and $\balpha'$ in the domain. 
Suppose we want to obtain $D_{jk} [u(\balpha)]$ given $\Ucal$. Since $u(\cdot)$ is a GP, its derivative $D_{jk}[u]$ is a also a GP, and the covariance/kernel between $D_{jk} u(\balpha)$ and $u(\balpha')$ is derived from the kernel differentiation at the same input variables with respect to which we take the derivative, 
\begin{align}
	\cov(D_{jk} [u(\balpha)], u(\balpha')) = D_{jk} [\kappa(\balpha, \balpha')]  = D_{jk} [ \kappa(\alpha_1, \ldots \alpha_{m+1}, \alpha'_1, \ldots, \alpha'_{m+1})]. \label{eq:ker-diff}
\end{align}
\cmt{Note that we expand the input vectors to the kernel function $\kappa$ to be more clear.} 
Therefore, we can use the new kernel to obtain the interpolation for $D_{jk}[u(\balpha)]$ from $\Ucal$, 
\begin{align}
	D_{jk}[u(\balpha)] = D_{jk} [\kappa(\balpha, \Zcal)] \K_{\Zcal\Zcal}^{-1} \vec(\Ucal) ,
 \label{eq:kint-2}
\end{align}
where $D_{jk}[\kappa(\balpha, \Zcal)]  = \left[D_{jk}[\kappa(\balpha, \balpha_1')], \ldots, D_{jk}[\kappa(\balpha, \balpha_N')]\right]$, all $\balpha'_n \in \Zcal$ and  $N = \prod_{j=1}^{m+1} N_j$. 
%$D_{jk}[\kappa(\z, \Zcal)] = \{D_{jk}[\kappa(\z, \z')]|\z' \in \Zcal\}$.

 Let us denote the evaluated values of each candidate operator on $\Zcal$  by $\bphi_1, \ldots, \bphi_M$. To identify the operators in the governing equation, for each $\bphi_j$, we use a spike-and-slab prior to sample a selection indicator $s_j$ and the operator coefficient $w_j$, 
 \begin{align}
 	p(s_j) = \text{Bern}(s_j|\rho_0) = \rho_0^{s_j}(1 - \rho_0)^{1 - s_j}, \;\;p(w_j | s_j) = s_j\N(w_j|0, \sigma_0^2) + (1 - s_j)\delta(w_j), \label{eq:ss-2}
 \end{align}
 We then sample a virtual dataset $\wy = \0$, given the candidate operators' evaluation results  on $\Zcal$, 
\begin{align}
	p(\wy|\bPhi, \w) = \N\left(\wy|\h - \bPhi \w, \tau_2^{-1} \I\right) , 
 \label{eq:vll}
\end{align}
where $\bPhi = [\bphi_1, \ldots, \bphi_M]$, $\w = [w_1, \ldots, w_M]^\top$, and $\h$ is $\frac{\partial u}{\partial t}$ evaluated at $\Zcal$. Note that each $\bphi_j$ ($1\le j \le M$) and $\h$ have been flattened into a vector. The equation likelihood \eqref{eq:vll} measures how consistent the selected equation is with the estimate of $\Ucal$ and the operators. During the learning,  the data likelihood~\eqref{eq:gp-ll} and equation likelihood~\eqref{eq:vll} will synergize to learn $\Ucal$ and the underlying equation.    

%The supervised signal comes from the data likelihood  \eqref{eq:gp-ll}, and propagate through the kernel interpolation \eqref{eq:kint-1} and \eqref{eq:kint-2}. The equation likelihood \eqref{eq:vll} combined with the spike-and-slab prior \eqref{eq:ss-2} then identify the operators and their coefficients. During the learning, the data likelihood and equation likelihood will synergize for the estimation of $\Ucal$ and the equation identification.  
%evaluations $\{\bphi_j\}$, computed based on $\g$, and $\g$ are coupled with $\u$ in the GP prior \eqref{eq:gp-joint}. Via fitting the data by \eqref{eq:gp-ll}, the supervised information is propagated to $\g$ and then each $\bphi_j$. 
The joint distribution of our model is given by
\begin{align}
	%p(\u, \g, \y, \wy, \w) =  \N\left([\u; \g]|\0, \bSigma\right) \prod\nolimits_{j} p_{\text{HS}}(w_j)  \N(\y|\f, \sigma_0^2\I) \N\left(\wy|\brho - \sum\nolimits_{j} w_j  \bphi_j, \sigma_1^2\I\right).
	p(\Ucal, \f, \y, \w, \s, \wy) &= \N(\vec(\Ucal)| \0, \K_{\Zcal\Zcal})\cdot \prod\nolimits_{j=1}^M  \text{Bern}(s_j|\rho_0) \left( s_j\N(w_j|0, \sigma_0^2) + (1 - s_j)\delta(w_j)\right) \notag \\
	&\cdot \N(\y|\f, \tau_1^{-1}\I) \N\left(\wy|\h - \bPhi \w, \tau_2^{-1} \I\right) , 
 \label{eq:joint-2}
\end{align} 
 where $\s = [s_1, \ldots, s_M]^\top$. 
}

\michael{Would you make a pass and make the entire discussion more organized.  It seems very low level, with ``and wait, there's more ...'' so it's hard to see what is new with us.  Also, we will need to weave in some comments about hierarchical and complex geometries, like we discussed last week.} \shandian{Yes, will do}

\section{\MakeUppercase{Algorithm}}
\label{sec:algo}
Given the measurement data $\Dcal$ and the dictionary $\Ocal$, the learning of our model amounts to estimating $\Ucal$, the kernel parameters, and the posterior distribution of $\s$ and $\w$.  This is challenging. First, when the number of mesh points is large, the kernel interpolation that requires the inverse of $\K_{\Mcal,\Mcal}$ can be extremely costly or even infeasible; see \eqref{eq:ker-int} and \eqref{eq:ker-int-der}.\cmt{ because the time complexity is cubic to the number of mesh points.} Second, 
 because the spike-and-slab prior \eqref{eq:ssprior} mixes binary and continuous random variables, the posterior distribution is analytically intractable. 
 To address these challenges, we induce a Kronecker product in kernel computation, and use tensor algebra to avoid operating on full matrices. We then develop an expectation-propagation expectation-maximization (EP-EM) algorithm. In each iteration, our algorithm performs two steps. In the E step, we fix $\Ucal$ and the kernel parameters, and use EP~\citep{minka2001expectation} to estimate the posterior of $\s$ and $\w$. In the M step, we maximize the expected model likelihood to update $\Ucal$ and the kernel parameters. The alternating of the E and M steps keeps mutually improving equation discovery and function estimation until convergence. See Algorithm \ref{algo:ep-em} for a summary.

%computation trick --> E step --> M step --> alternate 
Specifically, thanks to the usage of a mesh for placing the function values to be estimated, $\Ucal$ is a $d_1 \times d_2 \times d_3$ tensor, and if we use a product-kernel, $\kappa(\z, \z') = \kappa(t, t')\kappa(x_1, x_1)\kappa(x_2, x_2)$, 
%since $\Ucal$ is placed on the mesh $\Mcal$, \ie  $\Ucal$ is a $d_1 \times d_2 \times d_3$ tensor, if we use a product-kernel, like the popular SE kernel in \eqref{eq:se}, $\kappa(\z, \z') = \kappa(t, t')\kappa(x_1, x_1)\kappa(x_2, x_2)$  
%\begin{align}
%	\kappa(\z, \z') = \kappa(t, t')\kappa(x_1, x_1)\kappa(x_2, x_2)
%\end{align}
where $\z = (t, x_1, x_2)$ and $\z' = (t', x_1', x_2')$, we can immediately induce a Kronekcer product structure in the kernel matrix, %$\K_{\Mcal,\Mcal}$, 
\begin{align}
	\K_{\Mcal,\Mcal} = \K_1 \kron \K_2 \kron \K_3 ,
\end{align}
where $\K_1= \kappa(\bgamma_1, \bgamma_1)$, $\K_2= \kappa(\bgamma_2, \bgamma_2)$, and $\K_3= \kappa(\bgamma_3, \bgamma_3)$ are the kernel matrices of the inputs at each dimension of $\Mcal$. Note that the popular square exponential (SE) kernel is a product kernel. One can also construct a product kernel from any other kernels. 

We can then leverage the nice properties of Kronecker product~\citep{minka2000old} and tensor algebra~\citep{kolda2006multilinear} to avoid computing the full kernel matrix and tremendously improve the efficiency. First, to compute \eqref{eq:ker-int}, we can derive that $u(\z)=\kappa(t, \bgamma_1) \kron \kappa(x_1, \bgamma_2)  \kron \kappa(x_2, \bgamma_3) \left( \K_1 \kron \K_2 \kron \K_3\right)^{-1}  \vec(\Ucal)  = \left( \kappa(t, \bgamma_1) \K_1^{-1} \kron \kappa(x_1, \bgamma_2)\K_2^{-1} \kron \kappa(x_2, \bgamma_3)\K_3^{-1}\right) \vec(\Ucal) = \Ucal \times_1 \kappa(t, \bgamma_1) \K_1^{-1}  \times_2 \kappa(x_1, \bgamma_2) \K_2^{-1} \times_3 \kappa(x_2, \bgamma_3)\K_3^{-1}$,
\cmt{
\begin{align}
	& \kappa(t, \bgamma_1) \kron \kappa(x_1, \bgamma_2)  \kron \kappa(x_2, \bgamma_2) \left( \K_1 \kron \K_2 \kron \K_3\right)^{-1}  \vec(\Ucal)  \notag \\
	&= \left( \kappa(t, \bgamma_1) \K_1^{-1} \kron \kappa(x_1, \bgamma_2)\K_2^{-1} \kron \kappa(\x_2, \bgamma_3)\K_3^{-1}\right) \vec(\Ucal) \notag \\
	&= \Ucal \times_1 \kappa(t, \bgamma_1) \K_1^{-1}  \times_2 \kappa(x_1, \bgamma_2) \K_2^{-1} \times_3 \kappa(\x_2, \bgamma_3)\K_3^{-1}, \notag %\label{eq:kint-kron} 
\end{align}
}
where $\times_k$ is the tensor-matrix product at mode $k$~\citep{kolda2006multilinear}. Hence, we only need to compute the inverse of each $\K_j$ ($1 \le j \le 3$), which takes time complexity $O(d_1^3 + d_2^3 + d_3^3)$,  and the tensor-matrix product takes $O(d)$ time complexity ($d=d_1d_2d_3$) ---  linear in the size of $\Ucal$. Hence, it is much more efficient than the naive computation --- $O(d^3)$ time complexity. 

Next, to compute a derivative of $u$ at $\Mcal$, we find that, because \textit{the product kernel is decomposed over individual input variables, the corresponding kernel differentiation still maintains the product structure.} For example, to obtain $\pa{x_1x_2} u$, we have the corresponding kernel derivative (see \eqref{eq:ker-int-der}) as 
\begin{align}
    &\pa{x_1x_2}\kappa(\z, \brho_j) = \pa{x_1x_2}\left[\kappa(t, \rho_{j1})\kappa(x_1, \rho_{j2})\kappa(x_2, \rho_{j3})\right] \notag \\
    &= \kappa(t, \rho_{j1})\cdot \pa{x_1}\kappa(x_1, \rho_{j2})\cdot \pa{x_2}\kappa(x_2, \rho_{j3}).
\end{align}
%$\pa{x_1x_2}\kappa(\z, \brho_j) = \pa{x_1x_2}\left[\kappa(t, \rho_{j1})\kappa(x_1, \rho_{j2})\kappa(x_2, \rho_{j3})\right] = \kappa(t, \rho_{j1})\cdot \pa{x_1}\kappa(x_1, \rho_{j2})\cdot \pa{x_2}\kappa(x_2, \rho_{j3})$.
Accordingly, to compute $\pa{x_1x_2} \Ucal  \overset{\Delta}{=} \{\pa{x_1x_2}u(\z)|\z \in \Mcal\}$, we have 
%$\vec(\pa{x_1x_2} \Ucal) = \left(\K_1 \kron D_1 [\K_2] \kron D_1[\K_3] \right)\left(\K_1 \kron \K_2 \kron \K_3\right)^{-1} \vec(\Ucal)=\left(\I \kron D_1[\K_2] \K_2^{-1} \kron D_1[\K_3] \K_3^{-1}\right) \vec(\Ucal)$, and hence $\pa{x_1x_2} \Ucal=\Ucal \times_2 D_1[\K_2]\K_2^{-1}\times_3 D_1[\K_3] \K_3^{-1}$,
\begin{align}
	&\vec(\pa{x_1x_2} \Ucal)\notag \\
    &= \left(\K_1 \kron D_1 [\K_2] \kron D_1[\K_3] \right)\left(\K_1 \kron \K_2 \kron \K_3\right)^{-1} \vec(\Ucal) \notag \\
	&=\left(\I \kron D_1[\K_2] \K_2^{-1} \kron D_1[\K_3] \K_3^{-1}\right) \vec(\Ucal), \notag %\\
%        &=\Ucal \times_2 D_1[\K_2]\K_2^{-1}   \times_3 D_1[\K_3] \K_3^{-1},
\end{align}
and hence 
\begin{align}
    \pa{x_1x_2} \Ucal=\Ucal \times_2 D_1[\K_2]\K_2^{-1}\times_3 D_1[\K_3] \K_3^{-1}, 
\end{align}
where $D_1[\cdot ]$ means taking the partial derivative w.r.t the first input variable for each kernel function inside, $D_1 [\K_2] = [\pa{\gamma} \kappa(\gamma, \gamma')]_{\gamma, \gamma' \in \bgamma_2}$, and $D_1 [\K_3] = [\pa{\gamma} \kappa(\gamma, \gamma')]_{\gamma, \gamma' \in \bgamma_3}$.  The multilinear operation takes time complexity $O((d_2 + d_3)d)$. Hence, the overall time complexity is $O(d_1^3+ d_2^3 + d_3^3 + (d_2 + d_3)d)$ which again is much more efficient than the naive computation with  $O(d^3)$ complexity. As such, we can compute the values of all the required derivatives on $\Mcal$ highly efficiently.

\cmt{

We now present the model estimation algorithm. Given the training data $\Dcal$, we aim to estimate the function values at the grid $\Zcal$, \ie $\Ucal$, the kernel parameters $\btheta$, and the posterior distribution of the selection indicators $\s$ and the operator coefficients $\w$. 
This is challenging, because the spike-and-slab prior in \eqref{eq:ss-2} mixes binary and continuous random variables, making the posterior distribution analytically intractable. 
In addition, when the grid is dense, \ie the number of grid points  is large, computing the covariance matrix $\K_{\Zcal\Zcal}$ in the prior (see \eqref{eq:prior-u}) and kernel interpolation (see \eqref{eq:kint-1} and \eqref{eq:kint-2}) is prohibitively expensive. 
\michael{What is a ``dense grid'' versus a ``sparse grid'' in this context?} \shandian{dense grid means that we estimate function values at many input locations. Doing this will make our estimate of the function more accurate and fine-grained. In practice, we want dense grid. sparse grid will lose subtle local information of the target function. Like in numerical methods, it is not favored to use a sparse mesh, because it will lower the accuracy.}
To address these issues, we introduce a factorized approximation to the spike-and-slab prior for efficient posterior estimation.
\michael{I'm not familiar with the spike-and-slab prior, so we should explain it so the reasonable reader will know the context.  Here it sounds like one of our contributions is to couple that prior with the Kronocker construction, exploiting the underlying geometry.}
\michael{Also, if the ``factorized approximation'' approximation is the Kronocker thing that comes from the geometry, then we are presenting it here as part of the algorithm, but I think it may be better if we present it as part of the model, i.e., in the previous section, and they say that the algorithm exploits that. This should be appropriate for some problems and not others.}\shandian{The factorized approximation is to estimate the posterior distribution of the selection indicators and operator coefficients. So it is solely in the spike-and-slab inference by EP. Kronecker product is used to compute the kernel related things, it was int he M step.  }
We leverage the grid structure in $\Ucal$, and use a product kernel to evade computing the full covariance matrix. 
\michael{Do PDE people say ``leverage the grid structure''?  So far as I know, ML people haven't done that.} \shandian{I will be more explicit, i will say leverage the grid structure to derive a Kronecker product. }
Bases on these methods, we develop an efficient expectation-propagation expectation-maximization (EP-EM) algorithm. 
In each iteration, our algorithm performs two steps. 
In the E step, we fix $\Ucal$ and $\btheta$, and use the expectation propagation (EP) framework~\citep{minka2001expectation} to estimate the posterior of $\s$ and $\w$. 
In the M step, we maximize the expected log joint probability to update $\Ucal$ and $\btheta$. 
The alternating of the E and M steps keeps mutually improving the function estimate and equation discovery until convergence. 
}

\textbf{E Step.} With this efficient computational method, we perform EP-EM steps for model estimation. In the E step, we estimate $p(\s, \w|\Dcal, \wy)$ given the current estimate of $\Ucal$. The joint distribution is 
\begin{align}
    &p(\w, \s, \Dcal, \wy) \propto  \N(\h|\bPhi \w, \tau\I) \cdot \label{eq:ep-joint-dist} \\
    &\hspace{1mm}\prod\nolimits_{j=1}^A \text{Bern}(s_j|\rho_0)\left(\s_j \N(w_j|0, \sigma^2_0) + (1-s_m)\delta(w_j)\right). \notag 
\end{align}
Note that we drop the constant terms to $\w$ and $\s$, such as the measurement data likelihood~\eqref{eq:data-ll}. We can see that the problem is reduced to the inference for Bayesian linear regression with the spike-and-slab prior. We therefore use a similar approach to~\citep{fang2020online} to develop an efficient EP method.  EP approximates each non-exponential-family factor in the joint distribution with an exponential-family term. Then from the approximate joint distribution, we can obtain a closed-form posterior, since the exponential family is closed under multiplication. EP iteratively updates each factor approximation until convergence. We leave the details in Appendix Section~\ref{sect:ep-ss}.   
After our EP inference, we obtain a posterior approximation, 
\begin{align}
	& \hspace{-2mm} p(\s, \w|\Dcal, \wy) \notag \\
 &\approx q(\s, \w) = \prod\nolimits_{j=1}^A     \text{Bern}(s_j|\sigma(\widehat{\rho}_j)) \cdot \N(\w|\bbeta, \bSigma), \label{eq:qws}
\end{align}
where $\sigma(\cdot)$ is the sigmoid function. We then prune the operators according to $q(\s)$ to decide the currently discovered equation.  

\textbf{M Step.}
In the M step, we maximize the expected log model likelihood under $q(\w)$ over the remaining operators to update $\Ucal$ and the kernel parameters, 
\begin{align}
	\Lcal =&  - \frac{1}{2}\vec(\Ucal)^\top \K_{\Mcal\Mcal}^{-1} \vec(\Ucal) - \frac{1}{2v} \| \y - \f\|^2 \notag \\
    &- \frac{1}{2\tau}\EE_q\left[ \left|\h - \widehat{\bPhi} \widehat{\w}\right\|^2\right] + \text{const}, \label{eq:expt-ll}
\end{align}
where $\widehat{\bPhi}$ and $\widehat{\w}$ are the evaluation and weights of the remaining operators, respectively. Note that the first term is the RKHS norm of the target function estimate \eqref{eq:ker-int}, which regularizes the learning of $u(\cdot)$. With the Kronecker product and tensor algebra, the computation of the RKHS norm is highly efficient. We can use any gradient based optimization to maximize $\Lcal$.

\textbf{Algorithm Complexity.} The time complexity of our model estimation algorithm is $O(\sum_{j=1}^3d_j^3  + (N + \sum_j d_j)d )$, where $d=d_1d_2d_3$. The space complexity is $O( \sum_{j=1}^3 d_j^2 + d +A^2)$, for the storage of the kernel matrices at each dimension, $\Ucal$ and $q(\s, \w)$. 

\begin{algorithm}[t]
	\small 
	\caption{\ours($\Dcal$, $\Ocal$\cmt{, $\sigma_0^2$, $\tau$, $T$, $\xi$}) } 
	\begin{algorithmic}[1]                    % enter the algorithmic environment
		\STATE Learn a kernel regression model from $\Dcal$ to obtain an initial estimate of $\Ucal$ and the kernel parameters $\btheta$.%; obtain the likelihood variance $v$ in \eqref{eq:data-ll} with cross-validation. 
		\REPEAT
		\STATE E-step: Fix $\Ucal$ and $\btheta$, run EP to estimate $q(\s, \w)$ in \eqref{eq:qws}. Prune the operators with $q(s_j=1)<\alpha$.   
		\STATE M-step: Fix $q(\s, \w)$, run ADAM for 100 steps to maximize \eqref{eq:expt-ll}  to update $\Ucal$ and $\btheta$.  
		\UNTIL {the maximum iteration number  is reached or the relative change of $\Ucal$ is less than a tolerance level.}
		\RETURN $q(\s, \w)$, $\Ucal$ and the kernel parameters $\btheta$. 
	\end{algorithmic}\label{algo:ep-em}
	\vspace{-0.05in}
\end{algorithm}

\cmt{
Specifically, to compute each $f_j$ in \eqref{eq:kint-1}, we have 
\begin{align}
	f_n &= \left(\kappa(x_{n,1}, \z_1) \kron \ldots \kron \kappa(x_{n,{m+1}},\z_{m+1})\right) (\K_1^{-1} \kron \ldots \kron \K_{m+1}^{-1})\vec(\Ucal) \notag \\
	&= \left(\kappa(x_{n,1}, \z_1) \K_1^{-1} \kron \ldots \kron \kappa(x_{n,m+1},\z_{m+1})\K_{m+1}^{-1}\right) \vec(\Ucal) \notag \\
	&= \Ucal \times_1 \kappa(x_{n,1}, \z_1) \K_1^{-1} \times_2 \ldots \times_{m+1} \kappa(x_{n,m+1},\z_{m+1})\K_{m+1}^{-1}
\end{align}
where $\times_j$ is tensor-matrix product at mode $j$. Therefore, we can compute $\kappa(x_{n,j}, \z_j) \K_j^{-1}$ at each dimension $j$, and then use the tensor-matrix product to compute the $f_n$. 
}

\cmt{
According to the properties of the Kronecker product, we have 
\begin{align}
	\log |\K_{\Zcal\Zcal}| &= \sum\nolimits_{j} (N/N_j)\log|\K_j|, \notag \\
	\K_{\Zcal\Zcal}^{-1} \vec(\Ucal) &= 
\end{align}
where $N = \prod_j N_j$
}

%We now present the model estimation algorithm. Given the training data $\Dcal$ and collocation points $\Zcal$, we aim to estimate the posterior distribution of $\u$, $\g$, the coefficients $\w$,  and the kernel parameters. This is challenging, because the HS density in \eqref{eq:hs} does not have a closed-form. In addition, when the number of collocation points is large, the covariance matrix in the multivariate Gaussian prior \eqref{eq:gp-joint} is prohibitively expensive to compute. To address these issues, we introduce  auxiliary variables to represent the HS prior,  use the variational inducing-point method to evade computing the huge covariance matrix, and  develop an efficient structural variational inference algorithm.% with the re-parameterization trick. 

\cmt{
Specifically, we first introduce auxiliary variables to sample each $\lambda_j^2$ and $s^2$ in the HS prior \eqref{eq:hs} via a nested inverse Gamma (NIG) structure~\citep{wand2011mean}, which is equivalent to the half-Cauchy prior of $\lambda_j$ and $s$, 
\begin{align}
	\lambda_j^2 | v_j  \sim \text{IG}(\frac{1}{2}, \frac{1}{v_j}), \;\; v_j \sim \text{IG}( \frac{1}{2}, \frac{1}{a_0}), \notag \\
	s^2 | \alpha \sim \text{IG}(\frac{1}{2}, \frac{1}{\alpha}), \;\; \alpha \sim \text{IG}(\frac{1}{2}, \frac{1}{a_1}), \label{eq:nig}
\end{align}
where $\text{IG}(x|\beta_0,\beta_1) = \frac{\beta_1^{\beta_0}}{\Gamma(\beta_0)}x^{-\beta_0 - 1}\exp(-\frac{\beta_1}{x})$ and $\Gamma(\cdot)$ is the Gamma function.
Similar to the whitening method often used in GP inference~\citep{murray2010slice}, we represent $w_j = \lambda_j s \xi_j$ according to \eqref{eq:hs} where $\xi_j \sim \N(0, 1)$, so as to reduce the prior dependency and ease the posterior inference. 
 Therefore, the joint distribution of our model with the auxiliary variables is 
\begin{align}
	&p(\u, \g, s^2, \alpha, \{\lambda_j^2, v_j\}_{j=1}^M) =  \N\left([\u; \g]|\0, \bSigma\right) \text{NIG}(s^2, \alpha|a_1)  \notag \\
	& \cdot \prod\nolimits_{j=1}^M \text{NIG}(\lambda_j^2, v_j|a_0) \N(\xi_j|0, 1) \cdot \N(\y|\f, \sigma_0^2\I) \N\left(\wy|\brho - \sum\nolimits_{j} w_j \bphi_j, \sigma_1^2\I\right), \label{eq:joint-2}
\end{align}
where $\text{NIG}(\eta, \gamma|a) = \text{IG}(\eta|\frac{1}{2}, \frac{1}{\gamma})\text{IG}(\gamma|\frac{1}{2}, \frac{1}{a}) $.  

We then consider the variational inference approach for posterior estimation~\citep{wainwright2008graphical}. Given a variational posterior $q(\Theta)$, we construct a variational model evidence lower bound (ELBO), 
\begin{align}
	\Lcal = \EE_{q}\left[\frac{p(\u, \g, s^2, \alpha, \{\lambda_j^2, v_j\}_{j=1}^M)}{q(\Theta)}\right] \label{eq:elbo}
\end{align}
where $\Theta = \{\u, \g, s^2, \alpha, \{\lambda_j^2, v_j\}\}$ includes all the latent random variables. We maximize $\Lcal$ to estimate $q(\Theta)$. For efficient computation, we design that $q(\Theta)$ is factorized across the variables.  Specifically, similar to \citep{ghosh2019model}, we introduce a log normal posterior (LN) for $s^2$ to ensure  positivity, which is also convenient to apply the reparameterization trick~\citep{kingma2013auto} for efficient stochastic optimization. Since  each $w_j = s \lambda_j \xi_j$,  the local scales $\{\lambda_j\}$ and local Gaussian noises $\{\xi_j\}$  can be highly correlated in the true posterior. To capture their dependency, we introduce a matrix Gaussian variational posterior for $\A =[\blambda, \bxi]$ where $\blambda = [\log(\lambda_1^2); \ldots; \log(\lambda_M^2)]$ and $\bxi = [\xi_1; \ldots; \xi_M]$, 
\begin{align}
	q(\A) = \mathcal{MN}(\A | \bar{\A}, \bOmega_1, \bOmega_2) = \N(\vec(\A)| \vec(\bar{\A}), \bOmega_1 \kron \bOmega_2),
\end{align}
where $\bar{\A}$ is the mean, $\bOmega_1$ is the $M\times M$ row covariance matrix, and $\bOmega_2$ is the $2 \times 2$  column covariance matrix. To ensure the positive definiteness, we further parameterize $\bOmega_1 = \L_1 \L_1^\top$ and $\bOmega_2 = \L_2 \L_2^\top$ where $\L_1$ and $\L_2$ are lower-triangular matrices. In this way, not only can we capture the full posterior correlations in $\A$, we also save the parameters and do not need to estimate the $2M \times 2M$ entire covariance matrix.  The matrix Gaussian is also convenient for us to use the reparameterization trick.

A severe computational challenge arises when the number of collocation points $G$ is large ---  the covariance matrix of the multivariate Gaussian prior $p(\u, \g) =  \N\left([\u; \g]|\0, \bSigma\right)$ in \eqref{eq:joint-2} is huge and prohibitively costly to calculate. To address this issue, we use an idea similar to the variational inducing-point method~\citep{hensman2013gaussian}. We observe that the joint prior can be decomposed as $p(\u, \g) = \N(\u|\0, \bSigma_{uu}) p(\g|\u)$, where $p(\g|\u)$ is a giant conditional Gaussian distribution. We then introduce the variational posterior for $\u$ and $\g$ as %$q(\u, \g) = q(\u) p(\g|\u)$, 
\begin{align}
	q(\u, \g) = q(\u) p(\g|\u), 
\end{align}
where $q(\u) = \N(\u|\bmu_u, \L_u\L_u^\top)$ is a Gaussian distribution, and $\L_u$ is a lower triangular matrix. Accordingly, when we compute the log fraction inside the ELBO in \eqref{eq:elbo}, the conditional Gaussian will cancel out, 
\begin{align}
	&\Lcal = \EE_{q}\left[\frac{\N(\u|\0, \bSigma_{uu}) \bcancel{p(\g|\u)}  p(\wy|\bPhi, \w)\cdot \text{Others}}{q(\u) \bcancel{p(\g|\u)} q(\text{Others})}\right] \notag \\
	&= -\text{KL}(q(\u) \| \N(\u|\0, \bSigma_{uu})) + \sum\nolimits_{n=1}^G \EE_{q}\left[\log  \N(\widehat{y}_n | \rho_n - \sum\nolimits_j w_j \phi_{jn}, \sigma_1^2) \right] + \text{Others}, 
\end{align} 
where $\text{KL}(\cdot \| \cdot)$ is the Kullback-Leibler divergence, and $\bSigma_{uu} = \L_u \L_u^\top$. Since the virtual likelihood is a diagonal Gaussian (see \eqref{eq:vll}), the ELBO is additive over each collocation point. That means, we no longer need to compute the giant full conditional Gaussian $p(\g|\u)$. Instead, we compute the KL divergence regarding $\u$ only (based on the small training dataset), and  the conditional distribution of the elements of $\g$ associated with each individual collocation point, which is a small subset, and hence much cheaper to compute.  The additive form over the collocation points further enables us to  sample a mini-batch of collocation points each time, and use the reparameterization trick to update the posterior parameters with stochastic optimization.  %We sample a mini-batch of collocation points each time, and use the reparameterization trick to update the posterior parameters. 

Finally, our variational posterior has the following form, 
\begin{align}
	q(\Theta) =  \LN(s^2|\mu_s, \tau_s) q(\alpha) \prod\nolimits_{j} q(v_j) \cdot \mathcal{MN}(\A|\bar{\A}, \L_1\L_1^\top, \L_2\L_2^\top)q(\u) p(\g|\u). \notag 
\end{align}
We substitute it into \eqref{eq:elbo} to obtain the actual ELBO, and then apply stochastic mini-batch  optimization.  We use the mean-field variational update~\citep{bishop2006pattern} for $q(\alpha)$ and each $q(v_j)$, which gives the optimal form, \ie inverse Gamma, and thereby we do not need to design an additional posterior approximation term.  Each step, we first update the posterior of the other variables via stochastic gradient ascent, and then update $q(\alpha)$ and each $q(v_j)$ accordingly.   
}

%equation discovery important --> symbolic regression --> SINDy --> PINN-SNR  (add more references), then talk about PINNs --> PINN related work --> then GP based work 

\section{\MakeUppercase{Related Work}}
\label{sxn:related-work}
%\vspace{-0.1in}
The recent break-through SINDy~\citep{brunton2016discovering} uses sparse linear regression to select PDE operators from a pre-specified dictionary. The sparsity is fulfilled by a sequential threshold ridge regression (STRidge) method, which repeatedly conducts least-mean-square estimation and weight truncation. 
%SINDy has been used and extended for many domains, such as fluid flows~\citep{loiseau2018constrained,loiseau2018sparse}, chemical and biological physics~\citep{mangan2016inferring,hoffmann2019reactive}, implicit nonlinear dynamics discovery~\citep{kaheman2020sindy}, and active matter~\citep{cichos2020machine}. 
While SINDy was originally developed to discover ODEs, by augmenting the dictionary with partial derivatives over spatial variables, SINDy can be extended for discovering PDEs for spatial-temporal systems~\citep{rudy2017data,schaeffer2017learning}. 
%Such extension has been used to find parametric PDEs from molecular simulation~\citep{zhang2020data}, biological transport models~\citep{lagergren2020learning}, \etc 
The SINDy family of approaches use numerical differentiation  to evaluate candidate operators, and it can suffer from  scarce and noisy data. To alleviate this issue,  recent works~\citep{berg2019data,both2021deepmod,chen2021physics} use deep NNs to approximate the solution, applying automatic differentiation to evaluate and select the operators. These  methods can therefore be viewed as instances or extension of PINNs~\citep{raissi2019physics,krishnapriyan2021characterizing}. 
In \citep{berg2019data}, a deep NN is first used to fit the data, and then a sparse linear regression with $L_1$ regularization is applied to select the operators. 
In \citep{both2021deepmod}, the NN and sparse linear regression are jointly trained.  
In \citep{chen2021physics}, a joint training strategy is also used, but it performs an alternating optimization of the NN loss and sparse  regression, and it uses STRidge for weight truncation.  
The most recent work~\citep{sun2022bayesian} develops a Bayesian spline learning (BSL) method, which uses splines to approximate the solution and a student-$t$ prior, \ie relevance vector machines~\citep{tipping2001sparse}, for sparse linear regression. It uses a similar alternating optimization strategy to \citep{chen2021physics}. To quantify the uncertainty, BSL then applies Stochastic Weight Averaging-Gaussian (SWAG)~\citep{maddox2019simple} for posterior approximation of the  weights.

Our work uses  kernel methods to estimate the solution function for PDE discovery. GP or kernel methods have been applied for solving differential equations for a long time. For example, \citet{graepel2003solving,raissi2017machine} used GPs to solve linear PDEs with the noisy source term measurements. The recent work \citep{chen2021solving}  developed a general kernel method to solve nonlinear PDEs.  
\citet{fang2023solving} developed a spectral mixture kernel for solving high frequency and multi-scale PDEs. 
However, all these methods assume the equations are given, and thus they cannot discover the equations from data.  GPs have also been used to model the equation components or incorporate the equations for better training, such as~\citep{heinonen2018learning,alvarez2009latent,barber2014gaussian,macdonald2015controversy,wenk2019fast,wenk2020odin}. \citet{long2022autoip} proposed a general framework to integrate PDEs/ODEs into GP training. 
%For example, \citet{heinonen2018learning} used GPs to model the dynamics of unknown ODEs.  \citet{alvarez2009latent} proposed latent force models to incorporate incomplete equations into GP training. Other examples include 
%~\citep{barber2014gaussian,macdonald2015controversy, heinonen2018learning,lorenzi2018constraining,wenk2019fast,wenk2020odin,pan2020physics}. 
The GP community has realized the computational advantage of the Kronecker product~\citep{saatcci2012scalable}, and there has been work in leveraging the Kronecker product properties to improve the training speed and scalability, such as high-dimension output regression~\citep{zhe2019scalable,li2021scalable,wang2021multi} and sparse approximation based on inducing points~\citep{wilson2015kernel}. 
However, in typical machine learning applications, the data are not observed at a grid and the Kronecker product has a limited usage. By contrast, for PDE solving, it is natural to estimate the solution values on a mesh (which is consistent with the practice of traditional numerical methods).  This opens the possibility of using Kronecker products to enable efficient PDE solution estimation. %To our knowledge, our work is first to realize this point and use the Kronecker  product structure to enable efficient PDE discovery and solution estimation.

\section{\MakeUppercase{Empirical Results}}
\label{sec:experiments}

For evaluation, we considered several benchmark ODEs/PDEs in the literature of data-driven equation discovery~\citep{chen2021physics,sun2022bayesian}. We examined wether \ours can recover these equations from sparse, noisy measurements. We used the SE kernel and implemented \ours with Jax~\citep{frostig2018compiling}. We compared with state-of-the-art data-driven equation discovery methods, including SINDy~\citep{brunton2016discovering}, PINN-SR~\citep{chen2021physics} and BSL~\citep{sun2022bayesian}. For the competing methods, we used the original implementation released by the authors. For a fair comparison, we carefully tuned every method to achieve the best performance (to our best effort). We leave the details about the settings and tuning of all the methods in Appendix Section \ref{sect:expr-setting}. 

\textbf{Van del Pol (VDP) oscillator.} We first tested the discovery of a nonlinear ODE system, $x_t = y, y_t =  \mu (1 - x^2)y - x$
%\begin{align}
%    x_t = y,\;\; y_t = \mu (1 - x^2)y - x, \;\;\; t \in [0, 8]
%\end{align}
where $\mu = 2.5$ and $x(0) = y(0) = 1$. We used two test settings.  The first setting employed only 10 evenly-spaced training examples in $[0, 8]$ while the second setting used $25$ such examples. For each setting, we considered three noise levels: 0\%, 1\% and 20\%. The dictionary includes the combination of the polynomials of each state variable up to 4th order, which gives $24$ candidates in total. To evaluate the discovery performance, we followed~\citep{sun2022bayesian} to examine the normalized root-mean-square error (RMSE) of the weight estimation (including the relevant and irrelevant operators), and the precision and the recall of the selected operators.  %The results are shown in Table XX. As we can see ...

\textbf{Lorenz 96.} We next tested on Lorenz 96, a well-known chaotic ODE system, $\frac{\d x_i}{\d t} = (x_{i+1} - x_{i-2})x_{i-1} - x_i + F$, where $t \in [0, 5]$, $1 \le i \le n$, $x_0 = x_n$, $x_{-1} = x_{n-1}$ and $x_{n+1} = x_1$. 
We set $n=6$ to have six state variables and set forcing term $F = 8$. The initial condition is 2 for every state. The dictionary includes the polynomials of the state variables up to 3rd order, leading to $84$ candidates in total. We performed two tests: one used 12 evenly-spaced examples, and the other used 50 such examples in the domain. 
\begin{figure*}[t]
				\centering
				\setlength{\tabcolsep}{0pt}
				\begin{tabular}[c]{ccccc}
					\begin{subfigure}[b]{0.22\textwidth}
						\centering
						\includegraphics[width=\linewidth]{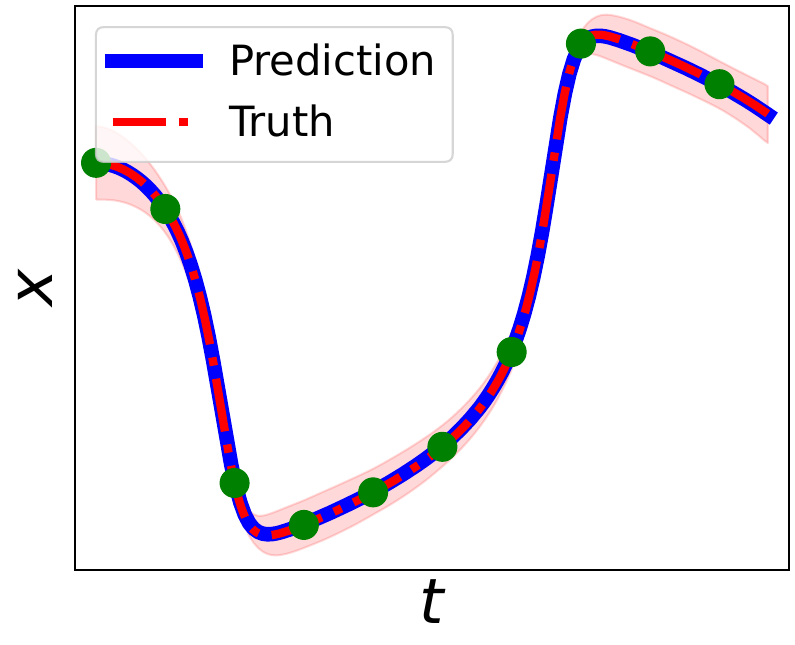}
						\caption{}
					\end{subfigure} &
					\begin{subfigure}[b]{0.22\textwidth}
						\centering
						\includegraphics[width=\linewidth]{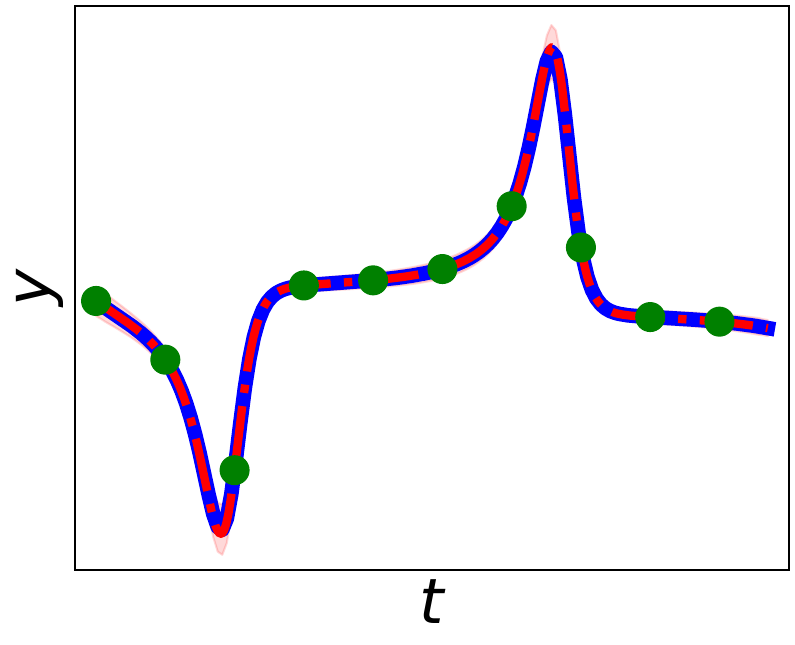}
						\caption{}
					\end{subfigure} &
				\begin{subfigure}[b]{0.218\textwidth}
					\centering
					\includegraphics[width=\linewidth]{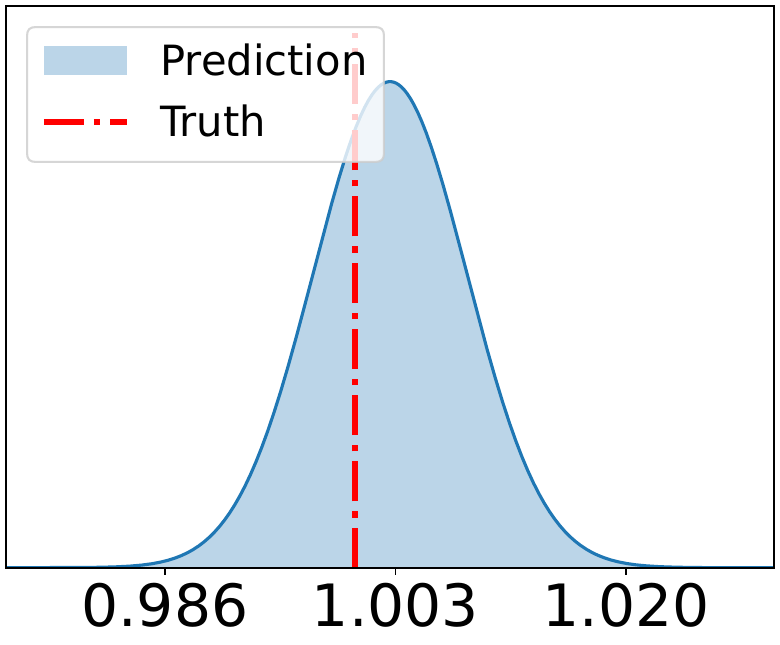}
                    \caption{}
					%\label{fig:poisson1d-mix_sin-GP-SE}
				\end{subfigure} &
				\begin{subfigure}[b]{0.218\textwidth}
					\centering
					\includegraphics[width=\linewidth]{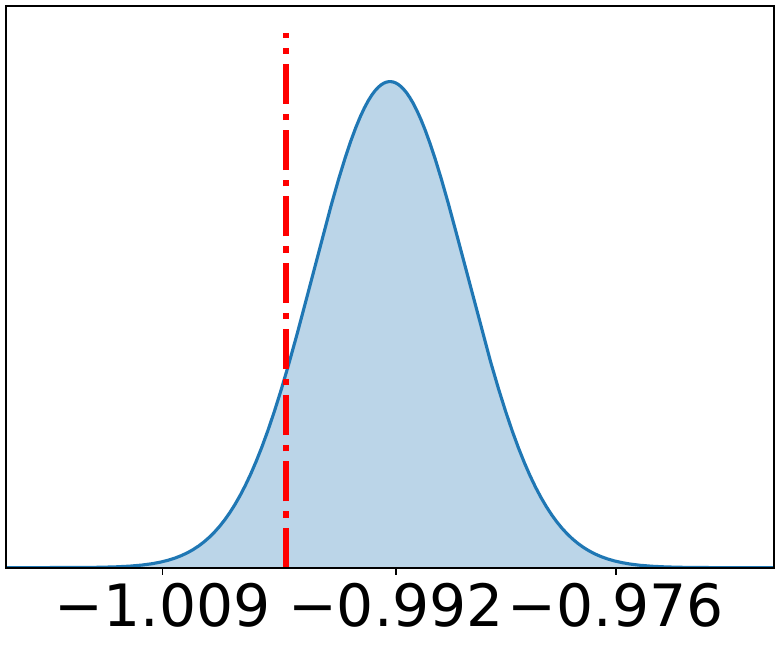}
					%\caption{$q(w_2)$}
                    \caption{}
					%\label{fig:poisson1d-mix_sin-GP-Matern}
				\end{subfigure} &
    %             \begin{subfigure}[b]{0.2\textwidth}
				% 	\centering
				% 	\includegraphics[width=\linewidth]{./NeurIPS2023/fig/osc/os_w4.pdf}
				% 	\caption{$q(w_4)$}
				% 	%\label{fig:poisson1d-mix_sin-GP-Matern}
				% \end{subfigure} 
				\end{tabular}
				%\vspace{-0.1in}
				\caption{\small Solution and weight posterior estimation for the VDP equation with $10$ training examples (marked as green); (c) and (d) show the weight posterior for terms $y$ and $x$, respectively. Their posterior selection probabilities were both estimated as $1.0$.}
				%\vspace{0.02in}
				\label{fig:vdp}
\end{figure*}
\begin{figure*}[t]
				\centering
				\setlength{\tabcolsep}{0pt}
				\begin{tabular}[c]{ccccc}
					\begin{subfigure}[b]{0.24\textwidth}
						\centering
						\includegraphics[width=\linewidth]{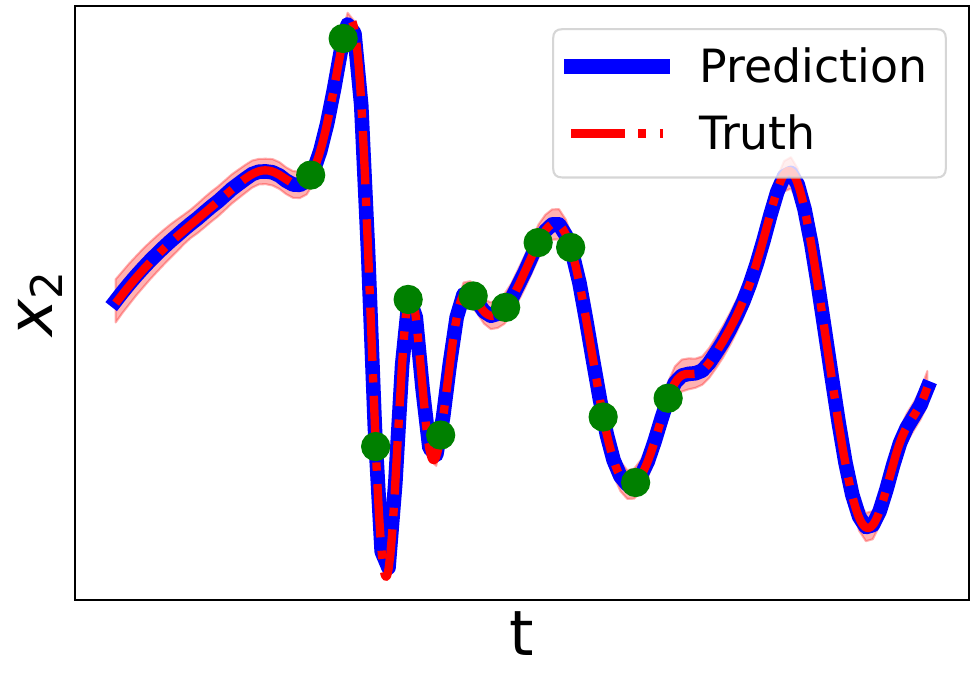}
						\caption{\cmt{50 * std}}
						%\label{fig:poisson1d-mix_sin-GP-HM-SMM}
					\end{subfigure} &
					\begin{subfigure}[b]{0.24\textwidth}
						\centering
						\includegraphics[width=\linewidth]{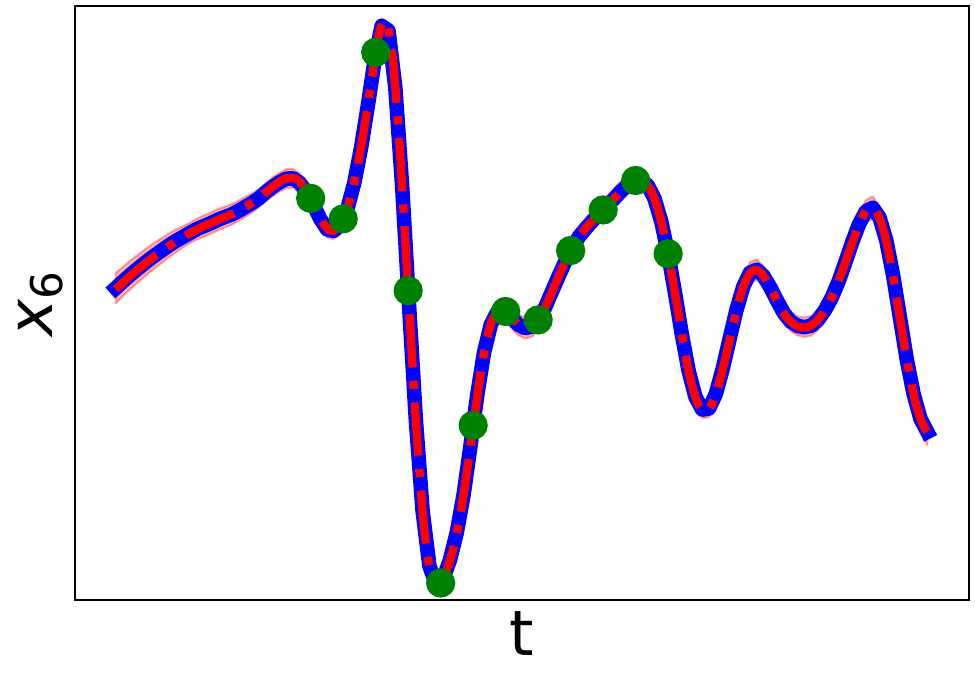}
						\caption{}
						%\label{fig:poisson1d-mix_sin-GP-HM-SM}
					\end{subfigure} &
				\begin{subfigure}[b]{0.20\textwidth}
					\centering
					\includegraphics[width=\linewidth]{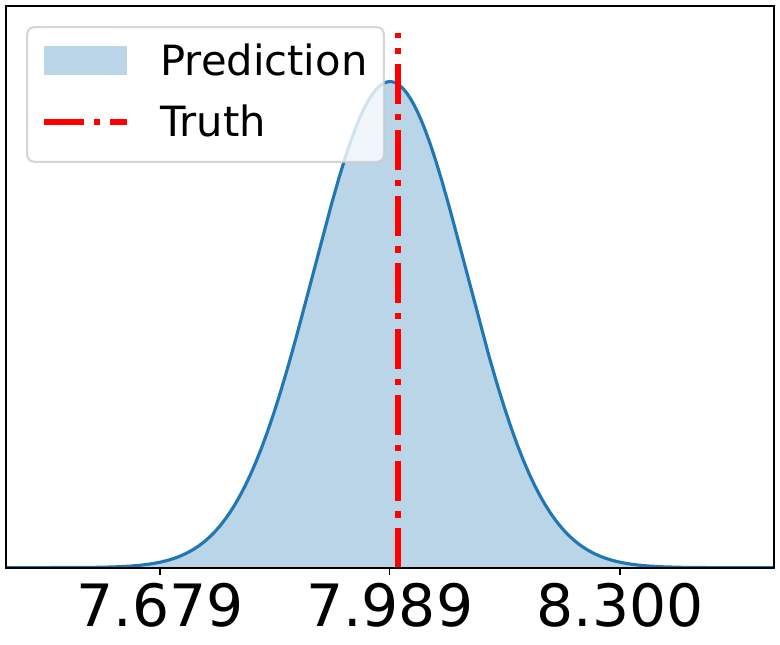}
					\caption{}
					%\label{fig:poisson1d-mix_sin-GP-Matern}
				\end{subfigure} &
                \begin{subfigure}[b]{0.20\textwidth}
					\centering
					\includegraphics[width=\linewidth]{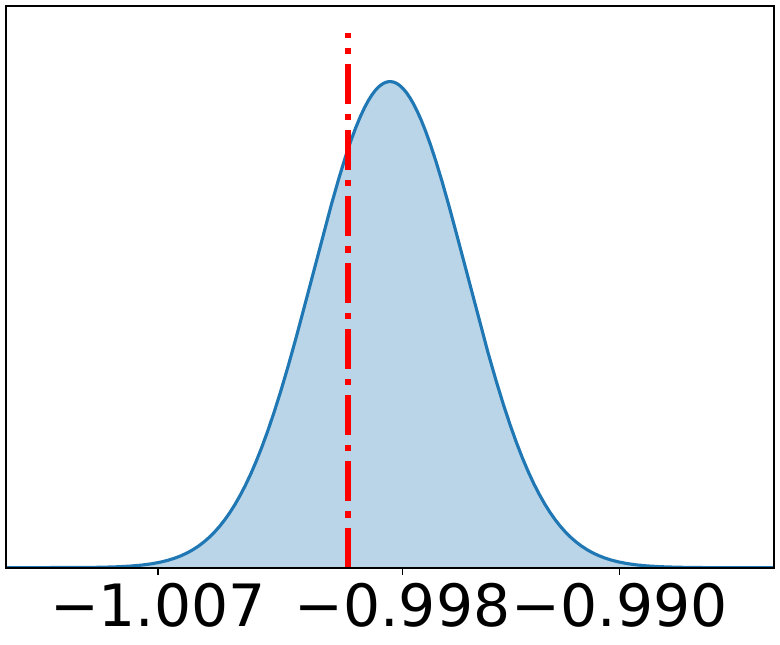}
					\caption{}
					%\label{fig:poisson1d-mix_sin-GP-Matern}
				\end{subfigure} 
				\end{tabular}
				%\vspace{-0.1in}
				\caption{\small Solution and weight posterior estimation for Lorenz 96 using 12 training examples; (c) and (d) show the weight posterior for the force term $F$ and $x_5$. The posterior selection probabilities were estimated as 1.0.}
				%\vspace{-0.1in}
				\label{fig:lorenz}
\end{figure*}
\begin{figure*}[h!]
	\centering
	\setlength{\tabcolsep}{0pt}
	\begin{tabular}[c]{cccc}
	    \begin{subfigure}[b]{0.27\textwidth}
		\centering
		\includegraphics[width=\linewidth]{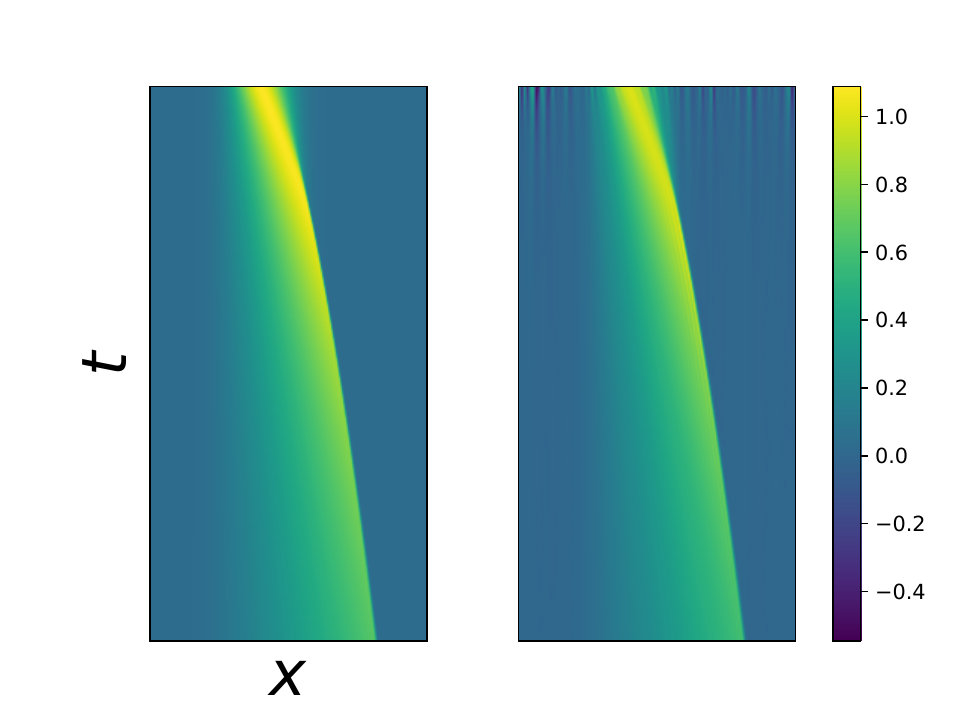}
		\caption{Left: truth; right: prediction}
		\end{subfigure} &
        \begin{subfigure}[b]{0.19\textwidth}
        \includegraphics[width=\textwidth]{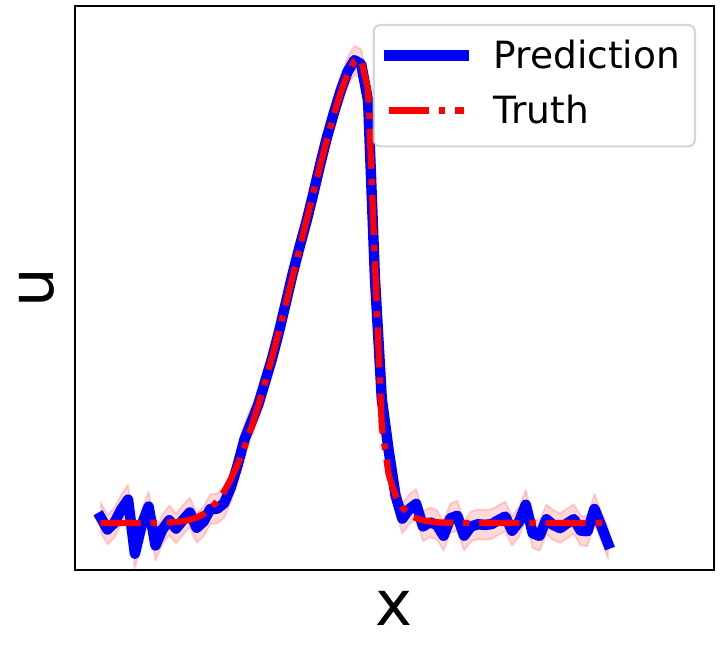}
        \caption{\small $t = 1.1$\cmt{, 15 * std}}
        \label{fig:f3}
        \end{subfigure} &
	    \begin{subfigure}[b]{0.19\textwidth}
        \includegraphics[width=\textwidth]{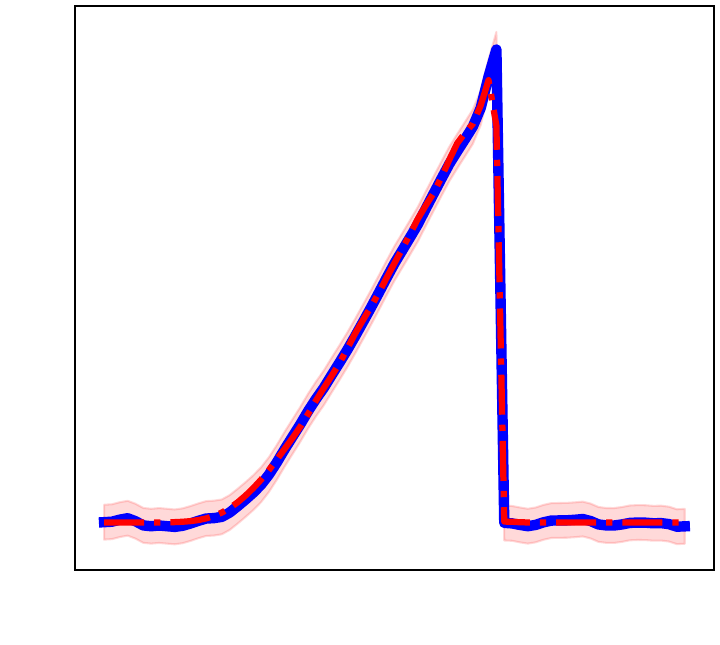}
        \caption{\small $t = 4.0$}
         \label{fig:f2}
         \end{subfigure} &
	       \begin{subfigure}[b]{0.19\textwidth}
          \includegraphics[width=\textwidth]{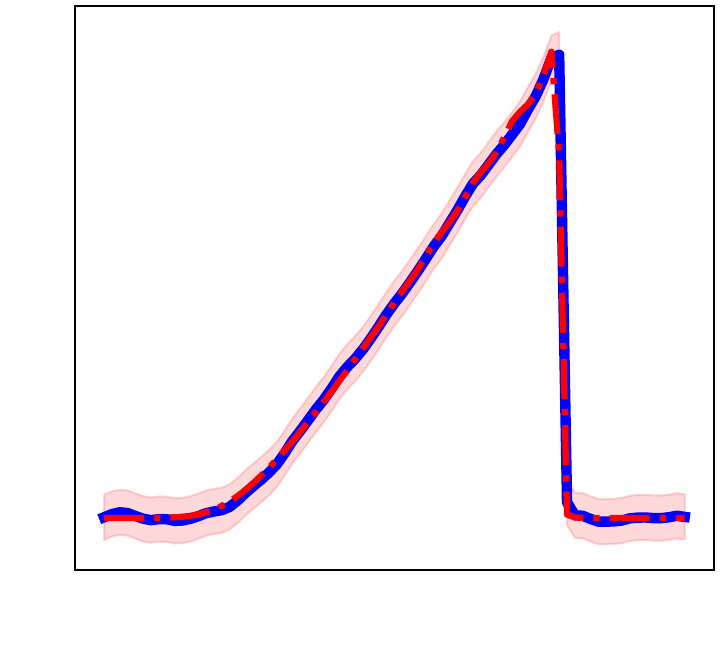}
          \caption{\small $t = 7.0$}
         \label{fig:f4}
         \end{subfigure}
	\end{tabular}
	%\vspace{-0.1in}
	\caption{\small Solution estimate for Burger's equation with $\nu = 0.005$ with 20\% noise on the training data.}
	%\vspace{0.02in}
	\label{fig:Burgers}
\end{figure*}
\begin{figure*}[h!]
	\centering
	\setlength{\tabcolsep}{0pt}
	\begin{tabular}[c]{cccc}
	    \begin{subfigure}{0.30\textwidth}
        \includegraphics[width=\textwidth]{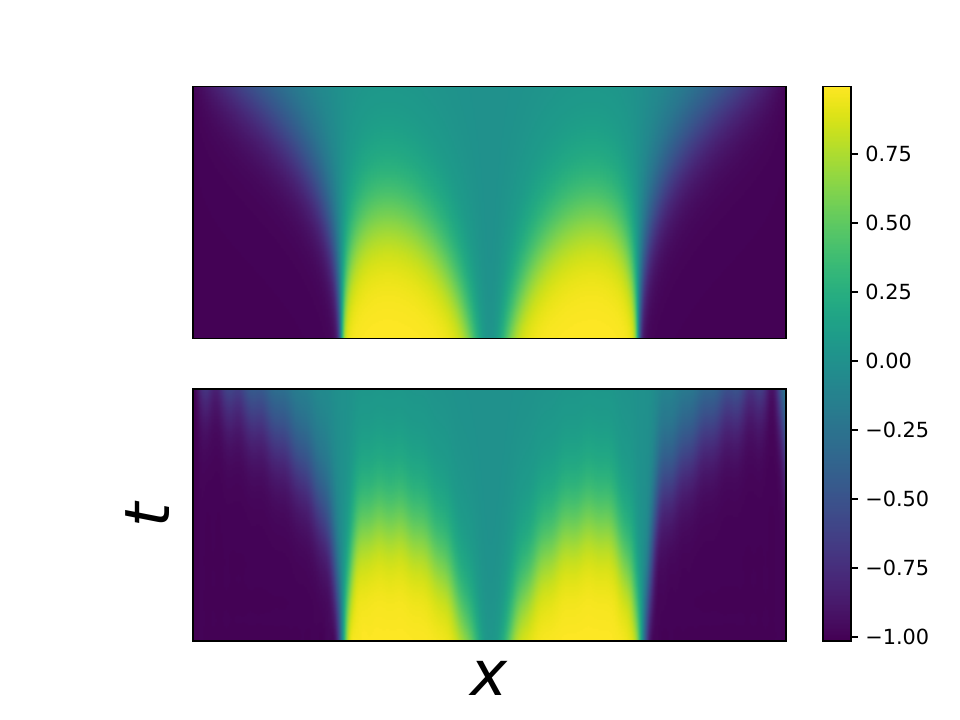}
        \caption{Top: truth; bottom: prediction}
  \end{subfigure} &
       \begin{subfigure}[b]{0.20\textwidth}
        \includegraphics[width=\textwidth]{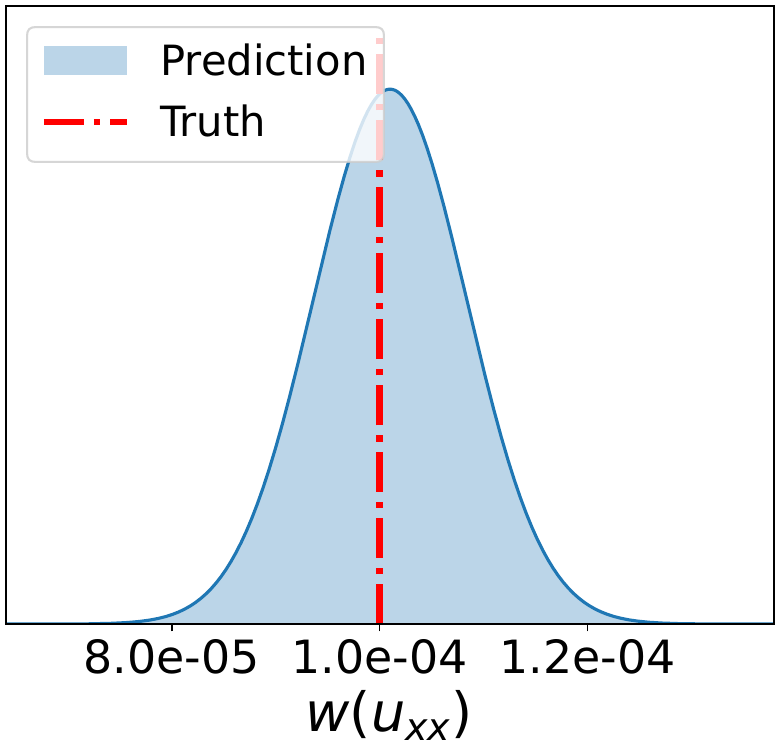}
        \caption{\small $q(s(u_{xx})=1)=1.0$} \label{fig:allen-cahn-uxx}
     \end{subfigure} &
  \begin{subfigure}[b]{0.20\textwidth}
    \includegraphics[width=\textwidth]{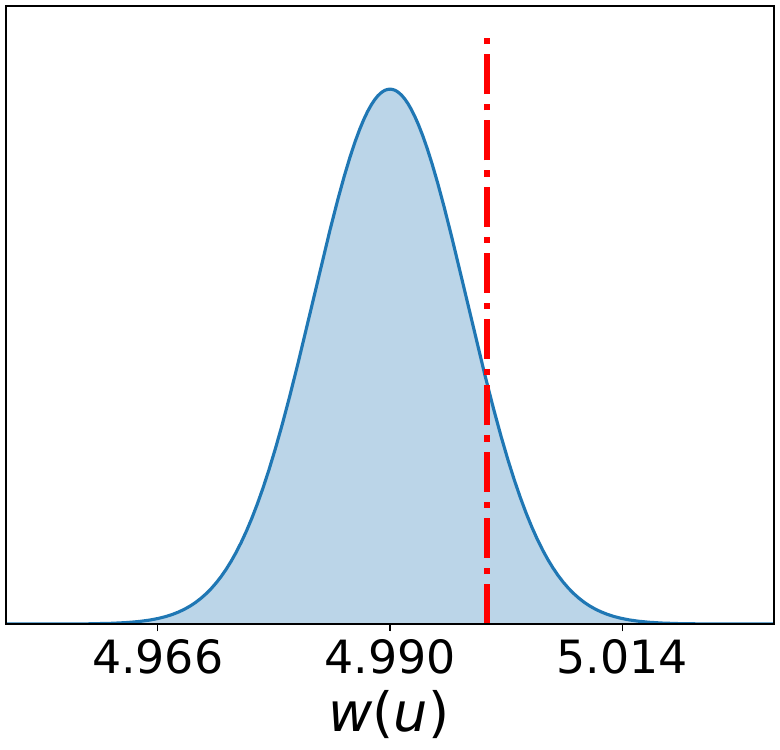}
    \caption{\small $q(s(u)=1)=1.0$}
  \end{subfigure} &
  \begin{subfigure}[b]{0.20\textwidth}
    \includegraphics[width=\textwidth]{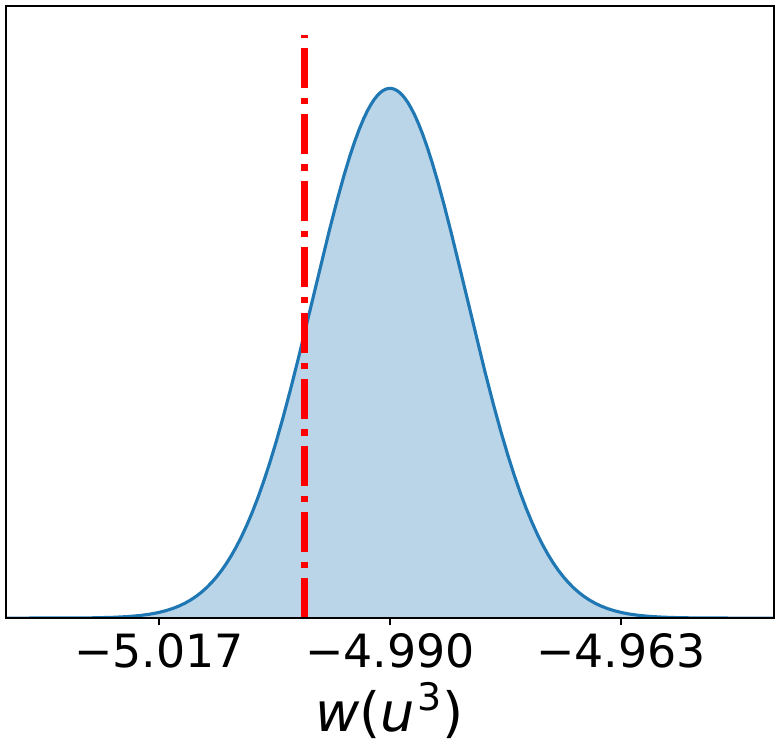}
    \caption{\small $q(s(u^3)=1)=1.0$}
  \end{subfigure}
	\end{tabular}
	\vspace{-0.1in}
	\caption{\small Solution and weight posterior estimation on Allen-cahn equation; $w(\cdot)$ and $s(\cdot)$ denote the weight and selection indicator of the operator. Note the ground-truth weight for $u_{xx}$ is $10^{-4}$.}
	\vspace{-0.1in}
	\label{fig:allen-cahn}
\end{figure*}

\textbf{Burger's equation.} Third, we tested on 1D Burger's equations, 
\begin{align}
    u_t +  u u_x - \nu u_{xx} = 0, \;\;\; (x, t) \in  [0, 10] \times [0, 8] ,
\end{align}
where the initial condition is $u(x, 0) = \exp(-(x-4)^2)$. We tested with three viscosity values, $\nu = 0.1$, $\nu = 0.01$, and $\nu = 0.005$. The dictionary consists of the polynomials and derivatives of $u$ up to 4th order and their combinations, which in total includes 24 candidate operators. 
For $\nu = 0.1$, we first tested with 100 training examples on an evenly-spaced $10\times 10$ grid in the domain. We then tested with a densor dataset, including 400 examples at a $20 \times 20$ evenly-spaced grid. Next, for $\nu = 0.01$ and $\nu = 0.005$, we used training examples at a $50 \times 50$ evenlly-spaced grid.

\textbf{Kuramoto-Sivashinsky (KS) equation.} Fourth, we tested the discovery of a KS equation, % that includes a fourth-order derivative, 
\begin{align}
    u_t + uu_x + u_{xx} + u_{xxxx} = 0, \;\; (x, t)  \in  [0, 32\pi] \times [0, 150]. \notag 
\end{align}
We used a periodic boundary condition $u(x,0)= \cos(x/16)$ for data generation. 
The dictionary includes polynomials and derivatives of $u$ up to 4th order and their combinations, which in total gives 24 candidate operators. We used training examples at an evenly spaced $40 \times 40$ grid in the domain.%, and varied the noise level from $5\%$ and $10\%$.

\begin{table*}[t]
	\small
	\centering
	\begin{tabular}[c]{cccc|ccc}
		\toprule
		\multirow{2}{*}{\textit{Method}} & \multicolumn{3}{c}{10 examples} & \multicolumn{3}{c}{25 examples}\\
  \cline{2-7}
            & (noise)\; 0\% & 1\% & 20\% & 0\%  & 1\% & 20\%\\
		\hline 
        SINDy & F/0.5/1 & F/0.5/1 & F/0.5/1 & F/0.75/0.6 & F/0.75/0.6 & F/0.75/0.75\\
        PINN-SR & F/1/0.67 & F/0.5/0.13 & F/0.75/0.23 & F/0.25/0.33 & F/0.5/0.33 & F/0/0\\
        BSL & F/0.5/1 & F/0.5/1 & F/0.5/0.12& 0/1/1 & 0.0029/1/1 & F/0.75/1\\
        \ours & \textbf{0.0086/1/1} & \textbf{0.032/1/1} & \textbf{0.0995/1/1} & \textbf{0/1/1} & \textbf{0.0029/1/1} & \textbf{0.042/1/1}\\
		\bottomrule
	\end{tabular}
    \caption{\small Performance of discovering VDP oscillators ($t \in [0, 8]$), with 0\%, 1\% and 20\% noises in the training data. Each entry is normalized RMSE/Recall/Precision and ``F'' means Failed. } \label{tb:vdp}
\end{table*}
\begin{table*}[t]
	\small
	\centering
	\begin{tabular}[c]{cccc|ccc}
		\toprule
		\multirow{2}{*}{\textit{Method}} & \multicolumn{3}{c}{12 examples} & \multicolumn{3}{c}{50 examples}\\
  \cline{2-7}
            & (noise)\;0\%  & 1\% & 10\% & 0\%  & 1\% & 10\%\\
		\hline 
        SINDy & F/0.083/0.29 & F/0.125/0.38 & F/0.083/0.33 & F/0.17/0.8 & F/0.17/0.8 & F/0.083/0.5\\
        PINN-SR & F/0/0 & F/0/0 & F/0/0 & F/0/0 & F/0/0 & F/0/0\\
        BSL & F/0.17/0.5 & F/0.083/0.33 & F/0.083/0.33& 0.0011/1/1 & 0.0049/1/1 & F/0.67/0.25\\
        \ours & \textbf{0.0068/1/1} & \textbf{0.0125/1/1} & \textbf{0.134/1/1} & \textbf{0/1/1} & \textbf{0.0035/1/1} & \textbf{0.054/1/1}\\
		\bottomrule
	\end{tabular}
	\caption{\small Performance of discovering Lorenz 96 systems with 6 state variables and $t \in [0, 5]$.} \label{tb:lorenz}
\end{table*}
\begin{table*}[h!]
	\small
	\centering
 \begin{subtable}{\textwidth}
 \centering 
	\begin{tabular}[c]{cccc|ccc}
		\toprule
		\multirow{2}{*}{\textit{Method}} & \multicolumn{3}{c}{ $10\times 10$ examples} & \multicolumn{3}{c}{$20\times 20$ examples}\\
  \cline{2-7}
            & (noise)\;0\%  & 1\% & 20\% & 0\%  & 1\% & 20\% \\
		\hline 
        SINDy & F/0.5/0.2 & F/0.5/0.2 & F/0.5/0.67 & F/1/0.4 & F/1/0.4 & F/0/0\\
        PINN-SR  & 0/1/1 & F/1/0.4 & F/1/0.4 & 0/1/1 & F/1/0.4 & F/1/0.13\\
        BSL & F/0.5/1 & F/0/0 & F/0/0& 0/1/1 & 0/1/1& F/0/0\\
        \ours & \textbf{0/1/1} & \textbf{0.0032/1/1} & \textbf{0.038/1/1} & \textbf{0/1/1} & \textbf{0/1/1} & \textbf{0.038/1/1}\\
		\bottomrule
	\end{tabular}
 \caption{\small $\nu=0.1$}\label{tb:burgers0.1}
    \end{subtable}
    \begin{subtable}{\textwidth}
 \centering 
	\begin{tabular}[c]{cccc|ccc}
		\toprule
		\multirow{2}{*}{\textit{Method}} & \multicolumn{3}{c}{$\nu=0.01$} & \multicolumn{3}{c}{$\nu=0.005$}\\
  \cline{2-7}
            & 0\%  & 10\% & 20\% & 0\%  & 10\% & 20\% \\
		\hline 
        SINDy & F/0.5/0.2 & F/0.5/0.2 & F/0.5/0.2 & F/0.5/0.2 & F/1/0.4 & F/1/0.4\\
        PINN-SR & F/1/0.14 & F/0.5/0.2 & F/0.5/0.5 & F/0.5/0.2 & F/0.5/0.2 & F/0/0\\
        BSL & F/0.5/0.33 & F/0/0 & F/0/0& F/0.5/0.25 & F/0/0 & F/0/0\\
        \ours & \textbf{0.003/1/1} & \textbf{0.007/1/1} & \textbf{0.008/1/1} & \textbf{0.0013/1/1} & \textbf{0.0026/1/1} & \textbf{0.0047/1/1}\\
		\bottomrule
	\end{tabular}
 \caption{\small $\nu=0.01$ and $\nu=0.005$ on $50 \times 50$ examples.}\label{tb:burgers0.01}
    \end{subtable}
	\caption{\small Performance of discovering a 1D Burger's equation, with $(x, t) \in [0, 10]\times[0, 8]$.} \label{tb:burgers}
\end{table*}
\begin{table*}[h!]
	\small
	\centering
	\begin{tabular}[c]{cccc|ccc}
		\toprule
		\multirow{2}{*}{\textit{Method}} & \multicolumn{3}{c}{{KS equation}} & \multicolumn{3}{c}{{Allen-cahn equation}}\\
  \cline{2-7}
            & (noise)\;0\%  & 10\% & 20\% & 0\%  & 5\% & 10\%\\
		\hline 
        SINDy & F/0.33/0.2& F/0.33/0.2 & F/0.33/0.2 & F/0.67/1 & F/0.67/1 & F/0.67/1\\
        PINN-SR & F/0/0 & F/0/0 & F/0.33/0.33 & F/0.67/0.33 & F/0.67/0.33 & F/0.67/0.33\\
        BSL & F/0/0 & F/0/0 & F/0/0& F/0.67/0.67 & F/0.67/1 & F/0.67/0.5\\
        \ours & \textbf{0.013/1/1} & \textbf{0.023/1/1} & \textbf{0.025/1/1} & \textbf{0.0023/1/1} & \textbf{0.0071/1/1} & \textbf{0.045/1/1}\\
		\bottomrule
	\end{tabular}
	\caption{\small Performance of discovering a 1D KS equation with $(x, t) \in [0, 32\pi] \times [0, 150]$, and 1D Allen-cahn equation with $(x, t) \in [-1, 1] \times [0, 1]$.} \label{tb:ks-allen}
\end{table*}

\textbf{Allen-cahn equation.} Fifth, we tested on an Allen-cahn equation with a very small diffusion rate, 
\begin{align}
    u_t = 10^{-4} u_{xx} + 5u - 5u^3, \;\; (x, t) \in [-1, 1] \times [0, 1], \notag 
\end{align}
where $u(x,0) = x^2\cos(\pi x)$, $u(-1,t) = u(1, t)$ and $u_x(-1, t) = u_x(1, t)$. The dictionary consists of the polynomials and derivatives of $u$ up to 3rd order and their combinations, which in total includes 16 candidate operators.
We used training examples at a $26 \times 101$ grid in the spatial-temporal domain. 

\textbf{Real-world application.} Finally, we ran our method on the data of a real-world predator-prey system. The dataset and experimental details are given in Appendix Section \ref{sect:real-world}. %The data includes the population of hares and lynx in Hudson Bay Company during 1900-1920. The data and experimental  

\textbf{Results and discussion.} We report the discovery performance of each method in Table \ref{tb:vdp}, \ref{tb:lorenz}, \ref{tb:burgers}, and \ref{tb:ks-allen} and Appendix Table \ref{tb:real-world}. We can see that in all the settings, \ours successfully recovered the equations, and the error of the operator weight estimate is small. As a comparison, SINDy failed to find the equations for every case. This might be due to that the sampled measurements (\ie training examples) are too sparse. We found that SINDy started to successfully recover VDP, Lorenz 96, Burger's ($\nu=0.1$), and KS equations when we increased to 50, 250, 1.6K and 384.5K examples respectively, which takes 5x and 21x, 16x and 240x of the sample size needed by \ours. Nonetheless, the accuracy of the operator weight estimation is still much worse than \ours; see the results in Appendix Table \ref{tb:sindy-more}. 
PINN-SR failed to discover all the ODEs. This is consistent with the observation in~\citep{sun2022bayesian}. 
Though PINN-SR can exactly recover the Burger's equation ($\nu=0.1$) with $10\times 10$ and $20 \times 20$ examples, when adding a tiny amount of noise (1\%), PINN-SR immediately failed; see Table \ref{tb:burgers0.1}. Similarly, BSL can exactly recover the Burger's equation ($\nu=0.1$) using $20\times 20$ examples, with zero or 1\% noise. However, it failed with 20\% noise. %By contrast, \ours can discover the equation with both sample sizes and 20\% noises, 
Hence, it illustrates that \ours is much more robust to data noise. In addition, both BSL and PINN-SR failed to find the Burger's equation with $\nu=0.01$ and $\nu=0.005$, which are more challenging; see Table \ref{tb:burgers0.01}. So too did they for the KS and Allen-cahn equations; see Table \ref{tb:ks-allen}. Note that although in~\citep{chen2021physics}, PINN-SR successfully recovers a KS equation, it uses $320 \times 101$ examples ---  20x of the examples used by \ours. We were not able to use PINN-SR to recover the KS equation tested in our experiment even with the same amount of data. This might relate to our usage of a much larger domain, $[0, 32\pi] \times [0, 150]$ versus $[0, 32\pi] \times [0, 100]$.

Next, we showcased our solution prediction and posterior estimate of the operator weights in Fig. \ref{fig:vdp}, \ref{fig:lorenz}, \ref{fig:Burgers}, \ref{fig:allen-cahn}, and Appendix Fig. \ref{fig:ks-est}, \ref{fig:solution-lorenz}, \ref{fig:sol-predatory-prey}. The shaded region in the solution prediction indicates the  predictive standard deviation (scaled by a factor to be more clear). We used the Laplace's approximation (see details in Appendix Section \ref{sect:laplace}) to estimate the posterior of the solution prediction.

We can see that the solution prediction by \ours is quite accurate compared to the ground-truth, even when the training data is very sparse and noisy, \eg the prediction for Burger's equation with $\nu=0.005$; see Fig \ref{fig:Burgers}. This might be (partly) due to that the accurate discovery of the equation in our E step can simultaneously guide the solution learning (M step). From the posterior estimate for the operator weights, we can see that the posterior mean  is close to the ground-truth, while the Gaussian distribution allows us to further quantify the uncertainty. It is particularly interesting to see that in Fig. \ref{fig:allen-cahn-uxx}, \ours successfully identifies the operator $u_{xx}$ that has a very small weight $10^{-4}$ and provides a Gaussian posterior approximation. Note that the posterior selection probability $q(s(u_{xx})= 1) = 1.0$, reflecting that \ours has a very high confidence to select $u_{xx}$, even though the scale of the weight is very small. Hence, \ours does not need to use a weight threshold for operator selection.

\noindent\textbf{Computational efficiency.} We examined the running time of each method. The results are given in Appendix Section \ref{sect:run-time}.  As shown in Appendix Table \ref{tb:run-time}, \ours takes much less time than BSL. For example, on Burger's and KS , PINN-SR takes 9x and 20x running time but still failed to recover the KS equation. Together this shows our method also has a significant advantage in computational efficiency.

\section{\MakeUppercase{Conclusion}}
We have presented \ours, an efficient kernel learning method with Bayesian spike-and-slab priors for data-driven equation discovery. Our method has higher sample efficiency and is more resistant to data noise than existing methods. \ours has shown significant advantages on a series of PDE and ODE discovery tasks. Currently, \ours is limited to a pre-specified operator dictionary. In the future, we plan to develop methods that can dynamically expand the dictionary to further improve the accuracy and robustness. 
\section*{Acknowledgements}
This work has been supported by MURI AFOSR grant FA9550-20-1-0358, NSF CAREER Award IIS-2046295 and NSF OAC-2311685.
%\section*{Acknowledgements}
%%%%%%%%%%%%%%%%%%%%%%%%%%%%%%%%%%%%%%%%%%%%%%%%%%%%%%%%%%%%
\bibliographystyle{apalike}
\bibliography{GovEqGP}

\begin{thebibliography}{}

\bibitem[Alvarez et~al., 2009]{alvarez2009latent}
Alvarez, M., Luengo, D., and Lawrence, N.~D. (2009).
\newblock Latent force models.
\newblock In {\em Artificial Intelligence and Statistics}, pages 9--16.

\bibitem[Barber and Wang, 2014]{barber2014gaussian}
Barber, D. and Wang, Y. (2014).
\newblock Gaussian processes for {B}ayesian estimation in ordinary differential equations.
\newblock In {\em International Conference on Machine Learning}, pages 1485--1493.

\bibitem[Berg and Nystr{\"o}m, 2019]{berg2019data}
Berg, J. and Nystr{\"o}m, K. (2019).
\newblock Data-driven discovery of pdes in complex datasets.
\newblock {\em Journal of Computational Physics}, 384:239--252.

\bibitem[Both et~al., 2021]{both2021deepmod}
Both, G.-J., Choudhury, S., Sens, P., and Kusters, R. (2021).
\newblock Deep{M}o{D}: Deep learning for model discovery in noisy data.
\newblock {\em Journal of Computational Physics}, 428:109985.

\bibitem[Brunton et~al., 2016]{brunton2016discovering}
Brunton, S.~L., Proctor, J.~L., and Kutz, J.~N. (2016).
\newblock Discovering governing equations from data by sparse identification of nonlinear dynamical systems.
\newblock {\em Proceedings of the national academy of sciences}, 113(15):3932--3937.

\bibitem[Chen et~al., 2021a]{chen2021solving}
Chen, Y., Hosseini, B., Owhadi, H., and Stuart, A.~M. (2021a).
\newblock Solving and learning nonlinear {PDEs} with {G}aussian processes.
\newblock {\em arXiv preprint arXiv:2103.12959}.

\bibitem[Chen et~al., 2021b]{chen2021physics}
Chen, Z., Liu, Y., and Sun, H. (2021b).
\newblock Physics-informed learning of governing equations from scarce data.
\newblock {\em Nature communications}, 12(1):1--13.

\bibitem[Fang et~al., 2023]{fang2023solving}
Fang, S., Cooley, M., Long, D., Li, S., Kirby, R., and Zhe, S. (2023).
\newblock Solving high frequency and multi-scale pdes with gaussian processes.
\newblock {\em arXiv preprint arXiv:2311.04465}.

\bibitem[Fang et~al., 2020]{fang2020online}
Fang, S., Zhe, S., Lee, K.-c., Zhang, K., and Neville, J. (2020).
\newblock Online {B}ayesian sparse learning with spike and slab priors.
\newblock In {\em 2020 IEEE International Conference on Data Mining (ICDM)}, pages 142--151. IEEE.

\bibitem[Frostig et~al., 2018]{frostig2018compiling}
Frostig, R., Johnson, M.~J., and Leary, C. (2018).
\newblock Compiling machine learning programs via high-level tracing.
\newblock {\em Systems for Machine Learning}, 4(9).

\bibitem[Graepel, 2003]{graepel2003solving}
Graepel, T. (2003).
\newblock Solving noisy linear operator equations by {G}aussian processes: Application to ordinary and partial differential equations.
\newblock In {\em ICML}, pages 234--241.

\bibitem[Heinonen et~al., 2018]{heinonen2018learning}
Heinonen, M., Yildiz, C., Mannerstr{\"o}m, H., Intosalmi, J., and L{\"a}hdesm{\"a}ki, H. (2018).
\newblock Learning unknown {ODE} models with {G}aussian processes.
\newblock In {\em International Conference on Machine Learning}, pages 1959--1968.

\bibitem[Ishwaran and Rao, 2005]{ishwaran2005spike}
Ishwaran, H. and Rao, J.~S. (2005).
\newblock Spike and slab variable selection: Frequentist and {B}ayesian strategies.
\newblock {\em The Annals of statistics}, 33(2):730--773.

\bibitem[Kolda, 2006]{kolda2006multilinear}
Kolda, T.~G. (2006).
\newblock Multilinear operators for higher-order decompositions.
\newblock Technical report, Sandia National Laboratories (SNL).

\bibitem[Krishnapriyan et~al., 2021]{krishnapriyan2021characterizing}
Krishnapriyan, A., Gholami, A., Zhe, S., Kirby, R., and Mahoney, M.~W. (2021).
\newblock Characterizing possible failure modes in physics-informed neural networks.
\newblock {\em Advances in Neural Information Processing Systems}, 34:26548--26560.

\bibitem[Lagergren et~al., 2020]{lagergren2020learning}
Lagergren, J.~H., Nardini, J.~T., Michael~Lavigne, G., Rutter, E.~M., and Flores, K.~B. (2020).
\newblock Learning partial differential equations for biological transport models from noisy spatio-temporal data.
\newblock {\em Proceedings of the Royal Society A}, 476(2234):20190800.

\bibitem[Li et~al., 2021]{li2021scalable}
Li, S., Xing, W., Kirby, R.~M., and Zhe, S. (2021).
\newblock Scalable {G}aussian process regression networks.
\newblock In {\em Proceedings of the Twenty-Ninth International Conference on International Joint Conferences on Artificial Intelligence}, pages 2456--2462.

\bibitem[Long et~al., 2022]{long2022autoip}
Long, D., Wang, Z., Krishnapriyan, A., Kirby, R., Zhe, S., and Mahoney, M. (2022).
\newblock Autoip: A united framework to integrate physics into gaussian processes.
\newblock In {\em International Conference on Machine Learning}, pages 14210--14222. PMLR.

\bibitem[Macdonald et~al., 2015]{macdonald2015controversy}
Macdonald, B., Higham, C., and Husmeier, D. (2015).
\newblock Controversy in mechanistic modelling with {G}aussian processes.
\newblock {\em Proceedings of Machine Learning Research}, 37:1539--1547.

\bibitem[MacKay, 2003]{mackay2003information}
MacKay, D.~J. (2003).
\newblock {\em Information theory, inference and learning algorithms}.
\newblock Cambridge university press.

\bibitem[Maddox et~al., 2019]{maddox2019simple}
Maddox, W.~J., Izmailov, P., Garipov, T., Vetrov, D.~P., and Wilson, A.~G. (2019).
\newblock A simple baseline for {B}ayesian uncertainty in deep learning.
\newblock {\em Advances in neural information processing systems}, 32.

\bibitem[Minka, 2000]{minka2000old}
Minka, T.~P. (2000).
\newblock Old and new matrix algebra useful for statistics.

\bibitem[Minka, 2001a]{minka2001expectation}
Minka, T.~P. (2001a).
\newblock Expectation propagation for approximate {B}ayesian inference.
\newblock In {\em Proceedings of the Seventeenth conference on Uncertainty in artificial intelligence}, pages 362--369.

\bibitem[Minka, 2001b]{minka2001family}
Minka, T.~P. (2001b).
\newblock {\em A family of algorithms for approximate {B}ayesian inference}.
\newblock PhD thesis, Massachusetts Institute of Technology.

\bibitem[Mitchell and Beauchamp, 1988]{mitchell1988bayesian}
Mitchell, T.~J. and Beauchamp, J.~J. (1988).
\newblock Bayesian variable selection in linear regression.
\newblock {\em Journal of the american statistical association}, 83(404):1023--1032.

\bibitem[Mohamed et~al., 2012]{mohamed2012bayesian}
Mohamed, S., Heller, K.~A., and Ghahramani, Z. (2012).
\newblock Bayesian and l1 approaches for sparse unsupervised learning.
\newblock In {\em Proceedings of the 29th International Coference on International Conference on Machine Learning}, pages 683--690.

\bibitem[Mojgani et~al., 2022]{mojgani2022lagrangian}
Mojgani, R., Balajewicz, M., and Hassanzadeh, P. (2022).
\newblock Lagrangian {PINNs}: A causality-conforming solution to failure modes of physics-informed neural networks.
\newblock {\em arXiv preprint arXiv:2205.02902}.

\bibitem[Raissi et~al., 2017]{raissi2017machine}
Raissi, M., Perdikaris, P., and Karniadakis, G.~E. (2017).
\newblock Machine learning of linear differential equations using {G}aussian processes.
\newblock {\em Journal of Computational Physics}, 348:683--693.

\bibitem[Raissi et~al., 2019]{raissi2019physics}
Raissi, M., Perdikaris, P., and Karniadakis, G.~E. (2019).
\newblock Physics-informed neural networks: A deep learning framework for solving forward and inverse problems involving nonlinear partial differential equations.
\newblock {\em Journal of Computational Physics}, 378:686--707.

\bibitem[Ritter et~al., 2018]{ritter2018scalable}
Ritter, H., Botev, A., and Barber, D. (2018).
\newblock A scalable laplace approximation for neural networks.
\newblock In {\em 6th International Conference on Learning Representations, ICLR 2018-Conference Track Proceedings}, volume~6. International Conference on Representation Learning.

\bibitem[Rudy et~al., 2017]{rudy2017data}
Rudy, S.~H., Brunton, S.~L., Proctor, J.~L., and Kutz, J.~N. (2017).
\newblock Data-driven discovery of partial differential equations.
\newblock {\em Science advances}, 3(4):e1602614.

\bibitem[Saat{\c{c}}i, 2012]{saatcci2012scalable}
Saat{\c{c}}i, Y. (2012).
\newblock {\em Scalable inference for structured {G}aussian process models}.
\newblock PhD thesis, Citeseer.

\bibitem[Schaeffer, 2017]{schaeffer2017learning}
Schaeffer, H. (2017).
\newblock Learning partial differential equations via data discovery and sparse optimization.
\newblock {\em Proceedings of the Royal Society A: Mathematical, Physical and Engineering Sciences}, 473(2197):20160446.

\bibitem[Sun et~al., 2022]{sun2022bayesian}
Sun, L., Huang, D., Sun, H., and Wang, J.-X. (2022).
\newblock Bayesian spline learning for equation discovery of nonlinear dynamics with quantified uncertainty.
\newblock {\em Advances in Neural Information Processing Systems}, 35:6927--6940.

\bibitem[Tipping, 2001]{tipping2001sparse}
Tipping, M.~E. (2001).
\newblock Sparse {B}ayesian learning and the relevance vector machine.
\newblock {\em Journal of machine learning research}, 1(Jun):211--244.

\bibitem[Wainwright et~al., 2008]{wainwright2008graphical}
Wainwright, M.~J., Jordan, M.~I., et~al. (2008).
\newblock Graphical models, exponential families, and variational inference.
\newblock {\em Foundations and Trends{\textregistered} in Machine Learning}, 1(1--2):1--305.

\bibitem[Walker, 1969]{walker1969asymptotic}
Walker, A.~M. (1969).
\newblock On the asymptotic behaviour of posterior distributions.
\newblock {\em Journal of the Royal Statistical Society: Series B (Methodological)}, 31(1):80--88.

\bibitem[Wang et~al., 2021]{wang2021multi}
Wang, Z., Xing, W., Kirby, R., and Zhe, S. (2021).
\newblock Multi-fidelity high-order {G}aussian processes for physical simulation.
\newblock In {\em International Conference on Artificial Intelligence and Statistics}, pages 847--855. PMLR.

\bibitem[Wenk et~al., 2020]{wenk2020odin}
Wenk, P., Abbati, G., Osborne, M.~A., Sch{\"o}lkopf, B., Krause, A., and Bauer, S. (2020).
\newblock {ODIN}: {ODE}-informed regression for parameter and state inference in time-continuous dynamical systems.
\newblock In {\em AAAI}, pages 6364--6371.

\bibitem[Wenk et~al., 2019]{wenk2019fast}
Wenk, P., Gotovos, A., Bauer, S., Gorbach, N.~S., Krause, A., and Buhmann, J.~M. (2019).
\newblock Fast {G}aussian process based gradient matching for parameter identification in systems of nonlinear {ODEs}.
\newblock In {\em The 22nd International Conference on Artificial Intelligence and Statistics}, pages 1351--1360. PMLR.

\bibitem[Williams and Rasmussen, 2006]{williams2006gaussian}
Williams, C.~K. and Rasmussen, C.~E. (2006).
\newblock {\em Gaussian processes for machine learning}, volume~2.
\newblock MIT press Cambridge, MA.

\bibitem[Wilson and Nickisch, 2015]{wilson2015kernel}
Wilson, A. and Nickisch, H. (2015).
\newblock Kernel interpolation for scalable structured {G}aussian processes ({KISS-GP}).
\newblock In {\em International conference on machine learning}, pages 1775--1784. PMLR.

\bibitem[Zhang and Ma, 2020]{zhang2020data}
Zhang, J. and Ma, W. (2020).
\newblock Data-driven discovery of governing equations for fluid dynamics based on molecular simulation.
\newblock {\em Journal of Fluid Mechanics}, 892.

\bibitem[Zhe et~al., 2019]{zhe2019scalable}
Zhe, S., Xing, W., and Kirby, R.~M. (2019).
\newblock Scalable high-order {G}aussian process regression.
\newblock In {\em The 22nd International Conference on Artificial Intelligence and Statistics}, pages 2611--2620. PMLR.

\end{thebibliography}
\newpage
\appendix



% Supplementary material: To improve readability, you must use a single-column format for the supplementary material.
\onecolumn

\newpage
\section{APPENDIX}
\subsection{\MakeUppercase{EP Spike-and-Slab Inference}}\label{sect:ep-ss}

\subsubsection{{EP Tutorial}}\label{sect:ep-intro}
We first give a brief  introduction to the expectation propagation (EP) framework~\citep{minka2001expectation}.
Consider a general probabilistic model with latent parameters $\btheta$. Given the observed data $\Dcal = \{\y_1, \ldots, \y_N\}$, the joint probability distribution is 
\begin{align}
	p(\btheta, \Dcal) = p(\btheta) \prod_{n=1}^N p(\y_n|\btheta).
\end{align}
Our goal is to compute the posterior $p(\btheta|\Dcal) = \frac{p(\btheta, \Dcal)}{p(\Dcal)}$. However, it is usually infeasible to compute the exact marginal distribution $p(\Dcal)$, because the complexity of the likelihood and/or prior makes the integration or marginalization intractable. This further makes the exact posterior distribution infeasible to compute. EP therefore seeks to approximate each term in the joint distribution by an exponential-family term, 
\begin{align}
	p(y_n|\btheta) \approx   c_n f_n(\btheta), \;\;\;
	p(\btheta) \approx  c_0	f_0(\btheta),
\end{align}
where $c_n$ and $c_0$ are constants to ensure the normalization consistency (they will get canceled in the inference, so we do not need to calculate them), and 
$$f_n(\btheta) \propto \exp(\blambda_n^\top \bphi(\btheta)) \;\;(0 \le n \le N), $$
where $\blambda_n$ is the natural parameter and $\bphi(\btheta)$ is sufficient statistics. For example, if we choose a Gaussian term, $f_n = \N(\btheta| \bmu_n, \bSigma_n)$, then the sufficient statistics is $\bphi(\btheta) = \{\btheta, \btheta\btheta^\top\}$. The moment is the expectation of the sufficient statistics.%, \ie the mean and covariance 

We therefore approximate the joint distribution with 
\begin{align}
	&p(\btheta, \Dcal) = p(\btheta) \prod_{n=1}^N p(\y_n|\btheta) \approx f_0(\btheta) \prod_{n=1}^N f_n(\btheta) \cdot \text{const}. \label{eq: approx-joint}
\end{align}
Because the exponential family is closed under multiplication and division, we can immediately obtain a closed-form approximate posterior $q(\btheta) \approx p(\btheta|\Dcal)$ by merging the approximation terms in the R.H.S of \eqref{eq: approx-joint}, which is still a distribution in the exponential family.  

Then the task amounts to optimizing those approximation terms $\{f_n(\btheta) | 0 \le n \le N\}$. EP repeatedly conducts four steps to optimize each $f_n$. 
\begin{itemize}
	\item \textbf{Step 1.}\label{text:ep-step-1}  Obtain the calibrated distribution that integrates the context information of $f_n$, 
	\begin{align}
		q^{\backslash n}(\btheta) \propto \frac{q(\btheta)}{f_n(\btheta)}, \label{eq:ep-step-1} 
	\end{align}
	where $q(\btheta)$ is the current posterior approximation. 
	\item \textbf{Step 2.}  Construct a tilted distribution to combine the true likelihood,  
	\begin{align}
		\tp(\btheta) \propto q^{\bkh n} (\btheta) \cdot p(\y_n | \btheta).\label{eq:ep-step-2}
	\end{align}
	Note that if $ n = 0$, we have $\tp(\btheta) \propto q^{\bkh n} (\btheta) \cdot p( \btheta)$.
	\item \textbf{Step 3.}  Project the tilted distribution back to the exponential family, 
    \begin{align}
        q^*(\btheta) = \argmin_q \;\; \kl( \tp \| q), \label{eq:ep-step-3}
    \end{align}
 %$$q^*(\btheta) = \argmin_q \;\; \kl( \tp \| q)$$ 
 where $\text{KL}(\cdot \| \cdot)$ is the Kullback-Leibler divergence, and $q$ belongs to the exponential family. This can be done by moment matching, 
	\begin{align}
		\EE_{q^*}[\bphi(\btheta)] = \EE_\tp[\bphi(\btheta)]. \label{eq:mm}
	\end{align}
	That is, we compute the expected moment under $\tp$, with which to set the parameters of $q^*$.  For example, if $q^*(\btheta)$ is a Gaussian distribution, then we need to compute $\EE_\tp[\btheta]$ and $\EE_\tp[\btheta\btheta^\top]$, with which to obtain the mean and covariance for $q^*(\btheta)$. Accordingly, we obtain $q^*(\btheta) = \N(\btheta| \EE_\tp[\btheta],\EE_\tp[\btheta\btheta^\top] -  \EE_\tp[\btheta] \EE_\tp[\btheta]^\top )$.
	\item \textbf{Step 4.}  Update the approximation term by 
	\begin{align}
		f_n(\btheta) \propto \frac{q^*(\btheta)}{q^{\bkh n}(\btheta)}. \label{eq:ep-step-4}
	\end{align}
\end{itemize}
In practice, EP often updates all the $f_n$'s in parallel. %, and uses damping to avoid divergence. 
 It iteratively runs the four steps until convergence. In essence, this is a fixed point iteration to optimize an energy function (a mini-max problem)~\citep{minka2001expectation,minka2001family,wainwright2008graphical}.

\subsubsection{{EP Method for Our Model}}
We now present how we develop an EP method to estimate the posterior distribution of the spike-and-slab variables, $\s$ and $\w$, for our model. Let us write down the joint distribution here (see \eqref{eq:ep-joint-dist} of the main paper), 
\begin{align}
	&p(\s, \w, \Dcal) \propto  \prod_{j=1}^A \text{Bern}(s_j|\rho_0)   \left(s_j \N(w_j|0, \sigma^2_0) + (1-s_j)\delta(w_j)\right) \cdot \N(\h|\bPhi \w, \tau \I). \label{eq:ep-joint-2}
\end{align}
%Note that we ignore other terms in the joint probability because they are constant to $\s$ and $\w$. 
The mixture factors in the spike-and-slab prior  \eqref{eq:ep-joint-2} do not belong to the exponential family, and thereby make the posterior distribution analytically intractable. So we introduce an approximation, 
\begin{align}
	&s_j \N(w_j|0, \sigma^2_0) + (1-s_j)\delta(w_j) \approx c_j\text{Bern}(s_j|\sigma(\rho_j))\N(w_j|\mu_j,v_j), \label{eq:ep-approx}
\end{align}
where $\sigma(\cdot)$ is the sigmoid function. 
%Note that all the other factors in the joint distribution is in the exponential family, and we do not need to introduce any approximation. 
Substituting the above into \eqref{eq:ep-joint-2}, we can obtain the approximate posterior, 
\begin{align}
	q(\s, \w) = \prod_{j=1}^A \text{Bern}\left(s_j| \sigma(\widehat{\rho}_j)\right) \cdot \N(\w| \bbeta, \bSigma), \label{eq:post-ss-appendix}
\end{align}
where 
\begin{align}
	&\widehat{\rho}_j = \sigma^{-1}(\rho_0) + \rho_j, \;\; \bSigma = \left(\text{diag}^{-1}(\v) + \tau^{-1}\bPhi^\top \bPhi\right)^{-1}, \notag \\ 
     &\bbeta = \bSigma\left(\frac{\bmu}{\v} + \tau^{-1} \bPhi^\top \h\right),
\end{align}
where $\bmu = [\mu_1, \ldots, \mu_A]^\top$ and $\v = [v_1, \ldots, v_A]^\top$.

We optimize the approximation factor \eqref{eq:ep-approx} for every operator $j$. To this end, we first compute the calibrated distribution (STEP 1; see \eqref{eq:ep-step-1}), 
\begin{align}
	&q^{\bkh j}(s_j, w_j) \propto \frac{q(s_j, w_j)}{\text{Bern}(s_j|\sigma(\rho_j))\N(w_j|\mu_j, v_j)} =\text{Bern}(s_j|\sigma(\rho^{\bkh j})) \N(w_j|\mu^{\bkh j}, v^{\bkh j}), 
\end{align}
where according to the exponential family properties, we have
\begin{align}
	\rho^{\bkh j} &= 	\widehat{\rho}_j - \rho_j, \notag \\
	\left(v^{\bkh j}\right)^{-1} &= [\bSigma]_{jj}^{-1} - (v_j)^{-1}, \notag \\
	\left(v^{\bkh j}\right)^{-1} \mu^{\bkh j} &= \frac{\beta_j}{[\bSigma]_{jj}} - \frac{\mu_j}{v_j}.
\end{align}
Note that  $\beta_j$ and $[\bSigma]_{jj}$ are the marginal mean and variance of $w_j$, respectively, under the global posterior approximation $q(\s, \w)$.  

In STEP 2 (see \eqref{eq:ep-step-2}), we construct the titled distribution, 
\begin{align}
	&\tp(s_j, w_j) \notag \\
    &\propto q^{\bkh j}(s_j, w_j) \cdot \left(s_j \N(w_j|0, \sigma^2_0) + (1-s_j)\delta(w_j) \right) \notag \\
	&=\text{Bern}(s_j|\sigma(\rho^{\bkh j})) \N(w_j|\mu^{\bkh j}, v^{\bkh j}) \left(s_j \N(w_j|0, \sigma^2_0) + (1-s_j)\delta(w_j) \right). \label{eq:tilt}
\end{align}

In STEP 3 (see \eqref{eq:ep-step-3}), we perform moment matching. That is, we compute the moment of $s_j$ and $w_j$ under $\tp$. To this end, 
we use the following identity about the convolution of two Gaussians, 
\[
\N(\btheta|\m_1,\Sigma_1)\N(\btheta|\m_2,\Sigma_2) = \N(\btheta|\hat{\m},\hat{\Sigma})\N(\m_1|\m_2,\Sigma_1+\Sigma_2)
\]
where
\[
\hat{\Sigma} = \big( \Sigma_1^{-1} + \Sigma_2^{-1} \big)^{-1},\quad \hat{\m} = \hat{\Sigma}(\Sigma_1^{-1}\m_1 + \Sigma_2^{-1}\m_2).
\]

It is therefore straightforward to obtain  the marginal tilted distribution,
\begin{align}
	&\tp(s_j) \propto  \left(\sigma(\rho^{\bkh j})\N(\mu^{\bkh j}|0, v^{\bkh j} + \sigma_0^2)\right) s_j + \left((1-\sigma(\rho^{\bkh j}))\N(\mu^{\bkh j}|0, v^{\bkh j})\right)(1-s_j). \label{eq:ps}
\end{align}
We need to normalize the coefficients to get the probabilities, so we have 
\begin{align}
	\tp(s_j) = \text{Bern}(s_j|\rho^*), \label{eq: s-update}
\end{align}
where 
\begin{align}
	\rho^* = \rho^{\bkh j} + \log \frac{\N(\mu^{\bkh j}|0, v^{\bkh j} + \sigma_0^2)}{\N(\mu^{\bkh j}|0, v^{\bkh j})}.
\end{align}

Now, we need to compute the first and second moments of $w_j$ in \eqref{eq:tilt}. We first marginalize out $s_j$, 
\begin{align}
	&\tp(w_j) = \frac{1}{Z} \left(\sigma(\rho^{\bkh j}) \N(w_j|\mu^{\bkh j}, v^{\bkh j})\N(w_j|0, \sigma^2_0) 
    + (1-\sigma(\rho^{\bkh j}))\N(w_j|\mu^{\bkh j}, v^{\bkh j}) \delta(w_j)\right).
\end{align}
Look at the normalizer, it is the same as the one in \eqref{eq:ps}. Then we can obtain the analytical form, \begin{align}
	\tp(w_j) = \sigma(\rho^*)\N(w_j|\tmu, \tv) + \left(1 - \sigma(\rho^*)\right) \frac{\N(w_j|\mu^{\bkh j}, v^{\bkh j})}{\N(\mu^{\bkh j}|0, v^{\bkh j})}\delta(w_j), \label{eq:pv}
\end{align}
where 
\begin{align}
	\tv^{-1} &= {v^{\bkh j}}^{-1} + \sigma_0^{-2}, \notag \\
	\tv^{-1}\tmu &=  \frac{\mu^{\bkh j}}{v^{\bkh j}}.
\end{align}

Now, we can compute the moments of \eqref{eq:pv}, which is straightforward: 
\begin{align}
	\mu^{*}_j \overset{\Delta}{=}\EE_{\tp}[w_j] &= \sigma(\rho^*) \tmu, \notag \\
	\EE_{\tp}[\left(w_j\right)^2] &= \sigma(\rho^*)\left(\tv + \tmu^2 \right), \notag \\
	v^{*}_j \overset{\Delta}{=} \text{cov}_{\tp}(w_j) &= \sigma(\rho^*)\left(\tv + (1 - \sigma(\rho^*))\tmu^2\right).\label{eq: v-update}
\end{align}
We obtain the global posterior approximation, 
\begin{align}
	q^*(s_j, w_j) = \text{Bern}(s_j|\sigma(\rho^*))\N(w_j|\mu_j^*, v_j^*).
\end{align}
The approximation factor in \eqref{eq:ep-approx} is updated by $\frac{q^*(s_j, w_j)}{q^{\bkh j}(s_j, w_j)}$ (STEP 4; see \eqref{eq:ep-step-4}), which gives
\begin{align}
	\rho_j &= \rho^* - \rho^{\bkh j}, \notag \\
	\left(v_j\right)^{-1} &= \left(v^{*}_j\right)^{-1} - \left(v^{\bkh j}\right)^{-1}, \notag \\
	\frac{\mu_j}{v_j} &= \frac{\mu^{*}_j}{v^{*}_j} - \frac{\mu^{\bkh j}}{v^{\bkh j}}.
\end{align}

In each iteration, we in parallel update the approximation factor \eqref{eq:ep-approx} for every operator $j$. Initially, we set $\rho_j = 0$, $\mu_j = 0$ and $v_j = 10^6$ (so at the beginning, these factors are uninformative and nearly constant one). We repeat the EP iterations until convergence, and then return the posterior approximation \eqref{eq:post-ss-appendix}.

\subsection{\MakeUppercase{Experimental Details}}\label{sect:expr-setting}
\noindent\textbf{SINDy}. We used the PySINDy library (\url{https://pysindy.readthedocs.io/en/latest/index.html#id4})  in the experiment. PySINDy not only includes the original implementation of SINDy with sequential threshold ridge regression (STRidge), but it also supports other sparse promoting techniques, including L0 and L1 thresholding. The current library includes six optimizers: STLSQ, SR3-L0, SR3-L1, SSR, SSR-residual, and FROLS. We tried all the six optimizers and tuned the hyperparmeters to achieve the best performance with our best efforts. Specifically, we tuned the tolerance level in the range $[10^{-15}, 10^{-1}]$, regularization strength (for L0, L1 and L2) from range $[10^{-1}, 10^3]$, normalization (on/off), the max number of iterations from range $[10^4, 10^5]$, and  threshold for STLSQ from the range [$10^{-5}$, $0.5$]. 

\noindent{\textbf{PINN-SR}.} We used the implementation (\url{https://github.com/isds-neu/EQDiscovery}) of the authors. PINN-SR first performs a pre-training of the NN from the training data, and then conducts alternating direction optimization (ADO) for joint equation discovery and solution learning. We tuned the number of NN layers from the range $[3,8]$, the NN width from range $[20, 100]$,  the number of iterations for ADO from range $[12, 20]$, regularization strength from $[10^{-7}, 10^{-1}]$, L-BFGS pre-training iterations from $[10\text{K}, 80\text{K}]$, L-BFGS ADO training iterations from [1K, 2K], learning rate of ADAM from $[10^{-4}, 10^{-3}]$, and d\_tol hyper-parameter from $[10^{-5}, 5]$. 
We varied the number of collocation points from $[10\text{K}, 160\text{K}]$.

%number of hidden layers [3, 8], 2. number of LBFGS pretraining iterations [10000, 80000] 3. number of ADO training, [12, 20] 4. regualization stress [1e-1, 1e-7] 5. d_tol [1e-1, 5]

%For each experiment, we ran 30 alternating direction optimization (ADO) iterations in the outer-loop, which is enough for convergence. We used the default parameters in the inner loop, namely, alternately updating the PINN parameters with ADAM/L-BFGS and the operator coefficients with ridge regression and thresholding.  PINN-SR first conducted pre-training of the neural network (NN) using L-BFGS with 1K maximum iterations and the machine precision as the tolerance level, and then entered into the ADO iterations. We used the \texttt{tanh} activation for all the experiments. To select the NN architecture, we tuned the depth from [2, 9] and width from [20, 100], to run PINN-SR and report the results for the architecture that gave the smallest training loss for each experiment. We found the architecture is critical to identify the equations. For example, for the nonlinear pendulum system, we were never able to correctly recover the equation using NNs more than 4 layers. A two layer NN with 20 or 50 neurons per layer often gave the best estimations (but not necessarily the exactly correct). By contrast, for the nonlinear diffusion-reaction system, the most reasonable equations are found by NNs with more than 6 layers with width 50. Note that best architecture was also subject to different runs (\ie collocation points).     

\noindent\textbf{BSL}. We used the original implementation shared by the authors. We tuned the number of knots from range [30, 300] for each input dimension. We tuned the number of collocation points per dimension from [50, 300]. The weight value threshold was tuned from range $[10^{-5}, 0.5]$, and the regularization strength from range $[10^{-6}, 10^{-1}]$. We also tuned the pre-training iteration number from $[10^4, 10^5]$ and the number of ADO iterations from $[8, 20]$. The d\_tol hyper-parmeter was tuned from $[10^{-5}, 0.5]$. 

\noindent\textbf{\ours}. We tuned  the slab variance $\sigma_0^2$ from 0.5 to 5000 (very flat Gaussian slab),  and the equation likelihood variance $\tau$ from $10^{-4}$ to $1$. The data likelihood variance $v$ in \eqref{eq:data-ll} (of the main paper) is obtained via cross-validation in the initial training (see the first step in Algorithm \ref{algo:ep-em}). In the early iterations, we employed small values of $\alpha$, typically around $0.1$ or the smallest value of the current $q(s_j=1)$ among all $j$, to conservatively prune the operators. As the iteration progresses towards completion, we increased $\alpha$ to $0.5$ to conduct an unbiased pruning over the remaining operators, considering that our solution estimate is as optimal as possible. The performance is not sensitive to other hyperparameters and so we fixed them. Specifically, 
the maximum number of EP-EM iteration was set to 200K and the tolerance level is $10^{-8}$. We also used relative change of natural parameters less than $10^{-5}$ as the condition to stop the EP inference at the E step. The solution estimation $\Ucal$ was initialized as zero. We used an evenly-spaced mesh for each testing case, and the size is summarized in Table \ref{tb:mesh-size}.
\begin{table*}[h]
	%\begin{wraptable}{r}{0.7\textwidth}
	\vspace{-0.01in}
	\small
	\centering
	\begin{tabular}[c]{cc}
		\toprule
	    Test equation & Mesh size\\
		\hline 
        VDP & 200 \\
        Lorenz 96 & 300 \\
        Burger's ($\nu=0.1$) & $160 \times 160$\\
        Burger's ($\nu = 0.01$) & $180 \times 180$\\
        Burger's ($\nu = 0.005$) & $180 \times 180$\\
        KS & $200 \times 200$\\
        Allen-cahn &$200 \times 200$\\
        Predator-Prey & 500\\
		\bottomrule
	\end{tabular}
	\caption{\small The mesh size used by \ours.} \label{tb:mesh-size}
\end{table*}

\subsection{\MakeUppercase{Real-world Predator-Prey Data}} \label{sect:real-world}
\begin{figure*}[h]
			%	\vspace{-0.05in}
				\centering
				\setlength{\tabcolsep}{0pt}
				\begin{tabular}[c]{cc}
					\begin{subfigure}[b]{0.4\textwidth}
						\centering
						\includegraphics[width=\linewidth]{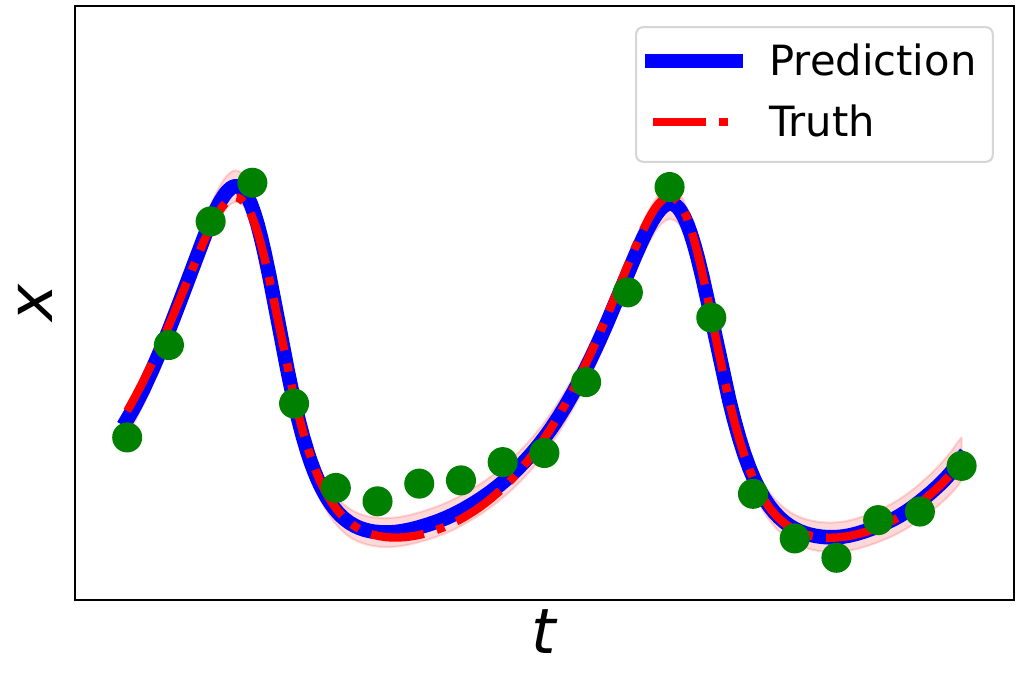}
						%\caption{30 * std}
					\end{subfigure} &
					\begin{subfigure}[b]{0.4\textwidth}
						\centering
						\includegraphics[width=\linewidth]{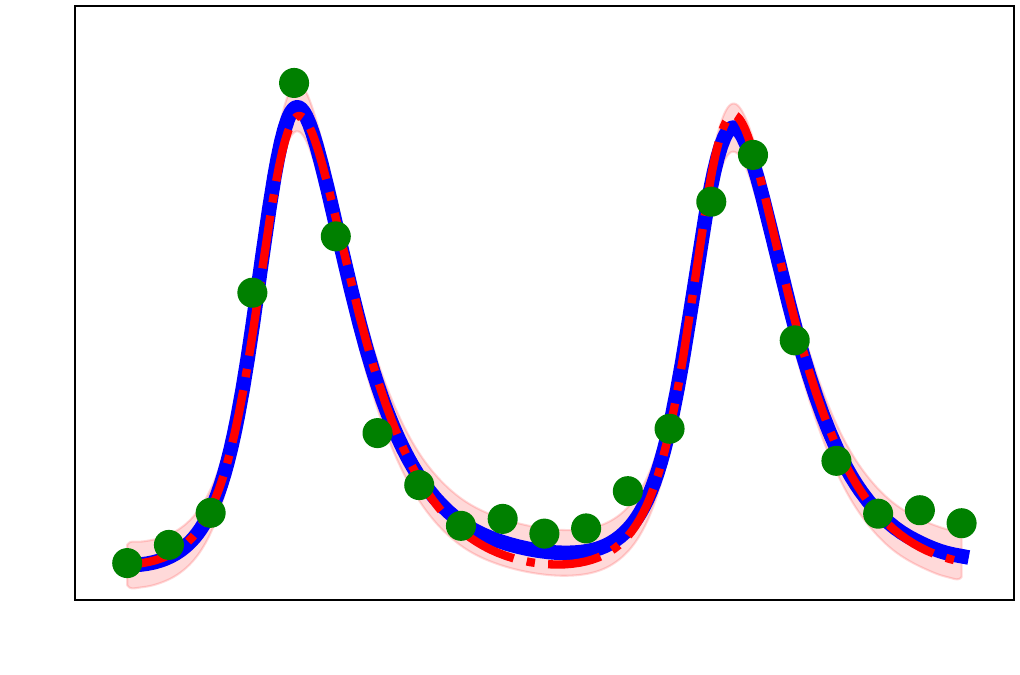}
						%\label{fig:poisson1d-mix_sin-GP-HM-SM}
					\end{subfigure}
				\end{tabular}
				\caption{\small Solution estimate for the real-world predator-prey system.}
				\label{fig:sol-predatory-prey}
\end{figure*}

We followed~\citep{sun2022bayesian} to test the equation discovery from a real-world dataset collected from a predator-prey system between lynx and hares. The dataset is the population of the lynx and hares from 1900 to 1920 at Hudson Bay Company, presented in Table 11 of~\citep{sun2022bayesian}. The dataset is very noisy. The governing equation from mathematical analysis is given by
\begin{align}
\frac{\d x}{\d t} &= 0.4807x - 0.0248xy, \notag \\
\frac{\d y}{\d t} &= -0.9272 y + 0.0276xy,
\end{align}
which we used as the golden standard. We used the same setting as in \citep{sun2022bayesian} to test \ours. The performance of the discovery is reported in Table \ref{tb:real-world}. As we can see, \ours not only correctly discovered the equation form, but also it achieves the smallest RMSE in the operator weight estimation. We also show the solution prediction of \ours in Fig.~ \ref{fig:sol-predatory-prey}.

\begin{table*}[t]
	%\begin{wraptable}{r}{0.7\textwidth}
	\small
	\centering
	\begin{tabular}[c]{cccc}
		\toprule
		\multirow{2}{*}{\textit{Method}} & \multicolumn{3}{c}{21 examples}\\
            & RMSE & Recall & Precision \\
		\hline 
        SINDy & F & 0.75 & 0.6 \\
        PINN-SR & F & 0.25 & 0.5 \\
        BSL & 0.0304 & 1 & 1\\
        \ours & \textbf{0.0244} & \textbf{1} & \textbf{1} \\
		\bottomrule
	\end{tabular}
	\caption{\small Performance of discovering a real-world predator-prey system. } \label{tb:real-world}
\end{table*}

\subsection{\MakeUppercase{Laplace's approximation for Posterior Estimation}} \label{sect:laplace}
Given the loss $L$ and parameters $\bbeta$,  the Laplace's approximation~\citep{walker1969asymptotic,mackay2003information} first minimizes the loss function to obtain the optimal parameter estimate $\bbeta^*$, and then constructs a multi-variate Gaussian posterior approximation, 
\begin{align}
q(\bbeta) = \N(\bbeta|\bbeta^*, \H^{-1}),
\end{align}
where 
\begin{align}
\H = \left.\frac{\partial^2 L}{\partial \bbeta^2} \right\vert_{\bbeta = \bbeta^*}
\end{align}
is the Hessian matrix at $\bbeta^*$. The approximation can be derived via the second-order Taylor approximation of $L$ at the optimum $\bbeta^*$. For the ODE systems, we applied the standard Laplace's approximation to obtain the posterior estimate of our solution prediction $\Ucal$. For the other equations, like Burger's and KS equations, since $\vec(\Ucal)$ is high-dimensional, computing the full Hessian matrix is very costly. Therefore, we use the block-diagonal Hessian~\citep{ritter2018scalable}. We compute the Hessian for each time slice and then approximate the posterior for the solution at that slice. 

\begin{figure*}
	\centering
	\setlength{\tabcolsep}{0pt}
	\begin{tabular}[c]{cccc}
	    \begin{subfigure}[b]{0.28\textwidth}
        \includegraphics[width=\textwidth]{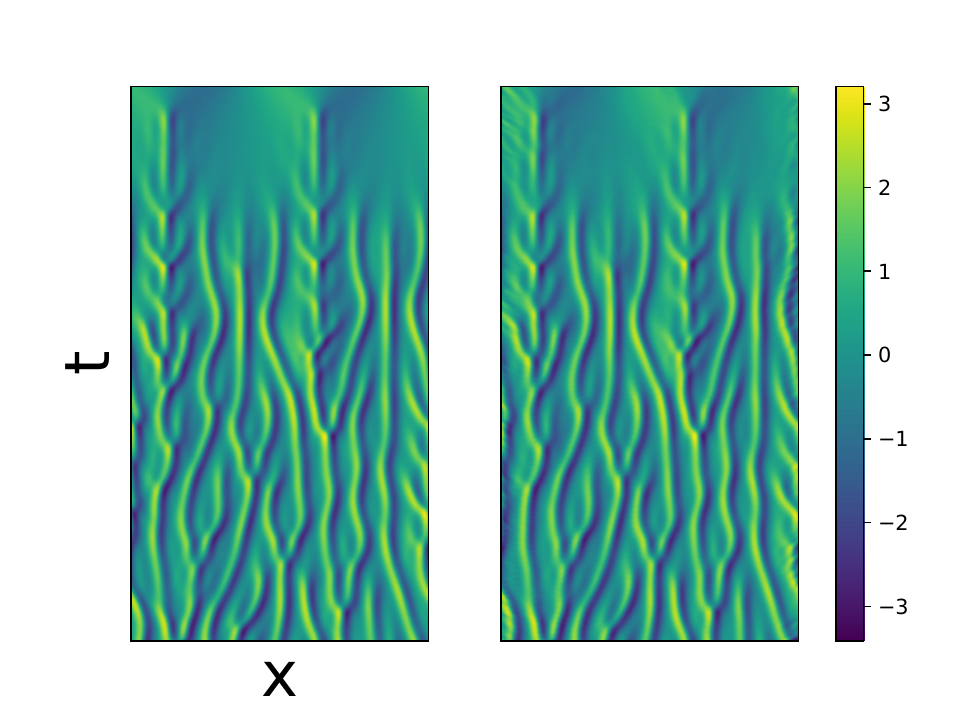}
        \caption{Left: truth; right: prediction}
        \label{fig:ks-whole}
    \end{subfigure} &
    \begin{subfigure}[b]{0.22\textwidth}
         \includegraphics[width=\textwidth]{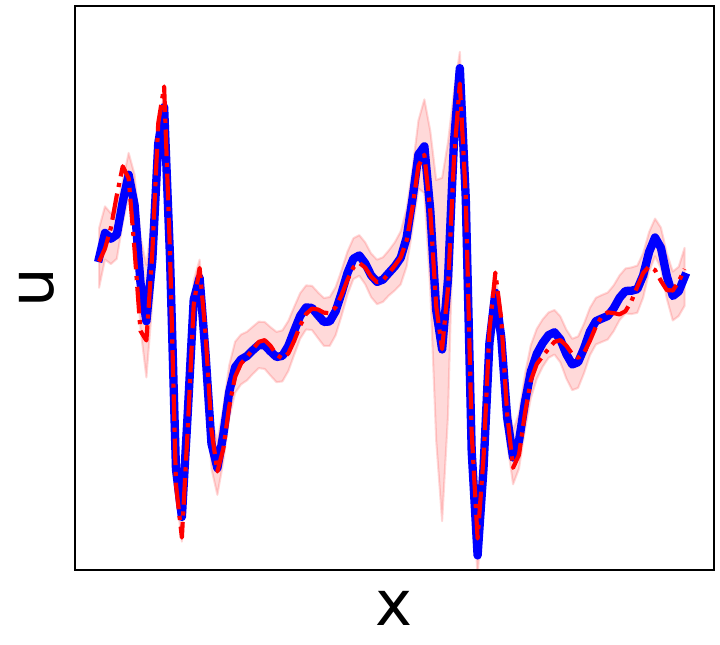}
            \caption{t = 24.1}
            \label{fig:ks-slice-1}
    \end{subfigure} &
	\begin{subfigure}[b]{0.22\textwidth}
        \includegraphics[width=\textwidth]{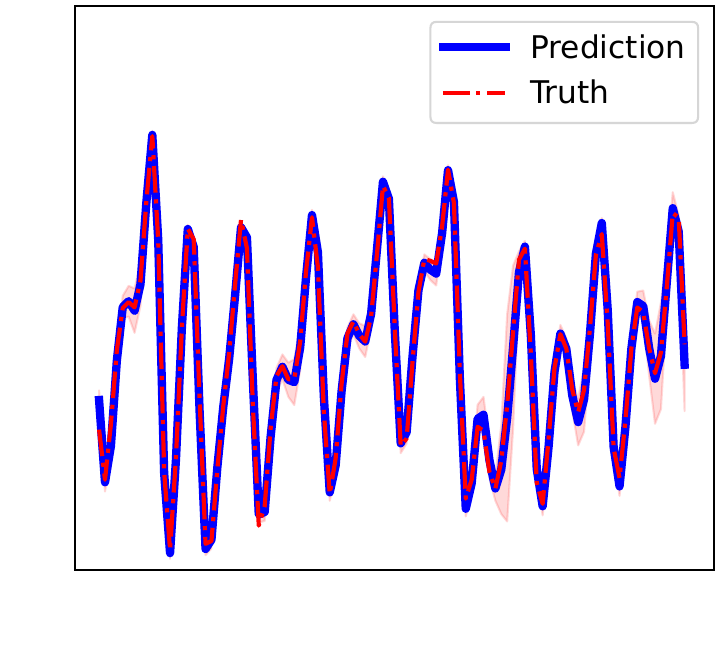}
        \caption{t = 75.4 \cmt{, 20 * std}}
         \label{fig:ks-slice-2}
     \end{subfigure} &
    \begin{subfigure}[b]{0.22\textwidth}
         \includegraphics[width=\textwidth]{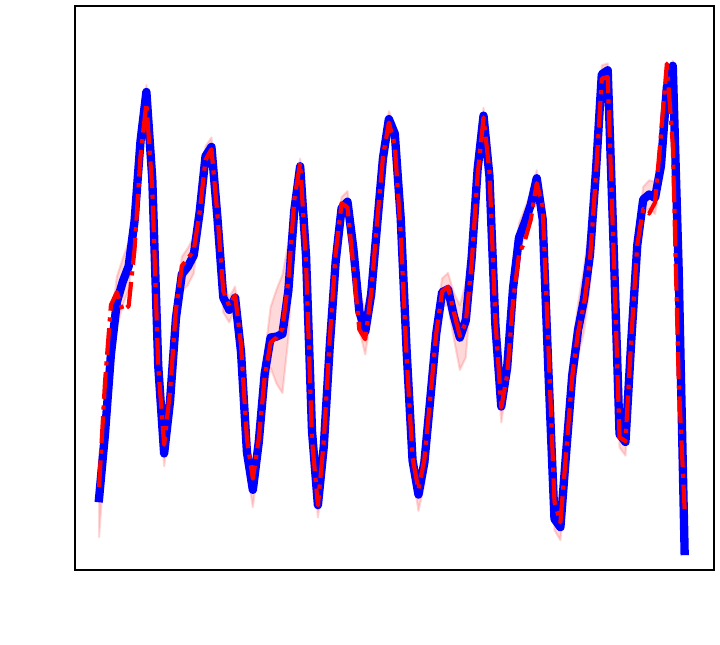}
          \caption{t = 128.1}
    \label{fig:ks-slice-3}
  \end{subfigure}
	\end{tabular}
	\caption{\small Solution estimate for the KS equation with 20\% noise on training data.}
	\label{fig:ks-est}
\end{figure*}

\cmt{
\subsection{More Results about Solution and Weight Posterior Estimation} \label{sect:more-results}
In Fig. \ref{fig:ks-est}, we show the solution estimate for the KS equation with 20\% noise on the training dataset. In Fig. \ref{fig:ks-whole}, we show the prediction on the entire domain. In Fig. \ref{fig:ks-slice-1}, \ref{fig:ks-slice-2} and \ref{fig:ks-slice-3}, we show the solution estimation at three time slices. %As we can see, the solution prediction is quite accurately. 
\zhec{to do: add a few more weight posteriors, including those for VDP, Lorenz, and KS}
}

\begin{figure*}[t]
			%	\vspace{-0.05in}
				\centering
				\setlength{\tabcolsep}{0pt}
				\begin{tabular}[c]{ccc}
					\begin{subfigure}[b]{0.3\textwidth}
						\centering
						\includegraphics[width=\linewidth]{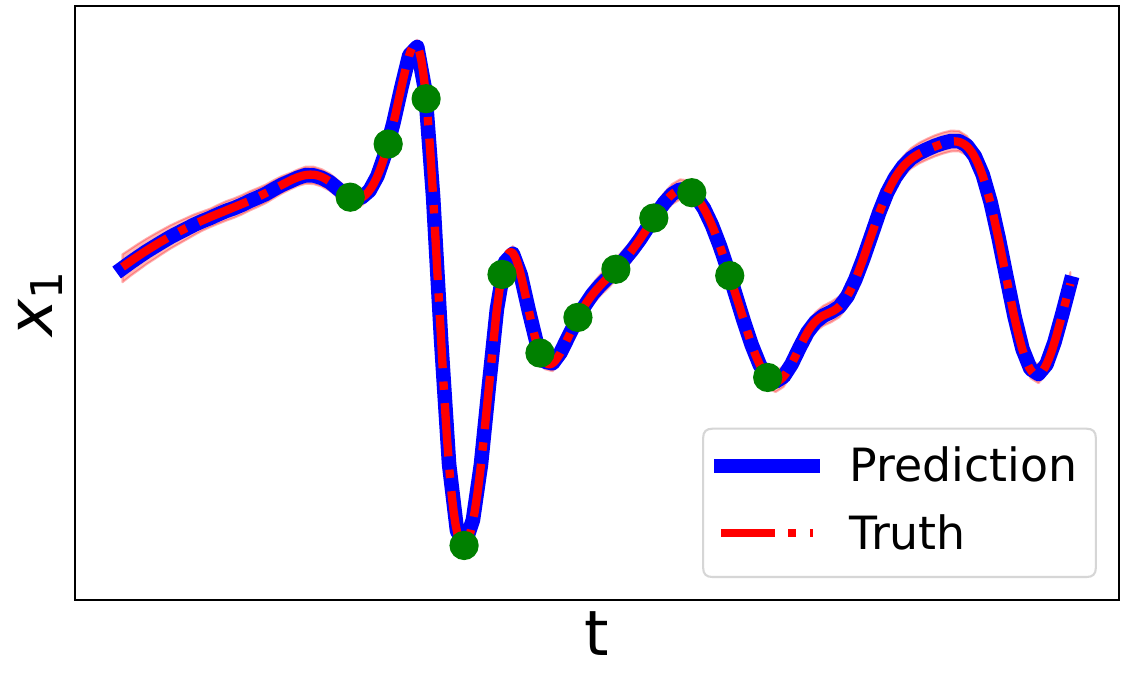}
						%\caption{30 * std}
					\end{subfigure} &
					\begin{subfigure}[b]{0.3\textwidth}
						\centering
						\includegraphics[width=\linewidth]{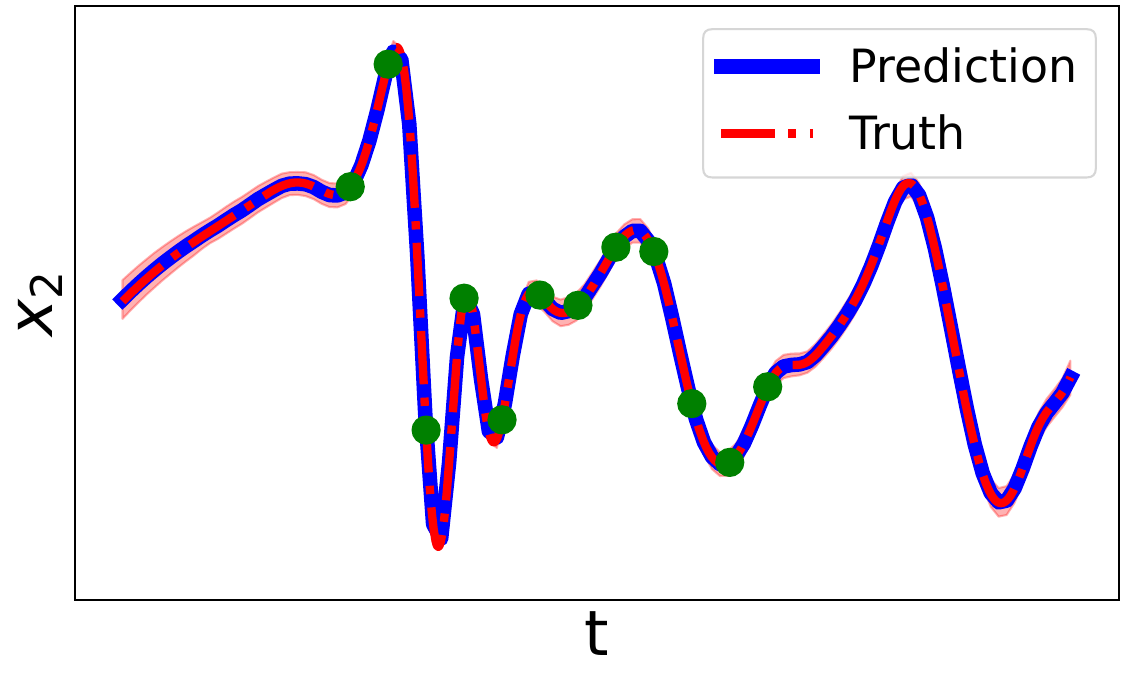}
						%\label{fig:poisson1d-mix_sin-GP-HM-SM}
					\end{subfigure} &
				\begin{subfigure}[b]{0.3\textwidth}
					\centering
					\includegraphics[width=\linewidth]{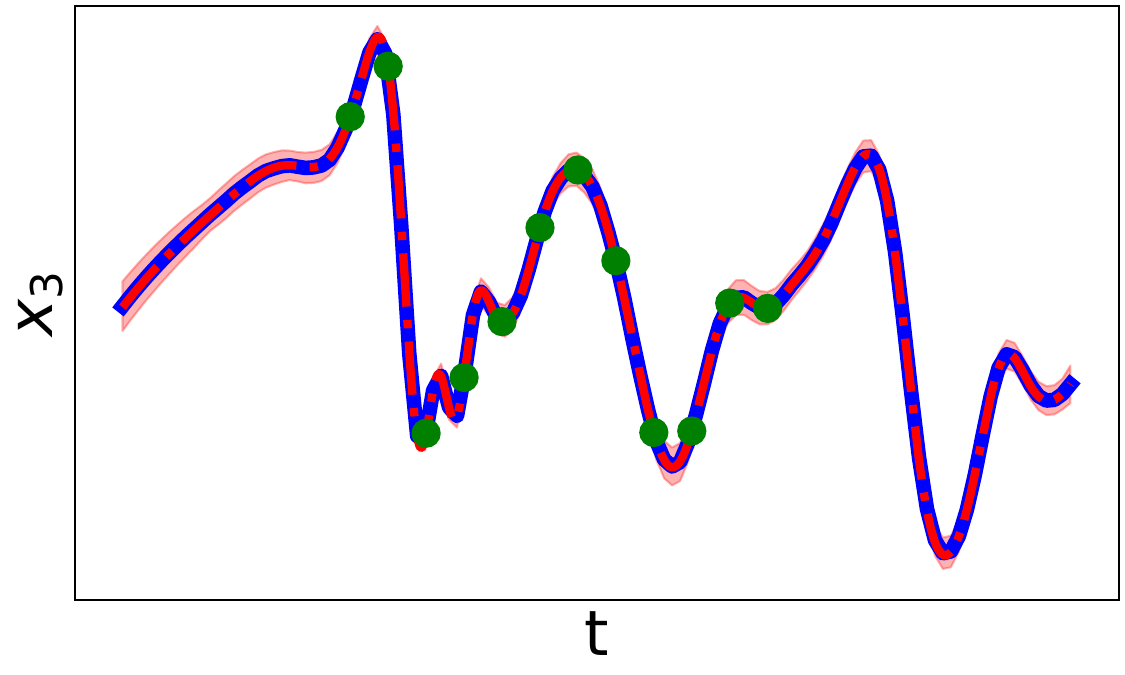}
					%\caption{$q(w_1)$}
					%\label{fig:poisson1d-mix_sin-GP-SE}
				\end{subfigure} \\
				\begin{subfigure}[b]{0.3\textwidth}
					\centering
					\includegraphics[width=\linewidth]{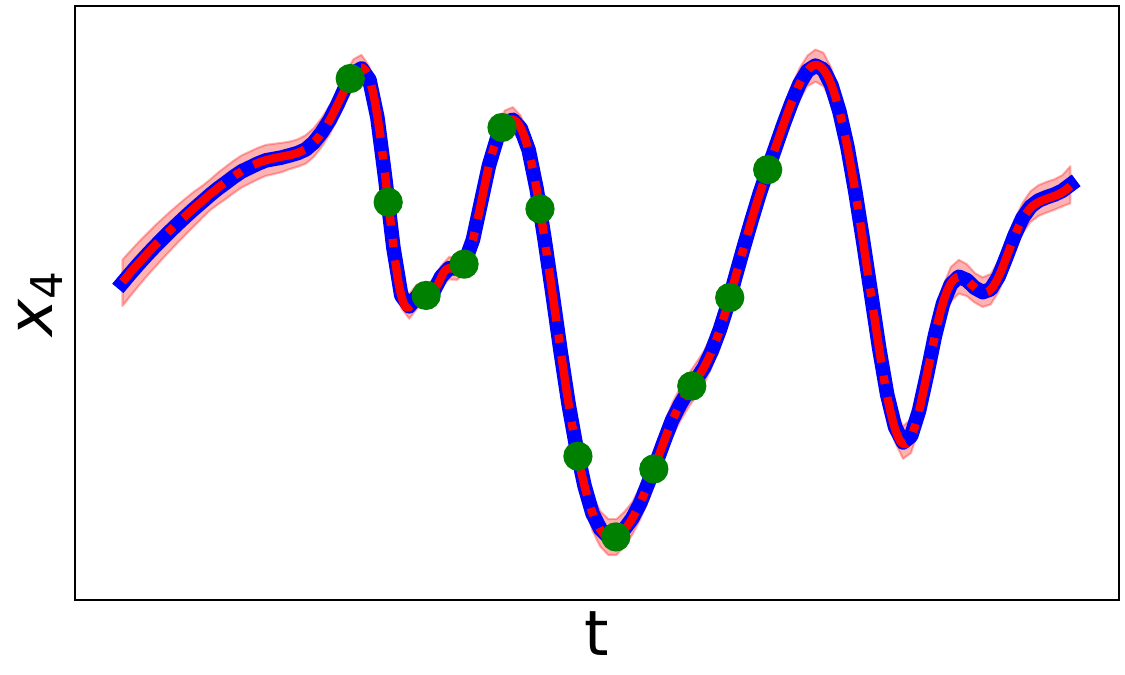}
					%\caption{$q(w_2)$}
                    \caption{}
					%\label{fig:poisson1d-mix_sin-GP-Matern}
				\end{subfigure} &
                    \begin{subfigure}[b]{0.3\textwidth}
					\centering
					\includegraphics[width=\linewidth]{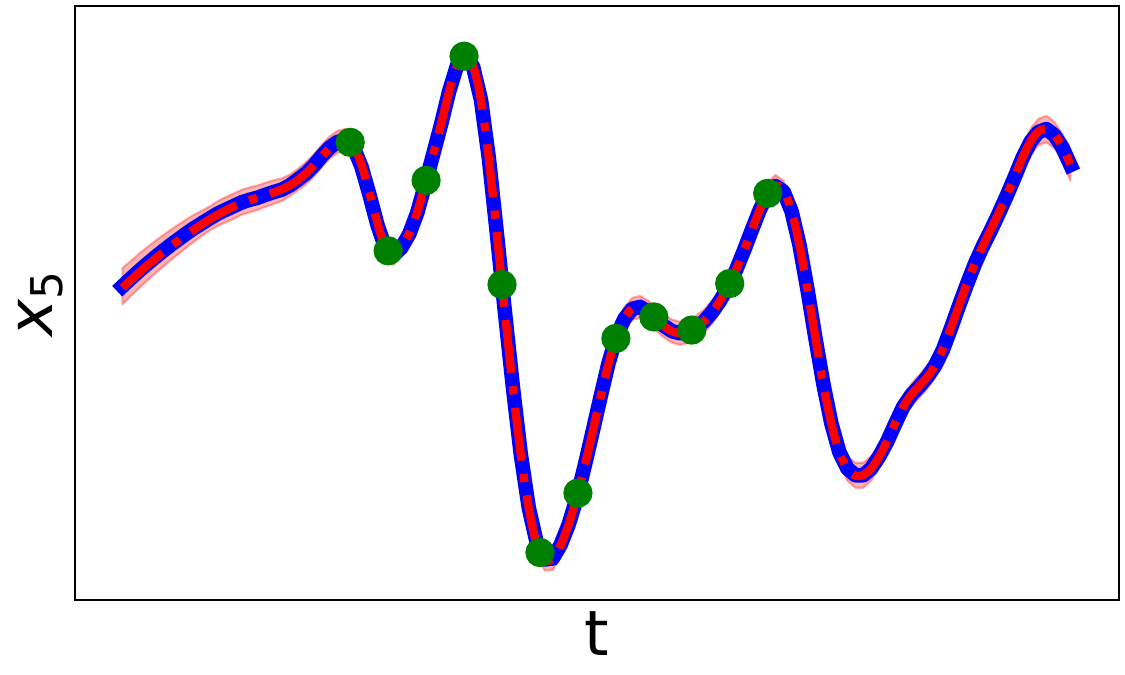}
					%\caption{$q(w_2)$}
                    \caption{}
					%\label{fig:poisson1d-mix_sin-GP-Matern}
				\end{subfigure} &
                    \begin{subfigure}[b]{0.3\textwidth}
					\centering
					\includegraphics[width=\linewidth]{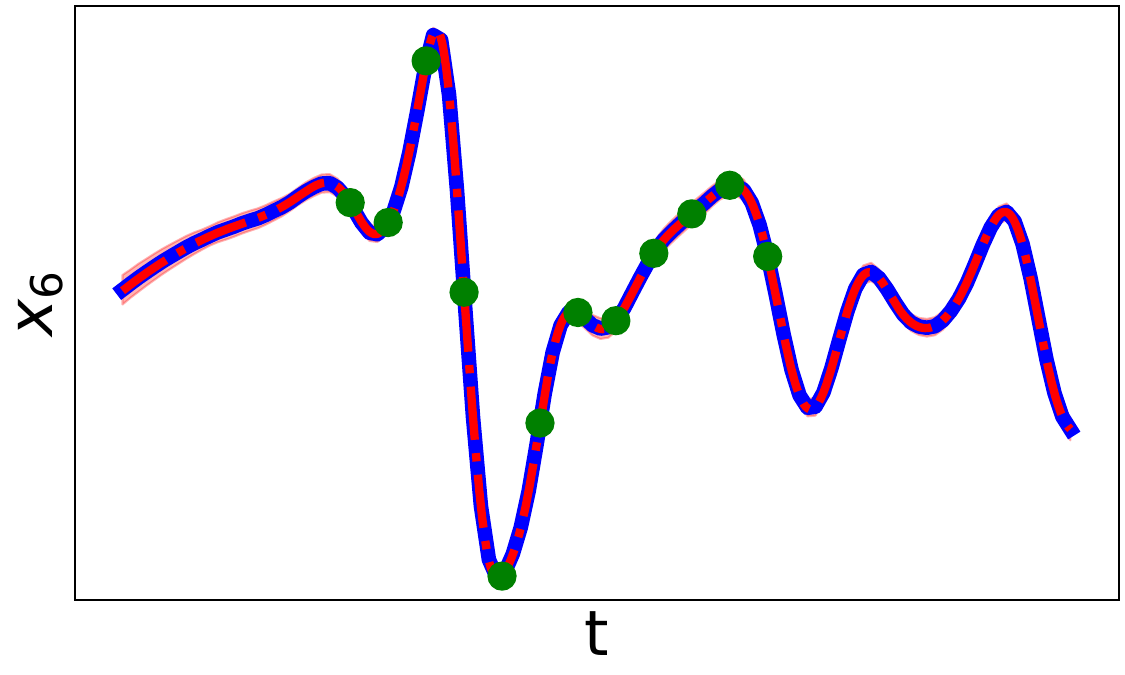}
					%\caption{$q(w_2)$}
                    \caption{}
					%\label{fig:poisson1d-mix_sin-GP-Matern}
				\end{subfigure} 
    %             \begin{subfigure}[b]{0.2\textwidth}
				% 	\centering
				% 	\includegraphics[width=\linewidth]{./NeurIPS2023/fig/osc/os_w4.pdf}
				% 	\caption{$q(w_4)$}
				% 	%\label{fig:poisson1d-mix_sin-GP-Matern}
				% \end{subfigure} 
				\end{tabular}
				\caption{\small Solution estimate for the Lorenz 96 system using 12 training examples.}
				\label{fig:solution-lorenz}
\end{figure*}
\subsection{\MakeUppercase{Running Time}}\label{sect:run-time}
We examined the running time of each method on a Linux workstation with a NVIDIA GeForce RTX 3090 GPU, with a memory size of 24576 MB. We tested on the discovery of VDP, Lorenz 96, Burger's and KS equations. The results are reported in Table \ref{tb:run-time}. It can be seen that \ours is consistently faster than BSL: 4.2x, 1.9x, 7.3x and 2.7x faster in the four testing cases, respectively. Although PINN-SR is faster than \ours in VDP and Lorenz 96, it failed to discover the correct equations. For KS, PINN-SR used 19.5x time and but still failed to recover the equation. SINDy is the fastest among all the four methods. It is reasonable, because SINDy does not estimate the solution function, and only performs sparse linear regression. 

\begin{table*}[h]
	%\begin{wraptable}{r}{0.7\textwidth}
	\small
	\centering
	\begin{tabular}[c]{ccc}
		\toprule
	    Test case & Number of examples & RMSE\\
		\hline 
        VDP & 54 & 0.127\\
        Lorenz 96 &250  & 0.229\\
        Burger's ($\nu=0.1$) & $40 \times 40$ & 0.0398 \\
        KS & $512 \times 751$ & 0.0173\\
		\bottomrule
	\end{tabular}
	\caption{\small SINDy with more training data for successful discovery. Note that the data is noise free.} \label{tb:sindy-more}
\end{table*}
\begin{table*}[h]
	\small
	\centering
	\begin{tabular}[c]{ccccc}
		\toprule
	    Test case & \ours& BSL& PINN-SR &SINDy\\
		\hline 
        VDP & 643& 2700& 421 (F)& 0.5 (F)\\
        Lorenz 96 &3994  & 7438& 2086 (F)& 2 (F)\\
        Burger's ($\nu=0.1$) &99 &720 & 853& 2 (F)\\
        KS &1174 & 3150 (F) & 22934 (F)& 0.5 (F)\\
		\bottomrule
	\end{tabular}
	\caption{\small Running time (in seconds) on a Linux workstation with a NVIDIA GeForce RTX 3090 GPU with 24576 MB memory. The number of training examples used for each test case is 25, 50, $20\times 20$ and $40\times 40$ for VDP, Lorenz 96, Burger's ($\nu=0.1$) and KS, respectively. The noise level is zero. ``F'' means failed to discover the equation. } \label{tb:run-time}
\end{table*}

\subsection{\MakeUppercase{Discovered Equations}} \label{sect:discovered-eq}
We show the examples of discovered equations by \ours and the competing methods in Table \ref{tb:discover-vdp}, \ref{tb:discover-lorenz}, \ref{tb:discover-burgers}, \ref{tb:discover-burgers2}, \ref{tb:discover-burgers3}, \ref{tb:discover-ks}, \ref{tb:discover-allen-cahn} and \ref{tb:discover-predator-prey}.

%\section{\zhec{MW's figure}}

\cmt{
\begin{figure*}
    \centering
    \begin{subfigure}[b]{0.22\textwidth}
         \includegraphics[width=\textwidth]{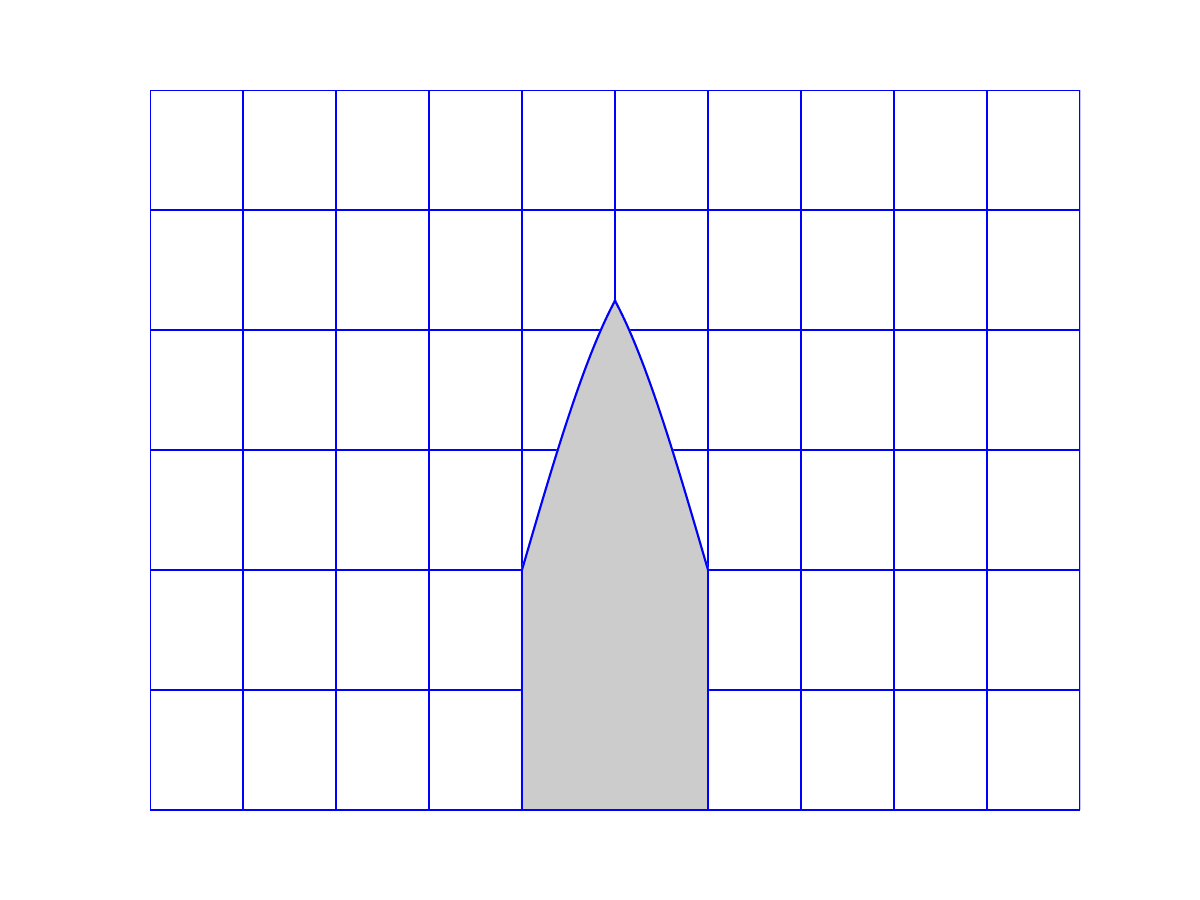}
            \caption{}
            \label{}
    \end{subfigure} 
     \begin{subfigure}[b]{0.22\textwidth}
         \includegraphics[width=\textwidth]{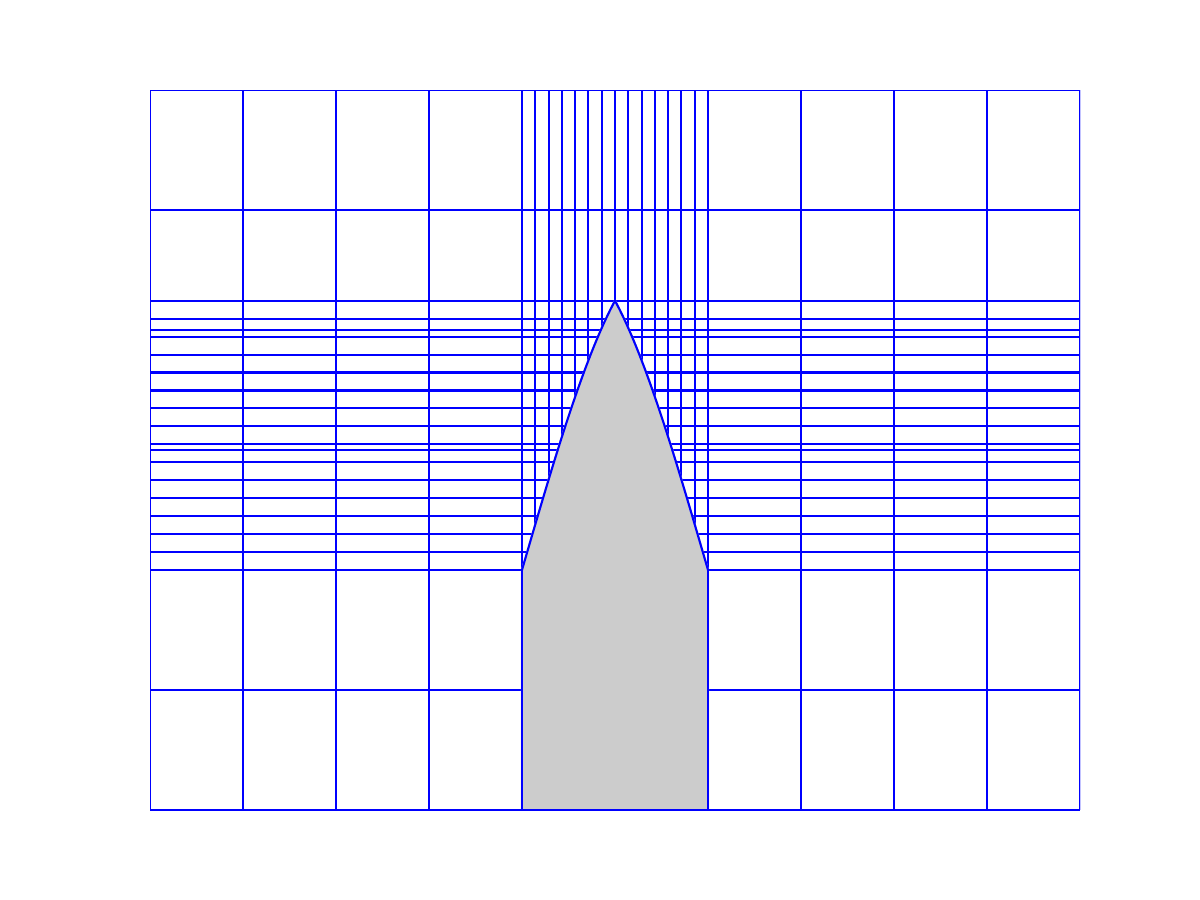}
            \caption{}
            \label{}
    \end{subfigure} 
    \caption{An example of the mesh design.} \label{fig:mesh-design}
\end{figure*} 
}

\begin{figure*}
	\centering
	\setlength{\tabcolsep}{0pt}
	\begin{tabular}[c]{cc}
	    \begin{subfigure}[b]{0.52\textwidth}
        \includegraphics[width=\textwidth]{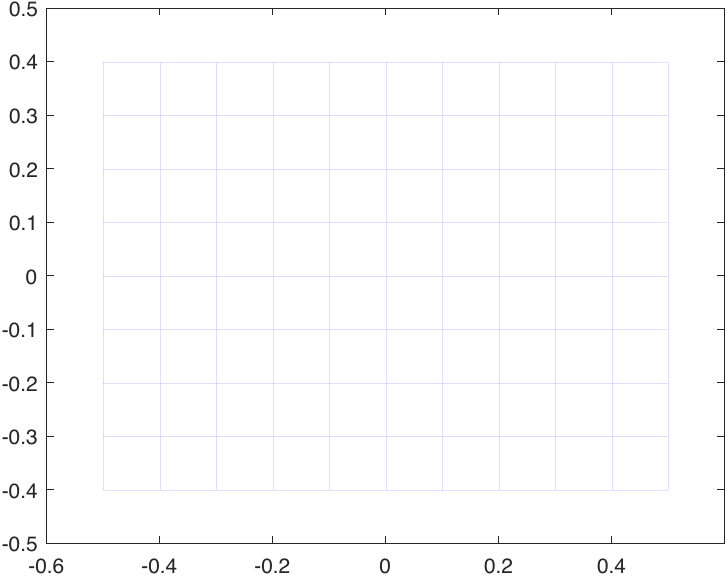}
        \caption{Regular mesh}
       % \label{fig:ks-whole}
    \end{subfigure} &
    \begin{subfigure}[b]{0.5\textwidth}
         \includegraphics[width=\textwidth]{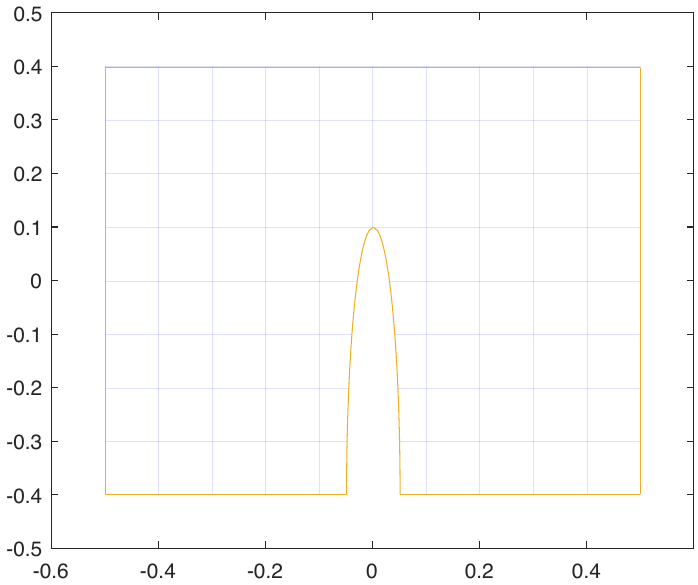}
            \caption{Regular mesh on a special domain}
            %\label{fig:ks-slice-1}
    \end{subfigure} \\
	\begin{subfigure}[b]{0.5\textwidth}
        \includegraphics[width=\textwidth]{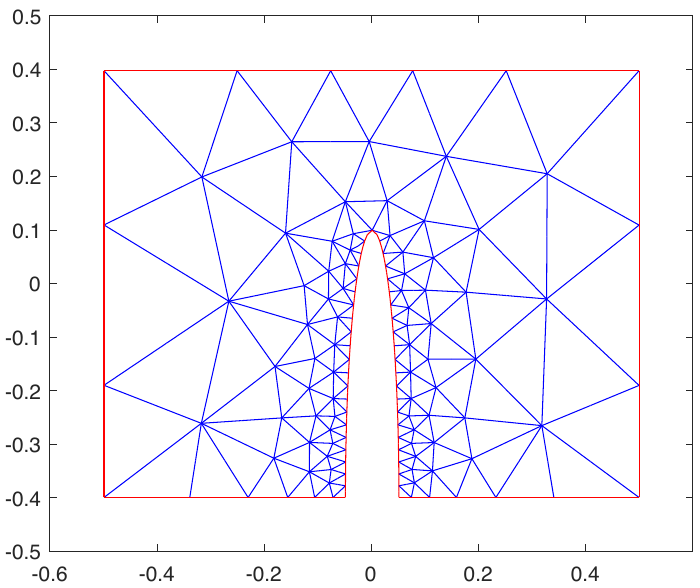}
        \caption{An example of finite element design.}
         %\label{fig:ks-slice-2}
     \end{subfigure} &
    \begin{subfigure}[b]{0.5\textwidth}
         \includegraphics[width=\textwidth]{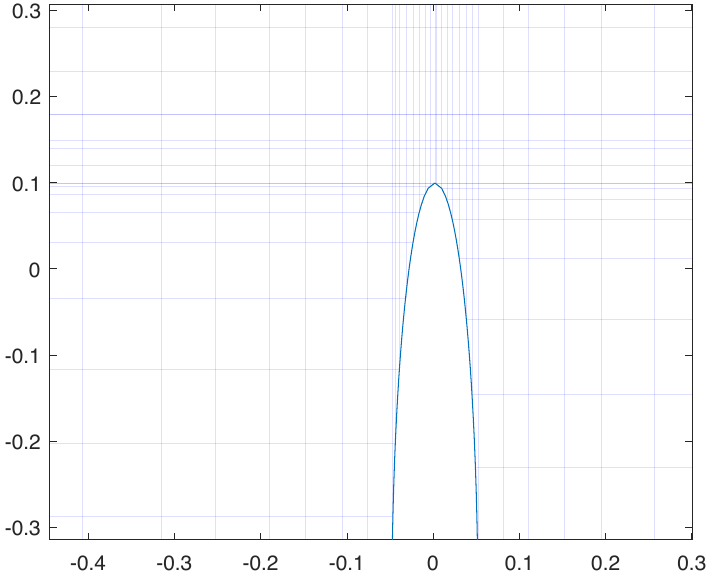}
          \caption{Unevenly-spaced mesh design for \ours.}
    %\label{fig:ks-slice-3}
  \end{subfigure}
	\end{tabular}
	\caption{\small An example of the mesh design. By varying the density regions of the mesh (d), our method can  adapt to the geometry of specific domains, as the traditional numerical methods do (c). }
	\label{fig:mesh-design}
\end{figure*}

\begin{table*}[h]
	\small
 \begin{subtable}{\textwidth}
 \centering
	\begin{tabular}[c]{cc}
		\toprule
	      \textit{Name}  & Equation\\
		\hline 
        True & $x_t=y$ \\
        \addlinespace
            & $y_t = 2.5y-x-2.5x^2y$\\
        \hline
        SINDy  & $x_t=0.807y$ \\
        \addlinespace
            & $y_t=-0.504x$\\
        \hline
        PINN-SR & $x_t=0.848y+1.23y^2+0.424y^3$ \\
        \addlinespace
            & $y_t = 1.14y-0.531x-0.742x^2y$\\
        \hline
        BSL & $x_t=1.236y$ \\
        \addlinespace
            & $y_t = -0.622x$\\
        \hline
        \ours & $x_t=y$ \\
        \addlinespace
            & $y_t = 2.47y-0.992x-2.49x^2y$\\
		\bottomrule
	\end{tabular}
	\caption{\small zero noise } \label{tb:mesh-szie}
 \end{subtable}
 \begin{subtable}{\textwidth}
     \centering
	\begin{tabular}[c]{cc}
		\toprule
	      \textit{Name}  & Equation\\
		\hline 
        True & $x_t=y$ \\
        \addlinespace
            & $y_t = 2.5y-x-2.5x^2y$\\
        \hline
        SINDy  & $x_t=0.724y$ \\
        \addlinespace
            & $y_t=-0.524x$\\
        \hline
        PINN-SR & $x_t=2.26y+0.208y^3-0.188y^4-0.303x^2+...$ \\
        \addlinespace
            & $y_t = 2.38y-1.5x^2y+1.52y^2-0.662y^4+...$\\
        \hline
        BSL & $x_t=0.505y+0.937xy-0.588x^2y+...$ \\
        \addlinespace
            & $y_t = 2.35xy-0.913x^2y-0.427x^3+...$\\
        \hline
        \ours & $x_t=0.961y$ \\
        \addlinespace
            & $y_t = 2.13y-1.04x-2.44x^2y$\\
		\bottomrule
        \hline
	\end{tabular}
	\caption{\small 20\% noise } \label{tb:mesh-szie}
 \end{subtable}
	
\caption{\small Discovery result for the VDP system with 10 measurement examples. }\label{tb:discover-vdp}
 
\end{table*}

\begin{table*}[h]
	%\begin{wraptable}{r}{0.7\textwidth}
	\vspace{-0.01in}
	\small
 \begin{subtable}{\textwidth}
 \centering
	\begin{tabular}[c]{cc}
		\toprule
	      \textit{Name}  & Equation\\
		\hline 
        True & ${x_{1}}_t=x_{2}x_{6}-x_5x_6-x_1+8$ , 
             ${x_{2}}_t=x_{1}x_{3}-x_1x_6-x_2+8$ \\
        \addlinespace
            & ${x_{3}}_t=x_{2}x_{4}-x_1x_2-x_3+8$ , ${x_{4}}_t=x_{3}x_{5}-x_2x_3-x_4+8$ \\
        \addlinespace
            & ${x_{5}}_t=x_{4}x_{6}-x_3x_4-x_5+8$ , ${x_{6}}_t=x_{1}x_{5}-x_4x_5-x_6+8$ \\
        \hline
        SINDy  & ${x_{1}}_t=0.688x_1+0.016x^2_4x_5-0.06x_3x^2_4$ , 
             ${x_{2}}_t=0.795x_3$ \\
        \addlinespace
            & ${x_{3}}_t=0$ , ${x_{4}}_t=0$ \\
        \addlinespace
            & ${x_{5}}_t=1.33$ , ${x_{6}}_t=1.38x_1-0.067x_3x^2_4$ \\
        \hline
        PINN-SR &${x_{1}}_t=-0.0288x_{1}x_{2}x_{6}$ , 
             ${x_{2}}_t=-0.0137{1}x_{2}x_{6}$ \\
        \addlinespace
            & ${x_{3}}_t=-0.0161x_{1}x_{2}x_{6}$ , ${x_{4}}_t=-0.015x_{1}x_{2}x_{6}$ \\
        \addlinespace
            & ${x_{5}}_t=-0.0117x_{1}x_{2}x_{6}$ , ${x_{6}}_t=-0.0117x_{1}x_{2}x_{6}$ \\
        \hline
        BSL & ${x_{1}}_t=0.944x_2$ , 
             ${x_{2}}_t=0$ \\
        \addlinespace
            & ${x_{3}}_t=-2.74-1.76x_1+1.24x_3+...$ , ${x_{4}}_t=6.35$ \\
        \addlinespace
            & ${x_{5}}_t=-1.8$ , ${x_{6}}_t=-4.52$ \\
        \hline
        \ours & ${x_{1}}_t=x_{2}x_{6}-x_5x_6-0.991x_1+7.96$ , 
             ${x_{2}}_t=0.999x_{1}x_{3}-0.999x_1x_6-x_2+7.94$ \\
        \addlinespace
            & ${x_{3}}_t=x_{2}x_{4}-x_1x_2-0.985x_3+7.92$ , ${x_{4}}_t=x_{3}x_{5}-x_2x_3-0.995x_4+7.98$ \\
        \addlinespace
            & ${x_{5}}_t=x_{4}x_{6}-x_3x_4-0.999x_5+7.97$ , ${x_{6}}_t=0.995x_{1}x_{5}-0.993x_4x_5-1.01x_6+7.93$ \\
		\bottomrule
	\end{tabular}
	\caption{\small zero noise } \label{tb:mesh-szie}
 \end{subtable}
 \begin{subtable}{\textwidth}
     \centering
	\begin{tabular}[c]{cc}
		\toprule
	      \textit{Name}  & Equation\\
		\hline 
        True & ${x_{1}}_t=x_{2}x_{6}-x_5x_6-x_1+8$ , 
             ${x_{2}}_t=x_{1}x_{3}-x_1x_6-x_2+8$ \\
        \addlinespace
            & ${x_{3}}_t=x_{2}x_{4}-x_1x_2-x_3+8$ , ${x_{4}}_t=x_{3}x_{5}-x_2x_3-x_4+8$ \\
        \addlinespace
            & ${x_{5}}_t=x_{4}x_{6}-x_3x_4-x_5+8$ , ${x_{6}}_t=x_{1}x_{5}-x_4x_5-x_6+8$\\
        \hline
        SINDy  & ${x_{1}}_t=0.169x_1+0.714x_2-0.053x_3x^2_4$ , 
             ${x_{2}}_t=0$ \\
        \addlinespace
            & ${x_{3}}_t=0$ , ${x_{4}}_t=0$ \\
        \addlinespace
            & ${x_{5}}_t=3.53-0.041x_2x^2_4$ , ${x_{6}}_t=-0.044x_3x^2_4$ \\
        \hline
        PINN-SR &${x_{1}}_t=0.587x_{1}x_{4}-2.85x_{2}x_{4}-0.206x_{1}x_{2}^2+...$ , 
             ${x_{2}}_t=0.0997x_{1}x_{5}+0.779x_{2}x_{6}+0.209x_{1}x_{3}^2+...$ \\
        \addlinespace
            & ${x_{3}}_t=-0.157x_{1}x_{5}+1.1x_{2}x_{6}+0.0907x_{1}x_{2}^2+...$ , ${x_{4}}_t=0.375x_{1}x_{4}-0.187x_{2}x_{5}+0.0273x_{1}^2x_{2}+...$ \\
        \addlinespace
            & ${x_{5}}_t=-0.923x_{1}x_{4}+0.07x_{1}x_{2}^2+0.263x_{1}x_{3}^2+...$ , ${x_{6}}_t=-0.923x_{1}x_{4}+0.07x_{1}x_{2}^2+0.263x_{1}x_{3}^2+...$ \\
        \hline
        BSL & ${x_{1}}_t=10.8x_2$ , 
             ${x_{2}}_t=-18.7x_6$ \\
        \addlinespace
            & ${x_{3}}_t=0.578$ , ${x_{4}}_t=15.9x_6$ \\
        \addlinespace
            & ${x_{5}}_t=2.1$ , ${x_{6}}_t=13.8x_2$ \\
        \hline
        \ours & ${x_{1}}_t=0.933x_{2}x_{6}-0.932x_5x_6-1.16x_1+9.22$ , 
             ${x_{2}}_t=1.06x_{1}x_{3}-0.969x_1x_6-1.28x_2+7.26$ \\
        \addlinespace
            & ${x_{3}}_t=0.906x_{2}x_{4}-0.988x_1x_2-0.594x_3+6.75$ , ${x_{4}}_t=1.1x_{3}x_{5}-1.05x_2x_3-1.26x_4+8.37$ \\
        \addlinespace
            & ${x_{5}}_t=0.919x_{4}x_{6}-0.947x_3x_4-0.81x_5+7.47$ , ${x_{6}}_t=0.985x_{1}x_{5}-0.992x_4x_5-1.02x_6+9.66$\\
		\bottomrule
	\end{tabular}
	\caption{\small 10\% noise } \label{tb:mesh-szie}
 \end{subtable}
	
\caption{\small Discovery result for the Lorenz 96 system with 12 measurement examples. }\label{tb:discover-lorenz}
 
\end{table*}

\begin{table*}[h]
	%\begin{wraptable}{r}{0.7\textwidth}
	\vspace{-0.01in}
	\small
 \begin{subtable}{\textwidth}
 \centering
	\begin{tabular}[c]{cc}
		\toprule
	      \textit{Name}  & Equation\\
		\hline 
        True & $u_t=-uu_x+0.1u_{xx}$ \\
        \hline
        SINDy  & $u_t=-0.135u_x-0.186uu_x-0.486uu_{xx}+...$\\
        \hline
        PINN-SR & $u_t=-uu_x+0.1u_{xx}$\\
        \hline
        BSL & $u_t=-0.218uu_x$\\
        \hline
        \ours & $u_t=-uu_x+0.1u_{xx}$\\
		\bottomrule
	\end{tabular}
	\caption{\small zero noise } \label{tb:mesh-szie}
 \end{subtable}
 \begin{subtable}{\textwidth}
     \centering
	\begin{tabular}[c]{cc}
		\toprule
	      \textit{Name}  & Equation\\
		\hline 
        True & $u_t=-uu_x+0.1u_{xx}$ \\
        \hline
        SINDy  & $u_t=-0.2u_x+0.147u_{xx}-0.715uu_{x}$\\
        \hline
        PINN-SR & $u_t=-1.03uu_x+0.0313u_{xx}+0.48u^3+...$\\
        \hline
        BSL & $u_t=-0.372u_x$\\
        \hline
        \ours & $u_t=-1.04uu_x+0.0873u_{xx}$\\
		\bottomrule
	\end{tabular}
	\caption{\small 20\% noise } \label{tb:mesh-szie}
 \end{subtable}
	
\caption{\small Discovery result for the Burgers' Equation ($\nu = 0.1$) with $10 \times 10$ measurement examples. }\label{tb:discover-burgers}
\end{table*}

\begin{table*}[h]
	%\begin{wraptable}{r}{0.7\textwidth}
	\vspace{-0.01in}
	\small
 \begin{subtable}{\textwidth}
 \centering
	\begin{tabular}[c]{cc}
		\toprule
	      \textit{Name}  & Equation\\
		\hline 
        True & $u_t=-uu_x+0.01u_{xx}$ \\
        \hline
        SINDy  & $u_t=-0.177u_x-1.101uu_x+0.143uu_{xxx}+...$\\
        \hline
        PINN-SR & $u_t=-uu_x+0.0593u_{xx}-0.0515u_x+...$\\
        \hline
        BSL & $u_t=-1.72uu_x+6.023u^3u_x-5.38u^4u_x$\\
        \hline
        \ours & $u_t=-0.998uu_x+0.01u_{xx}$\\
		\bottomrule
	\end{tabular}
	\caption{\small zero noise. } \label{tb:mesh-szie}
 \end{subtable}
 \begin{subtable}{\textwidth}
     \centering
	\begin{tabular}[c]{cc}
		\toprule
	      \textit{Name}  & Equation\\
		\hline 
        True & $u_t=-uu_x+0.01u_{xx}$ \\
        \hline
        SINDy  & $u_t=-0.192u_x-0.745uu_x+0.064uu_{xxx}+...$\\
        \hline
        PINN-SR & $u_t=-0.697uu_x-0.206u_x$\\
        \hline
        BSL & $u_t=-0.406u_x$\\
        \hline
        \ours &  $u_t=-0.992uu_x+0.00963u_{xx}$\\
		\bottomrule
	\end{tabular}
	\caption{\small 20\% noise } \label{tb:mesh-szie}
 \end{subtable}
	
\caption{\small Discovery result for the Burgers' Equation ($\nu = 0.01$) with $50 \times 50$ measurement examples. }\label{tb:discover-burgers2}

\end{table*}

\begin{table*}[h]
	%\begin{wraptable}{r}{0.7\textwidth}
	\vspace{-0.01in}
	\small
 \begin{subtable}{\textwidth}
 \centering
	\begin{tabular}[c]{cc}
		\toprule
	      \textit{Name}  & Equation\\
		\hline 
        True & $u_t=-uu_x+0.005u_{xx}$ \\
        \hline
        SINDy  & $u_t=-0.293u_x-0.651uu_x+0.077uu_{xxx}+...$\\
        \hline
        PINN-SR & $u_t=-0.958uu_x+0.262uu_{xx}-1.06u^2u_{xx}+...$\\
        \hline
        BSL & $u_t=-3.24uu_x+8.77u^2u_x-9.55u^3u_x+...$\\
        \hline
        \ours & $u_t=-uu_x+0.00634u_{xx}$\\
		\bottomrule
	\end{tabular}
	\caption{\small zero noise. } \label{tb:mesh-szie}
 \end{subtable}
 \begin{subtable}{\textwidth}
     \centering
	\begin{tabular}[c]{cc}
		\toprule
	      \textit{Name}  & Equation\\
		\hline 
        True & $u_t=-uu_x+0.005u_{xx}$ \\
        \hline
        SINDy  & $u_t=-0.222u_x+0.022u_{xx}-0.266uu_x+...$\\
        \hline
        PINN-SR & $u_t=-0.426u_x$\\
        \hline
        BSL & $u_t=-9.15u^2u_x+21.4u^3u_x-13.7u^4u_x$\\
        \hline
        \ours & $u_t=-1.0044uu_x+0.0066u_{xx}$\\
		\bottomrule
	\end{tabular}
	\caption{\small 20\% noise } \label{tb:mesh-szie}
 \end{subtable}
	
\caption{\small Discovery result for the Burgers' Equation ($\nu = 0.005$) with $50 \times 50$ measurement examples. }\label{tb:discover-burgers3}
\end{table*}

\begin{table*}[h]
	%\begin{wraptable}{r}{0.7\textwidth}
	\vspace{-0.01in}
	\small
 \begin{subtable}{\textwidth}
 \centering
	\begin{tabular}[c]{cc}
		\toprule
	      \textit{Name}  & Equation\\
		\hline 
        True & $u_t=-uu_x-u_{xx} - u_{xxxx}$\\
        \hline
        SINDy  & $u_t=-0.068uu_x+0.049uu_{xxx}-0.02u^2u_{xxxx}+...$\\
        \hline
        PINN-SR & $u_t=0.0856u_x+0.0684uu_{xx}-0.0566u^2u_{xxx}$\\
        \hline
        BSL & $u_t=-0.115u_x$\\
        \hline
        \ours & $u_t=-0.988uu_x-0.986u_{xx}-0.988u_{xxxx}$\\
		\bottomrule
	\end{tabular}
	\caption{\small zero noise} \label{tb:mesh-szie}
 \end{subtable}
 \begin{subtable}{\textwidth}
     \centering
	\begin{tabular}[c]{cc}
		\toprule
	      \textit{Name}  & Equation\\
		\hline 
        True & $u_t=-uu_x-u_{xx}-u_{xxxx}$ \\
        \hline
        SINDy  & $u_t=-0.035u-0.104uu_x+0.024uu_{xxx}+...$\\
        \hline
        PINN-SR & $u_t=0.0717uu_{xxx}+0.0374uu_{xx}-0.034u_{xx}$\\
        \hline
        BSL & $u_t=-0.115u_x$\\
        \hline
        \ours & $u_t=-0.97uu_x-0.978u_{xx}-0.977u_{xxxx}$\\
		\bottomrule
	\end{tabular}
	\caption{\small 20\% noise } \label{tb:mesh-szie}
 \end{subtable}
	
\caption{\small Discovery result for the KS Equation with $40 \times 40$ measurement examples. }\label{tb:discover-ks}
\end{table*}

\begin{table*}[h]
	%\begin{wraptable}{r}{0.7\textwidth}
	\vspace{-0.01in}
	\small
 \begin{subtable}{\textwidth}
 \centering
	\begin{tabular}[c]{cc}
		\toprule
	      \textit{Name}  & Equation\\
		\hline 
        True & $u_t=0.0001u_{xx}+5u-5u^3$\\
        \hline
        SINDy  & $u_t=4.99u-4.99u^3$\\
        \hline
        PINN-SR & $u_t=4.16u-3.7u^3+0.043u^4+...$\\
        \hline
        BSL & $u_t=4.53u-4.33u^3+0.128u^2$\\
        \hline
        \ours & $u_t=0.0001u_{xx}+4.99u-4.99u^3$\\
		\bottomrule
	\end{tabular}
	\caption{\small zero noise. } \label{tb:mesh-szie}
 \end{subtable}
 \begin{subtable}{\textwidth}
     \centering
	\begin{tabular}[c]{cc}
		\toprule
	      \textit{Name}  & Equation\\
		\hline 
        True & $u_t=0.0001u_{xx}+5u-5u^3$ \\
        \hline
        SINDy  & $u_t=3.85u-3.49u^3$\\
        \hline
        PINN-SR & $u_t=2.94u-2u^3+0.063u^4+...$\\
        \hline
        BSL & $u_t=-0.0099u_x+4.55u-4.42u^3+...$\\
        \hline
        \ours & $u_t=0.000104u_{xx}+4.79u-4.76u^3$\\
		\bottomrule
	\end{tabular}
	\caption{\small 10\% noise } \label{tb:mesh-szie}
 \end{subtable}
	
\caption{\small Discovery result for the Allen-cahn Equation with $26 \times 101$ measurement examples. }\label{tb:discover-allen-cahn}
\end{table*}

\begin{table*}[h]
	%\begin{wraptable}{r}{0.7\textwidth}
	\vspace{-0.01in}
	\small
     \centering
	\begin{tabular}[c]{cc}
		\toprule
	      \textit{Name}  & Equation\\
		\hline 
        True & $x_t=0.48x-0.0248xy$ \\
        \addlinespace
        & $y_t=-0.927y+0.0276xy$        \\
        \hline
        SINDy  & $x_t=0.581x-0.0261xy$ \\
        \addlinespace
        & $y_t=0.255x-0.27y$        \\
        \hline
        PINN-SR & $x_t=-13.9y$ \\
        \addlinespace
        & $y_t=-0.114y$        \\
        \hline
        BSL & $x_t=0.512x-0.0266xy$ \\
        \addlinespace
        & $y_t=-0.926y+0.0279xy$        \\
        \hline
        \ours & $x_t=0.506x-0.0255xy$ \\
        \addlinespace
        & $y_t=-0.925y+0.0273xy$        \\
		\bottomrule
	\end{tabular}
	
\caption{\small Discovery result for the real-world predator-prey system with 21 observed examples. }\label{tb:discover-predator-prey}
\end{table*}

%\section{More Discovered Solutions} \label{sect:discovered-eq}

% \clearpage
% \bibliographystyle{apalike}
% \bibliography{GovEqGP}

\end{document}